\documentclass[10pt,twocolumn,letterpaper]{article}

\usepackage[pagenumbers]{cvpr} 
\usepackage{multirow}
\usepackage{amssymb}
\usepackage{pifont}

\usepackage{colortbl}
\usepackage{tikzducks}
\usepackage{graphicx}
\usepackage[accsupp]{axessibility}
\newcommand{\red}[1]{{\color{red}#1}}
\newcommand{\blue}[1]{{\color{cvprblue}#1}}

\usepackage{soul}
\usepackage{color}
\usepackage{graphicx}
\usepackage{tabularx}
\newcolumntype{L}{X}
\newcolumntype{C}{>{\centering \arraybackslash}X}
\newcolumntype{R}{>{\raggedright \arraybackslash}X}

\usepackage{makecell}

\newcommand{\brho}[0]{\boldsymbol{\rho}}
\newcommand{\bx}[0]{\boldsymbol{x}}
\newcommand{\by}[0]{\boldsymbol{y}}
\newcommand{\btheta}[0]{\boldsymbol{\theta}}

\usepackage{algorithm}
\usepackage{algorithmic}

\definecolor{cvprblue}{rgb}{0.21,0.49,0.74}
\usepackage[pagebackref,breaklinks,colorlinks,citecolor=cvprblue]{hyperref}

\def\paperID{****} 
\def\confName{CVPR}
\def\confYear{2024}

\title{Think Twice Before Selection: Federated Evidential Active Learning for Medical Image Analysis with Domain Shifts}

\author{
Jiayi Chen$^{1^*}$~~~Benteng Ma$^{2^*}$~~~Hengfei Cui$^{1}$~~~Yong Xia$^{{1, 3}^\dag}$\\
$^{1}$ School of Computer Science and Engineering, Northwestern Polytechnical University, China\\
$^{2}$ Hong Kong University of Science and Technology, Hong Kong SAR, China\\
$^{3}$ Research \& Development Institute of Northwestern Polytechnical University in Shenzhen, China\\
{\tt\small jiayichen@mail.nwpu.edu.cn, bentengma@ust.hk, hfcui@nwpu.edu.cn, yxia@nwpu.edu.cn}
}

\begin{document}

\maketitle
\begin{abstract}
Federated learning facilitates the collaborative learning of a global model across multiple distributed medical institutions without centralizing data. Nevertheless, the expensive cost of annotation on local clients remains an obstacle to effectively utilizing local data. To mitigate this issue, federated active learning methods suggest leveraging local and global model predictions to select a relatively small amount of informative local data for annotation. However, existing methods mainly focus on all local data sampled from the same domain, making them unreliable in realistic medical scenarios with domain shifts among different clients. In this paper, we make the first attempt to assess the informativeness of local data derived from diverse domains and propose a novel methodology termed \textbf{F}ederated \textbf{E}vidential \textbf{A}ctive \textbf{L}earning (FEAL) to calibrate the data evaluation under domain shift. Specifically, we introduce a Dirichlet prior distribution in both local and global models to treat the prediction as a distribution over the probability simplex and capture both aleatoric and epistemic uncertainties by using the Dirichlet-based evidential model. Then we employ the epistemic uncertainty to calibrate the aleatoric uncertainty. Afterward, we design a diversity relaxation strategy to reduce data redundancy and maintain data diversity. Extensive experiments and analysis on five real multi-center medical image datasets demonstrate the superiority of FEAL over the state-of-the-art active learning methods in federated scenarios with domain shifts. The code will be available at \url{https://github.com/JiayiChen815/FEAL}.
\end{abstract}    
\vspace{-4mm}
\renewcommand{\thefootnote}{}
\footnote{$^*$Equal contribution. $^\dag$Yong Xia is the corresponding author. This work was supported in part by Shenzhen Science and Technology Program under Grants JCYJ20220530161616036, National Natural Science Foundation of China under Grants 62171377 and 62271405, Ningbo Clinical Research Center for Medical Imaging under Grant 2021L003 (Open Project: 2022LYKFZD06), and Foshan HKUST Projects under Grants FSUST21-HKUST10E and FSUST21-HKUST11E. 
}

\section{Introduction}
\begin{figure}[t]
    \centering
    \begin{minipage}[t]{\linewidth}
        \centering
        \includegraphics[width=0.95\linewidth]{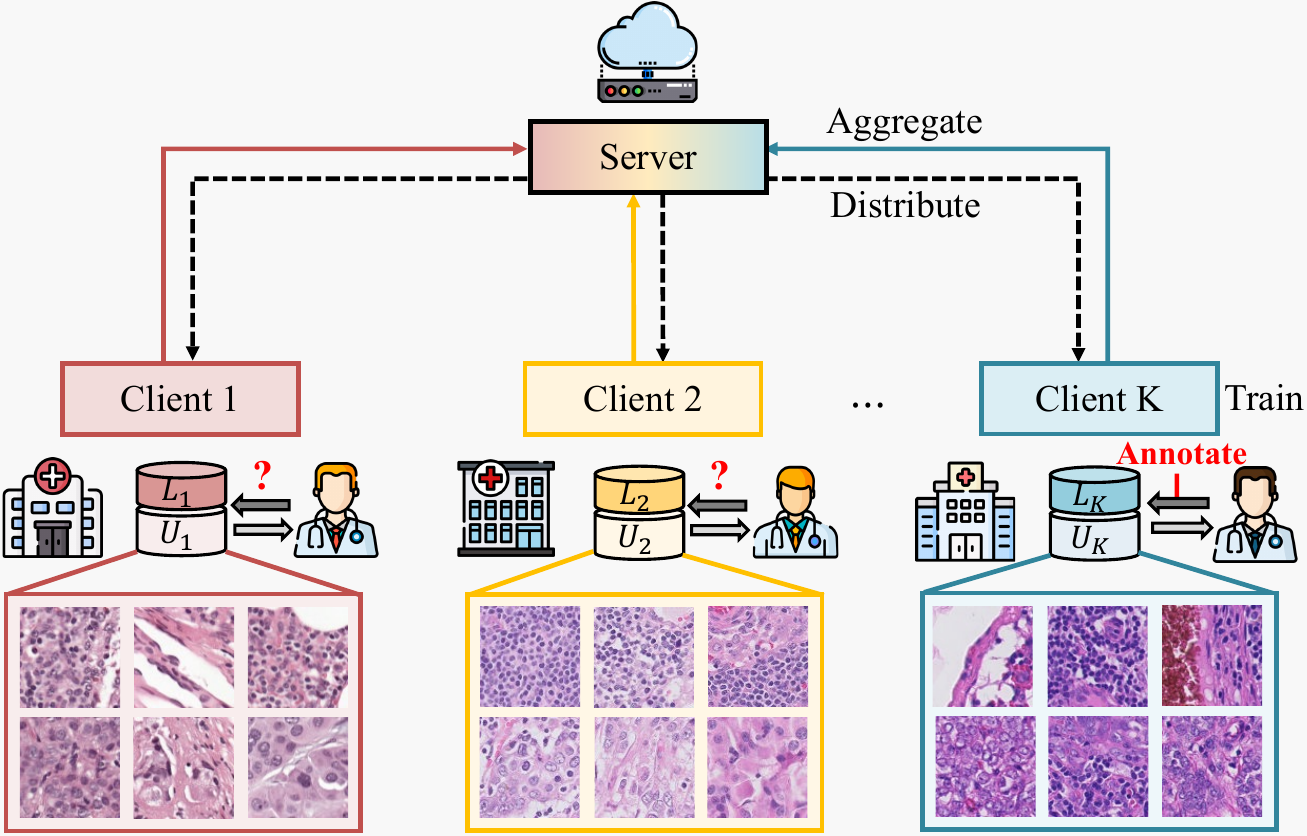}
        \subcaption{FAL scheme}
    \end{minipage}
    \\
    \begin{minipage}[t]{0.475\linewidth}
        \centering
        \includegraphics[width=\linewidth]{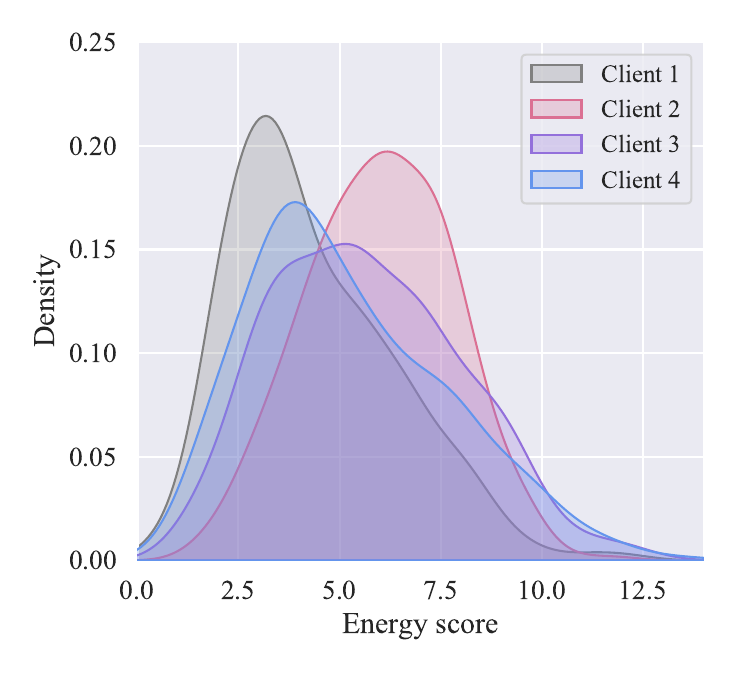}
        \subcaption{KDE of energy score}
    \end{minipage}
    \begin{minipage}[t]{0.475\linewidth}
        \centering
        \includegraphics[width=\linewidth]{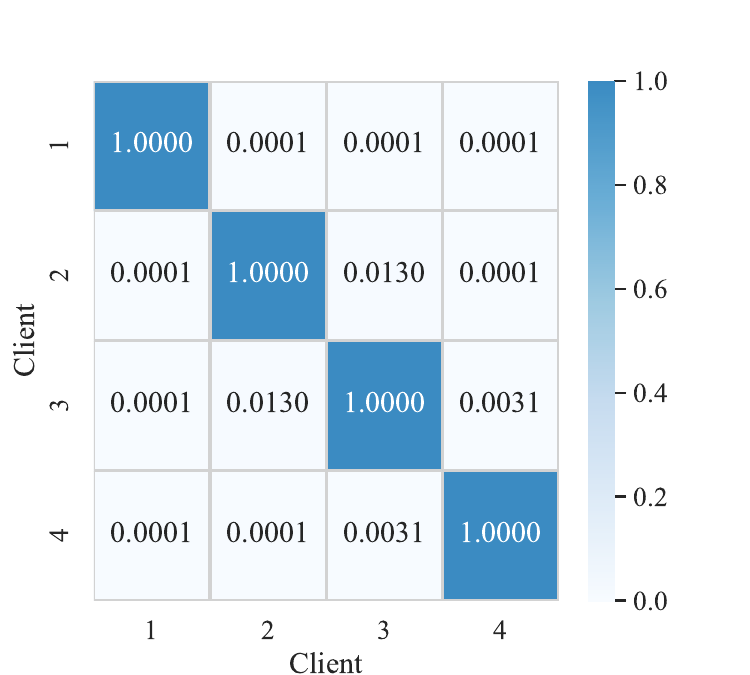}
        \subcaption{$p$-value}
    \end{minipage}
    \vspace{-2mm}
    \caption{\textbf{Illustration of federated active learning (FAL) in the presence of domain shift.} (a) FAL comprises model distribution, local training, model aggregation, and data annotation. (b) The KDE of energy scores depicts domain shifts across clients. (c) The low $p$-values in cross-client KDE of energy scores indicate the existence of significant domain shifts between all client pairs.}
    \vspace{-2mm}
    \label{fig:open}
\end{figure}

Federated learning enables collaborative learning across multiple clinical institutions (\textit{i.e.}, clients) to learn a unified model on the central server through model aggregation while preserving the data privacy at each client~\cite{konevcny2016federated, mcmahan2017communication, wang2022personalizing} (see Fig.~\ref{fig:open} (a)). 
Unfortunately, such a learning pipeline requires each client to prepare its own labeled data, whose scale is constrained by the available expertise, time, and budget for data annotation.

One possible solution to alleviate the annotation cost is to select a part of highly informative data to annotate. Active learning (AL) has shown great potential in guiding the data selection process~\cite{ahn2022federated, wu2022federated, kim2023re, cao2023knowledge}, leading to the federated AL (FAL) framework. Such a pipeline~\cite{monarch2021human, sener2017active, yoo2019learning, huang2021semi, caramalau2021sequential, parvaneh2022active} allows each client to assess the informativeness of unlabeled data using either the local model at each client or the global model from the server, greatly alleviating the heavy annotation costs while retaining great performance.
Nevertheless, when using a local model to select data, there is a bias toward prioritizing the data that improves the local updates while disregarding the overall generalizability of the global model.
Client models trained on diverse domains may exhibit significant divergence within the parameter space, making the use of a global model aggregated from these models for data selection unreliable.

Recent advances in FAL, \textit{e.g.}, LoGo~\cite{kim2023re} and KAFAL~\cite{cao2023knowledge}, tend to harness the knowledge of both local and global models to identify informative samples. Although this strategy has been proven to be more effective than employing a single model, these methods focus mainly on the class imbalance issue while assuming that the data at multiple clients is from the same domain.
However, the domain shift across clients is commonly seen in real-world applications, which is evidenced by the extremely low $p$-values of the kernel density estimation (KDE) of energy scores \cite{liu2020energy} (see Fig.~\ref{fig:open} (b) and (c)).
The existence of domain shift renders two major challenges for FAL.
\textbf{(1) Overconfidence;}  
Existing FAL methods evaluate data uncertainty based on the softmax prediction made by a deterministic model, which is essentially a point estimate and can be miscalibrated easily on data with domain shifts~\cite{liu2020energy, pearce2020uncertainty,ma2024VNAS}, resulting in unreliable uncertainty evaluation. 
\textbf{(2) Limited uncertainty representation.} 
Uncertainty can be divided into aleatoric uncertainty (or data uncertainty) and epistemic uncertainty (or knowledge uncertainty)~\cite{pearce2021understanding}. The former reflects the inherent complexity of data, such as class overlap and instance noise~\cite{ulmer2023prior}. The latter captures the restricted knowledge of a model caused by insufficient data or domain shifts. 
The softmax prediction can represent the aleatoric uncertainty but fails to capture the epistemic uncertainty, resulting in incomplete evaluations, which are particularly noticeable in the presence of domain shift.

To address both challenges, we propose the \textbf{Federated Evidential Active Learning (FEAL)} method. 
Built upon the Dirichlet-based evidential model~\cite{sensoy2018evidential, xie2023dirichlet}, FEAL treats the categorical prediction of a sample as following a Dirichlet distribution, thus allowing multiple potential predictions for a sample.
FEAL comprises two key modules, \textit{i.e.}, calibrated evidential sampling (CES) and evidential model learning (EML).
CES is a novel FAL sampling strategy that incorporates both uncertainty and diversity measures. It utilizes the expected entropy of potential predictions to quantify aleatoric uncertainty and aggregates the aleatoric uncertainty in both global and local models. Further, CES employs the differential entropy of the Dirichlet distribution to characterize the epistemic uncertainty~\cite{shen2023post} and utilizes the epistemic uncertainty in the global model to calibrate the aggregated aleatoric uncertainty. To enhance data selection, diversity relaxation is also employed with the local model to reduce redundancy and maintain diversity among the selected samples.
In addition to active sampling, we introduce evidence regularization in EML for accurate evidence representation and data assessment.
The main contributions of this work are summarized as follows:
\begin{itemize}
    \item We explore a rarely studied problem, FAL with domain shifts, which aims to attain a global model with a limited annotation budget for local clients amidst domain shifts. 
    \item We propose the FEAL method, with a sampling strategy CES and a local training scheme EML, to tackle the challenges in FAL with domain shifts. CES is designed to select informative samples by leveraging aleatoric and epistemic uncertainty with both global and local models and retaining sample diversity. EML is developed to regularize the evidence for improved data evaluation.
    \item We conduct extensive experiments on five real multi-center medical image datasets, comprising two datasets for classification and three datasets for segmentation. The results suggest the superiority of our FEAL method over its AL and FAL counterparts.
\end{itemize}

\section{Related Work}
\label{sec:related_work}
\subsection{Federated Learning with Domain Shifts}
Domain shift is a long-standing challenge for federated learning. Previous approaches can be divided into regularization-based, aggregation-based, and personalized ones.
\textbf{Regularization-based methods} implemented regularization on model parameters~\cite{li2020federated, shoham2019overcoming, karimireddy2020scaffold} or feature embeddings~\cite{li2021model, huang2023rethinking, guo2023fedbr, wu2023fediic} to address the objective inconsistency induced by domain shift. \textbf{Aggregation-based methods} dynamically adjust aggregation weights based on data quality~\cite{ma2023federated}, estimated client contribution~\cite{jiang2023fair}, generalization gap between global and local models~\cite{zhang2023federated}, layer-wise divergence~\cite{rehman2023dawa} or performance on proxy dataset~\cite{li2023revisiting}. \textbf{Personalized methods} aggregated domain-agnostic layers, while customizing domain-specific layers for local clients, including batch normalization (BN)~\cite{li2021fedbn}, high-frequency convolution~\cite{chen2021personalized} and prediction layers~\cite{wang2022personalizing}. Additionally, several methods enhanced data diversity~\cite{liu2021feddg, zhou2023fedfa} to refine data distribution and mitigate statistical heterogeneity~\cite{ye2023heterogeneous}. 
These approaches strive to mitigate the impact of domain shifts across clients in supervised scenarios with fully annotated training samples. Unfortunately, they ignore the substantial annotation costs for each client. In contrast, we further leverage active learning to reduce annotation costs by selecting the most informative data and propose a label-efficient method for federated learning with domain shifts.

\subsection{AL Methods}
Conventional AL methods can be categorized into uncertainty-based, diversity-based, and hybrid ones. 
\textbf{Uncertainty-based AL methods} aim to select the most ambiguous unlabeled samples for annotation. Classical approaches such as least confidence sampling~\cite{settles2009active}, margin-based sampling~\cite{monarch2021human}, and entropy-based sampling~\cite{shannon1948mathematical} evaluate the data uncertainty based on categorical probabilities. Yoo \textit{et al.}~\cite{yoo2019learning} and Huang \textit{et al.}~\cite{huang2021semi} estimated the loss for uncertainty assessment.
Moreover, several approaches assess the data uncertainty by analyzing the prediction inconsistency among multiple augmented samples~\cite{gao2020consistency}, standard and dropout inferences~\cite{gal2016dropout,gal2017deep}, or original and disturbed features~\cite{parvaneh2022active}. 
\textbf{Diversity-based AL methods} aim to identify a subset of samples that captures the distribution of the complete dataset. A variety of approaches have been proposed that exploit core-set techniques~\cite{sener2017active, caramalau2021sequential} or clustering methods~\cite{nguyen2004active, kutsuna2012active, urner2013plal} in the latent feature space, incorporate a diversity constraint in the optimization process~\cite{elhamifar2013convex,yang2015multi}, or model the distribution discrepancy between labeled and unlabeled samples~\cite{li2020attention} in order to identify a diverse collection of samples.
\textbf{Hybrid AL methods} exploit both uncertainty and diversity in their sampling strategies. Ash \textit{et al.}~\cite{ash2020deep} clustered the gradient embeddings to guarantee both uncertainty and diversity. A two-stage sampling strategy has also been implemented~\cite{parvaneh2022active, wang2023mhpl, yuan2023bi3d}.
However, these methods primarily focus on data selection driven by aleatoric uncertainty, often neglecting its sufficiency and reliability in practical scenarios. In this work, we developed a Dirichlet-based evidential model to capture both aleatoric and epistemic uncertainties. We further leveraged the epistemic uncertainty to calibrate uncertainty estimates, enhancing their reliability in the context of domain shifts.

\subsection{FAL Methods}
FAL aims to enhance the annotation efficacy of each local client in decentralized learning. In contrast to the centralized scenarios, there exist two potential query-selector models in FAL~\cite{kim2023re}, including the global model and the local model. 
Both Wu \textit{et al.}~\cite{wu2022federated} and Ahn \textit{et al.}~\cite{ahn2022federated} exclusively utilized a singular model for data evaluation. Specifically, Wu \textit{et al.}~\cite{wu2022federated} introduced a hybrid metric that considers both the locally predicted loss and the local feature distances between unlabeled and labeled samples. By contrast, Ahn \textit{et al.}~\cite{ahn2022federated} argued that evaluating samples with the global model contributes to the objectives of federated learning and recommended applying sampling strategies solely with the global model.
Nevertheless, as demonstrated in~\cite{kim2023re}, the superiority of the two query-selector models depends on the global and local heterogeneous levels, and it is necessary to leverage the knowledge of both global and local models. 
Kim \textit{et al.}~\cite{kim2023re} proposed a hybrid metric called LoGo, which applies $k$-means clustering technique~\cite{macqueen1967some} on the gradient space of the local model and subsequently conducts cluster-wise sampling using the global model. 
Cao \textit{et al.}~\cite{cao2023knowledge} proposed a knowledge-specialized sampling strategy, which leverages the discrepancy between the global model and local model to assess data uncertainty. However, these methods focus on the local data from a singular domain, which is less realistic. Though partial approaches~\cite{kim2023re, cao2023knowledge} account for heterogeneity caused by class imbalance, they often neglect another heterogeneous property known as domain shifts. In this work, we propose the uncertainty calibration method to achieve reliable uncertainty evaluation with domain shifts across multiple clients.
\section{Methodology}
\label{sec:method}
\begin{figure*}[t]
    \centering
    \includegraphics[width=\linewidth]{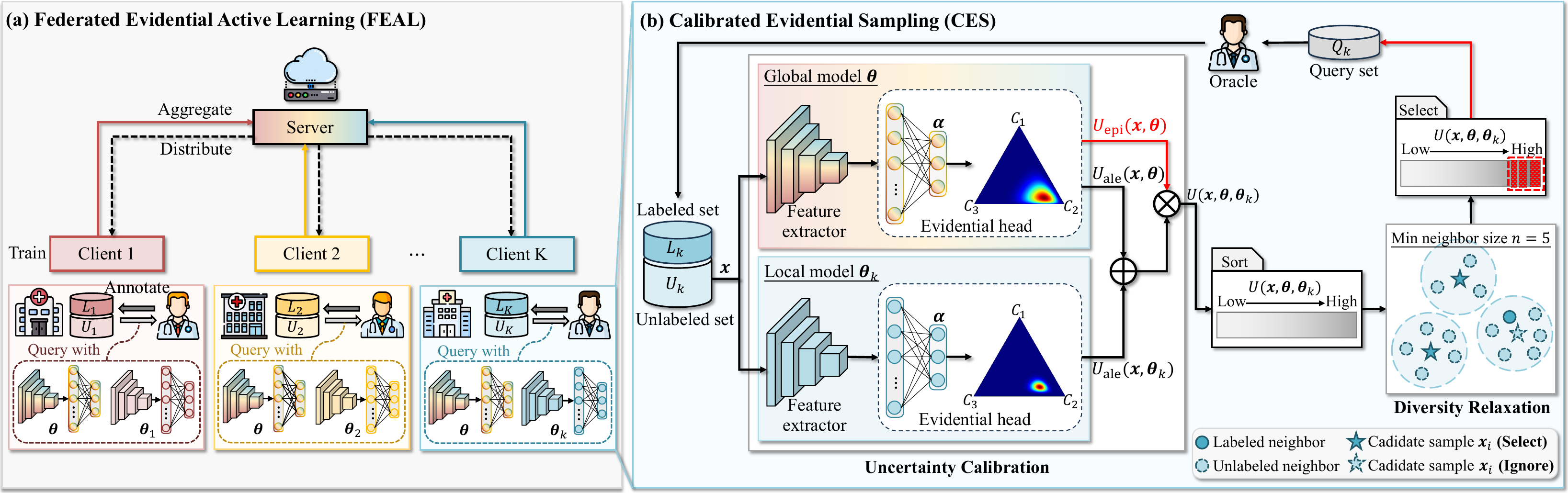}
    \caption{\textbf{Illustration of the proposed FEAL method. } (a) Overview of FEAL. (b) Illustration of CES module, including uncertainty calibration and diversity relaxation. }
    \label{fig:framework}
    \vspace{-2mm}
\end{figure*}

\subsection{Problem Formulation}
The overview of our FEAL framework is displayed in Fig.~\ref{fig:framework} and Appendix~A. Under this framework, we maintain $K$ local models $\{\btheta_k\}_{k=1}^K$ on clients and a global model $\btheta$ on the central server. The $k$-th local client contains a labeled set $L_k$ and an unlabeled set $U_k$.
FAL comprises two iterative phases: federated model training and local data annotation. Federated model training involves model distribution, local training, and model aggregation. In the first round, the $k$-th client randomly selects $B_k$ unlabeled samples and annotates them to form the initial labeled set $L_k^1=\{(\bx_i,\boldsymbol{y}_i)\}_{i=1}^{B_k}$, and the unlabeled set is updated to $U_k^1=U_k\setminus L_k^1$. In the $r$-th FAL round, the $k$-th client constructs the query set $Q_k^r=\{(\bx_j,\boldsymbol{y}_j)\}_{j=1}^{B_k}$ for annotation using the sampling strategy and updates the labeled set to $L_k^{r}=L_k^{r-1} \cup Q_k^r$, whereas the unlabeled set is updated to $U_k^{r}=U_k^{r-1}\setminus Q_k^{r}$. Subsequent federated model training proceeds with the updated labeled set $L_k^r$. The FAL process is repeated for $R$ times as required.

\subsection{Dirichlet-based Evidential Model in FAL}
For federated active learning, we employ a Dirichlet-based evidential model to effectively capture aleatoric and epistemic uncertainties in both global and local models. In this section, we begin by presenting the foundational formulation of the Dirichlet-based evidential model.

We start with the general $C$-class classification task. Given an input sample $\bx$ from the $k$-th client, a model $f$ parameterized with $\btheta$ projects $\bx$ into a $C$-dimensional logits $ f(\bx,\btheta)$. The classical CNN utilizes the softmax operator to transform the logits $ f(\bx,\btheta)$ into the prediction of class probabilities $\brho$. However, this approach essentially provides a single-point estimate of $\brho$ and can be easily miscalibrated on local data from diverse domains. The Dirichlet-based evidential model, on the other hand, views the categorical prediction $\brho$ as a random variable with a Dirichlet distribution $Dir(\brho|\boldsymbol{\alpha})$. The probability density function of $\brho$~\cite{sensoy2018evidential, xie2023dirichlet}, given $\bx$ and $\btheta$, is formulated as:
\begin{equation}
\small
\begin{aligned} 
p(\brho|\bx,\btheta)
    = \left\{
    \begin{aligned}
         \frac{\Gamma(\sum_{c=1}^C\alpha_c)}{\prod_{c=1}^C\Gamma(\alpha_c)}\prod_{c=1}^C\rho_c^{\alpha_c-1}&, (\brho\in \Delta^C) \\
         0\qquad\qquad&, (\text{otherwise}) 
    \end{aligned}
    \right.
\end{aligned}
\end{equation}
where $\boldsymbol{\alpha}$ denotes the parameters of the Dirichlet distribution for sample $\bx$, $\Gamma(\cdot)$ is the Gamma function, and $\Delta^C=\{\brho|\sum_{c=1}^C\rho_c{=}1\ \text{and}\ 0{<}\rho_c{<}1\}$ represents the $C$-dimensional unit simplex.

The posterior probability $P(y=c|\bx,\btheta)$ for class $c$, \textit{a.k.a.}, the expected categorical prediction $\overline{\rho}_c$, is given by:
\begin{equation}
\vspace{-1mm}
\small
    P(y=c|\bx,\btheta)=\int p(y=c|\brho)\cdot p(\brho|\bx,\btheta)\ \text{d}\brho=\frac{\alpha_c}{S},
    \label{eq:prosterior}
\end{equation}
where $S=\sum_{c=1}^C \alpha_c$ represents the Dirichlet strength. The derivation of Eq.~\ref{eq:prosterior} is provided in Appendix~B.1.

Drawing on concepts from Dempster-Shafer theory~\cite{sentz2002combination} and subjective logic~\cite{josang2016subjective}, the parameter $\boldsymbol{\alpha}$ is linked to the accumulated evidence $\boldsymbol{e}$ which quantifies the degree of support for the prediction on sample $\bx$. The parameter $\boldsymbol{\alpha}$ is derived as $\boldsymbol{\alpha} = \boldsymbol{e} + 1 = \mathcal{A}(f(\bx, \btheta)) + 1$, where $\mathcal{A}(\cdot)$ is a non-negative activation function that transforms the logits $f(\bx,\btheta)$ into evidence $\boldsymbol{e}$.

In our study, all local models adopt the same Dirichlet-based evidential architecture with the global model to communicate between local clients and the central server.

\subsection{Calibrated Evidential Sampling}
In the context of FAL with domain shifts, we integrate both uncertainty and diversity measures to identify the most informative samples for annotation (see Fig.~\ref{fig:framework}(b)). As for uncertainty evaluation, we leverage the epistemic uncertainty in the global model to calibrate the aleatoric uncertainty in both global and local models. We now delve into its details. 

\noindent\textbf{Aleatoric uncertainty.}
Dirichlet-based evidential models interpret the categorical prediction $\brho$ as a distribution rather than a singular point estimate, which acknowledges a range of possible predictions. We use the expected entropy of all possible predictions to deliver the aleatoric uncertainty~\cite{xie2023dirichlet} to quantify the inherent complexity or ambiguity present in local data. Given a sample $\bx$ and the global model $\btheta$, the aleatoric uncertainty of the sample $\bx$ in the global model $\btheta$ is represented as:
\begin{equation}
    \small
    \begin{aligned}
        U_{\text{ale}}(\bx, \btheta) &= \mathbb{E}_{p(\brho|\bx,\btheta)}[\mathcal{H}[P(y|\brho)]]\\
        &=-\sum_{c=1}^C \mathbb{E}_{p(\rho_c|\bx,\btheta)}[\rho_c \cdot \log\rho_c]\\
        &=\sum_{c=1}^C\frac{\alpha_c}{S}\cdot[\psi(S+1)-\psi(\alpha_c+1)],
    \end{aligned}
    \label{eq:ale}
\end{equation}
where $\mathcal{H}(\cdot)$ denotes the Shannon entropy~\cite{shannon1948mathematical}. Similarly, the aleatoric uncertainty in the local model $k$ is  $U_{\text{ale}}(\bx, \btheta_k)$. The derivation of Eq.~\ref{eq:ale} is in Appendix~B.2.

\noindent\textbf{Epistemic uncertainty.}
In the Dirichlet distribution, the differential entropy quantifies how dispersed the probabilities are across different categories~\cite{malinin2018predictive}. We employ the differential entropy of the Dirichlet distribution to quantify the epistemic uncertainty linked to domain shifts between the global model and local data. Specifically, given a sample $\bx$ and the global model $\btheta$, the epistemic uncertainty of the sample $\bx$ in the global model $\btheta$ is represented as:
\begin{equation}
    \small
    \begin{aligned}
        U_{\text{epi}}(\bx,\btheta) &= \mathcal{H}[p(\brho|\bx,\btheta)]\\
        &=-\int p(\brho|\bx,\btheta)\cdot \log p(\brho|\bx,\btheta)\ \text{d}\brho\\
        &=\sum_{c=1}^C\log\frac{\Gamma(\alpha_c)}{\Gamma(S)} - (\alpha_c-1)\cdot[\psi(\alpha_c)-\psi(S)].
    \end{aligned}
    \label{eq:epi}
\end{equation}
The derivation of Eq.~\ref{eq:epi} is in Appendix~B.2.

\noindent\textbf{Uncertainty calibration.}
Given a sample $\bx$, the global model $\btheta$, and the local model $\btheta_k$, we calculate the aleatoric uncertainty (Eq.~\ref{eq:ale}) in both global and local models and subsequently calibrate the aleatoric uncertainty by incorporating the epistemic uncertainty (Eq.~\ref{eq:epi}) from the global model. The overall calibrated uncertainty for sample $\bx$ is 
\begin{equation}
    \small
    U(\bx, \btheta, \btheta_k)=[U_{\text{ale}}(\bx, \btheta)+U_{\text{ale}}(\bx, \btheta_k)] \cdot U_{\text{epi}}(\bx, \btheta).
    \label{eq:uc}
\end{equation}

\noindent\textbf{Diversity relaxation.}
We adopt local constraints to ensure diversity among selected samples, contrasting with core-set techniques that impose global diversity constraints. As outlined in Alg.~\ref{alg:diversity}, we initially sort the unlabeled set $U_k^{r-1}$ by descending calibrated uncertainty $U(\bx, \btheta, \btheta_k)$, and then extract feature embeddings with local model $\btheta_k$. 
During the iteration over the unlabeled set $U_k^{r-1}$, we compute the cosine similarity $s(\bx_i, \bx_j)$ for each candidate sample $\bx_i$ against all other samples $\bx_j\in U_k^{r-1}\setminus \bx_i$ and form a neighbor set $N(\bx_i)$ based on the similarity threshold $\tau$. A sample $\bx_i$ is selected if its neighbor counts $|N(\bx_i)|$ are less than the minimum neighbor size $n$ or if these neighbors remain unlabeled. Following this criterion, $B_k$ unlabeled samples are chosen to constitute the final set $Q_k^r$ for annotation, effectively balancing diversity and uncertainty in data selection.
\vspace{-6mm}
\begin{algorithm}[htbp]
\small
\caption{\small{Diversity Relaxation for Local Client $k$}}
\label{alg:diversity}
\begin{algorithmic}[1]
\REQUIRE unlabeled set $U_k^{r-1}$, local model $\btheta_k$, annotation budget $B_k$, similarity threshold $\tau$, minimum neighbor size $n$
\ENSURE query set $Q_k^r$
\STATE Sort $U_k^{r-1}$ by descending calibrated uncertainty.
\STATE Initialize index $i=1$ and query set $Q_k^r=\emptyset$.
\WHILE{$|Q_k^{r}|<B_k$ \AND $i \le |U_k^{r-1}|$}
    \STATE Select a candidate sample $\bx_i$ from $U_k^{r-1}$.
    \STATE Compute feature similarity $s(\bx_i, \bx_j)$ using $\btheta_k$, where $\bx_j \in U_k^{r-1}\setminus\bx_i$.
    \STATE Form neighbor set $N(\bx_i)$, including $\bx_j$ with $\!s(\bx_i, \bx_j)$$\ge$$\tau\!$.
    \vspace{-3mm}
    \IF{$|N(\bx_i)|<n$ \OR $N(\bx_i) \cap Q_k^r = \emptyset$}
        \STATE Add $\bx_i$ to $Q_k^r$.
    \ENDIF
    \STATE Increment $i$.
\ENDWHILE
\RETURN $Q_k^r$
\end{algorithmic}
\end{algorithm}

\vspace{-4mm}
\subsection{Evidential Model Learning}
Dirichlet-based evidential models treat the categorical prediction of a sample as a distribution, enabling multiple potential predictions to occur with specific probabilities. Considering all possible predictions, we adopt the Bayes risk of cross-entropy loss~\cite{sensoy2018evidential} as the task loss for classification tasks, formulated as follows:
\begin{equation}
\small
    \begin{aligned}
        \mathcal{L}_{\text{task}}(\bx,\btheta_k,\by) &= \int (\sum_{c=1}^C -y_c\log{\rho_c}) \cdot p(\brho|\bx,\btheta_k)\ \text{d}\brho\\
        &=\sum_{c=1}^C y_c\cdot [\psi(S)-\psi(\alpha_c)],
    \end{aligned}
    \label{eq:cls_loss}
\end{equation}
where $\psi(\cdot)$ is the digamma function and $y_c$ is the label indicator for class $c$. Similarly, the Bayes risk of Dice loss~\cite{li2022region} for segmentation tasks is:
\begin{equation}
\small
    \begin{aligned}
        \mathcal{L}_{\text{task}}(\bx,\btheta_k,\by) &= \int (1{-}\frac{2}{C}\sum_{c=1}^{C}\frac{{\Vert\boldsymbol{y}_{c}\circ \brho_{c}\Vert}_1}{{\Vert\boldsymbol{y}_{c}^2\Vert}_1{+}{\Vert\brho_{c}^2\Vert}_1}) \cdot p(\brho|\bx,\btheta_k)\ \text{d}\brho\\
        &= 1-\frac{2}{C}\sum_{c=1}^C\frac{{\Vert\boldsymbol{y}_{c}\circ\overline{\brho}_{c}\Vert}_1}{{\Vert\boldsymbol{y}_{c}^2\Vert}_1+{\Vert\overline{\brho}_{c}^2\Vert}_1+{\Vert\frac{\overline{\brho}_{c}\circ(\boldsymbol{1}-\overline{\brho}_{c})}{\boldsymbol{S}+\boldsymbol{1}}\Vert}_1},
    \end{aligned}
    \label{eq:seg_loss}
\end{equation}
where $\circ$ is the Hadamard product and the expected categorical probability of $\bx$ is $\overline{\brho}_{c} = \frac{\boldsymbol{\alpha}_{c}}{\boldsymbol{S}}$. The derivation of Eq.~\ref{eq:cls_loss} and Eq.~\ref{eq:seg_loss} are in Appendix~B.3.

We incorporate evidence regularization to further reduce incorrect evidence~\cite{sensoy2018evidential} and improve correct evidence~\cite{pandey2023learn}.
\begin{equation}
    \small
    \mathcal{L}_{\text{reg}}(\bx,\btheta_k,\by) {=} KL[Dir(\brho|\tilde{\boldsymbol{\alpha}})\Vert Dir(\brho|\boldsymbol{1})] - \frac{C}{S}{\cdot}f(\bx,\btheta_k),
\end{equation}
where $\tilde{\boldsymbol{\alpha}}=\boldsymbol{y} + (\boldsymbol{1}-\boldsymbol{y})\odot\boldsymbol{\alpha}$ and $KL(\cdot)$ denotes the Kullback-Leibler divergence~\cite{kullback1951information}. Notably, we calculate the average pixel-wise $\mathcal{L}_{\text{reg}}$ in segmentation.

The overall training objective, combining task loss $\mathcal{L}_{\text{task}}$ and evidence regularization $\mathcal{L}_{\text{reg}}$, is formulated as:
\begin{equation}
   \small
   \mathcal{L}(\bx,\btheta_k,\by) = \mathcal{L}_{\text{task}}(\bx,\btheta_k,\by) + \lambda\cdot \mathcal{L}_{\text{reg}}(\bx,\btheta_k,\by),
   \label{eq:overall_loss}
\end{equation}
where $\lambda$ is the trade-off weight. between the task loss and the regularization term.

\section{Experiments}
\begin{figure*}[t]
\centering
\begin{subfigure}{0.3\linewidth}
\centering
\includegraphics[width=\linewidth]{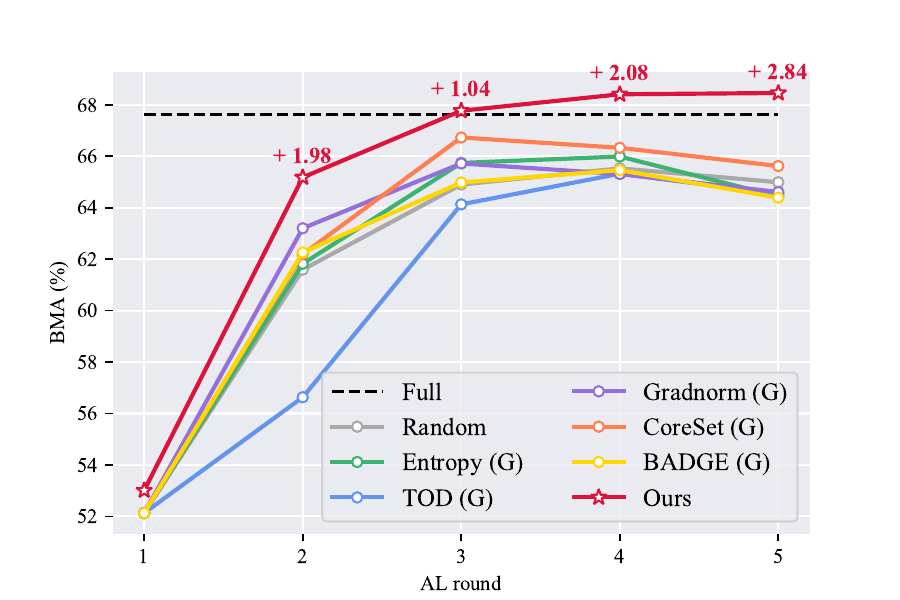}
\caption{Fed-ISIC ($G$)}
\end{subfigure}
\hspace{2mm}
\begin{subfigure}{0.3\linewidth}
\centering
\includegraphics[width=\linewidth]{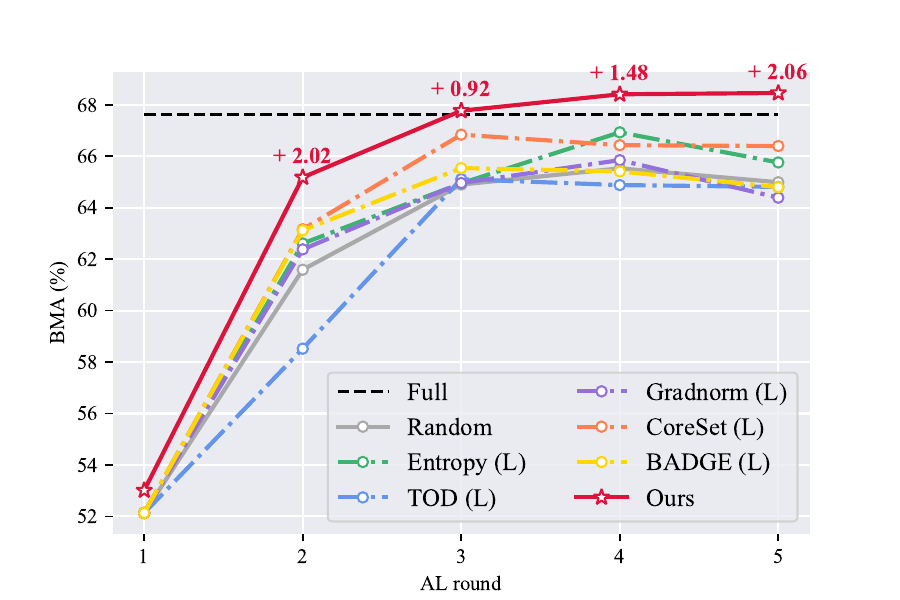}
\caption{Fed-ISIC ($L$)}
\end{subfigure}
\hspace{2mm}
\begin{subfigure}{0.3\linewidth}
\centering
\includegraphics[width=\linewidth]{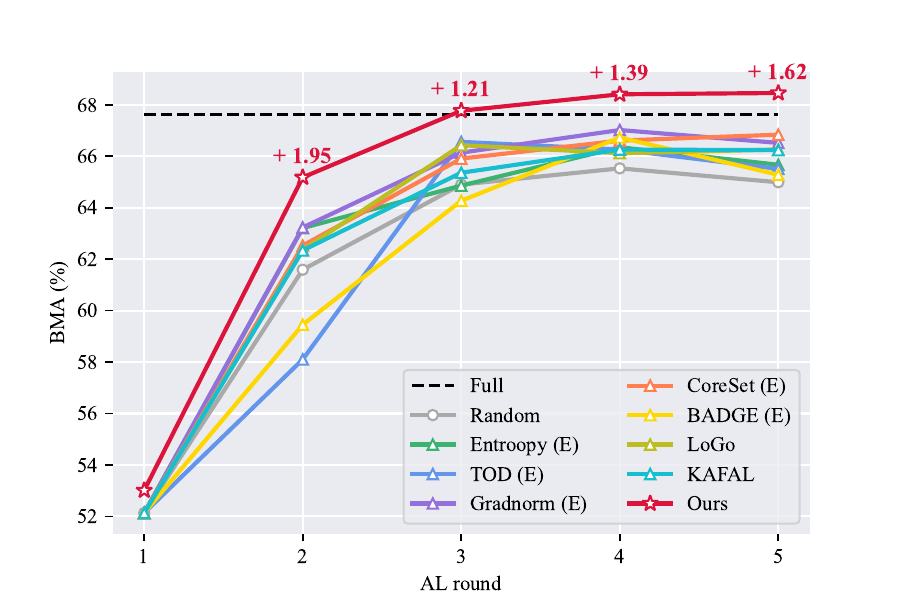}
\caption{Fed-ISIC ($E$)}
\end{subfigure}

\begin{subfigure}{0.3\linewidth}
\centering
\includegraphics[width=\linewidth]{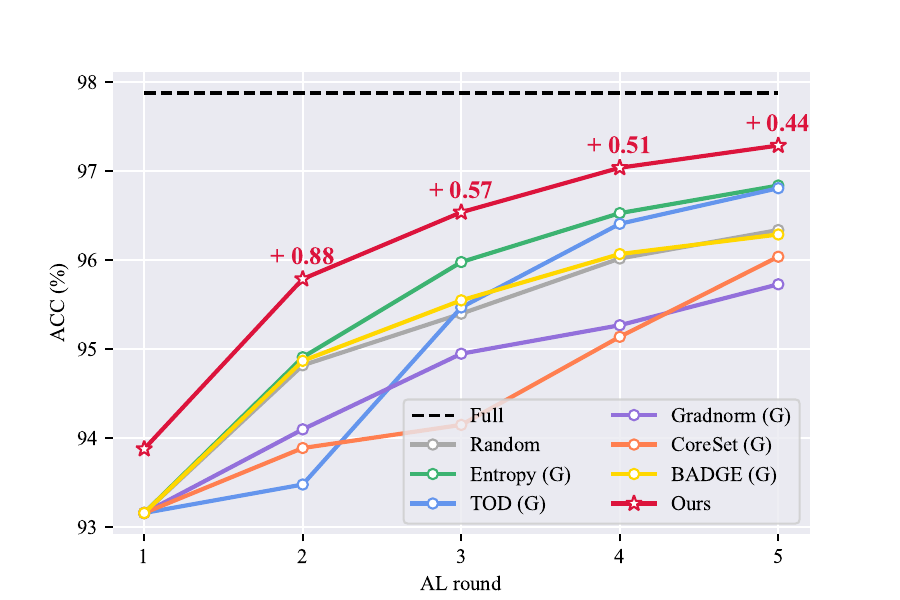}
\caption{Fed-Camelyon ($G$)}
\end{subfigure}
\hspace{2mm}
\begin{subfigure}{0.3\linewidth}
\centering
\includegraphics[width=\linewidth]{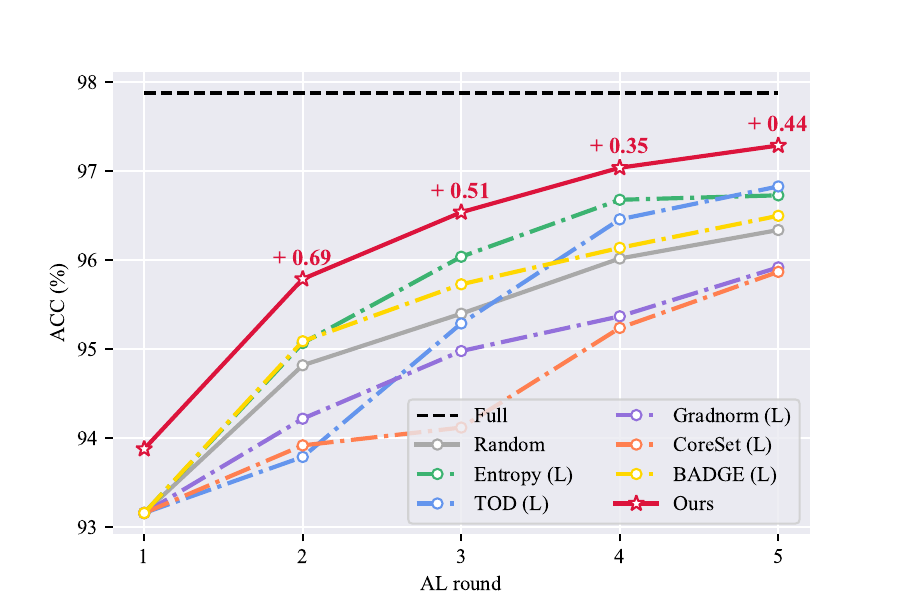}
\caption{Fed-Camelyon ($L$)}
\end{subfigure}
\hspace{2mm}
\begin{subfigure}{0.3\linewidth}
\centering
\includegraphics[width=\linewidth]{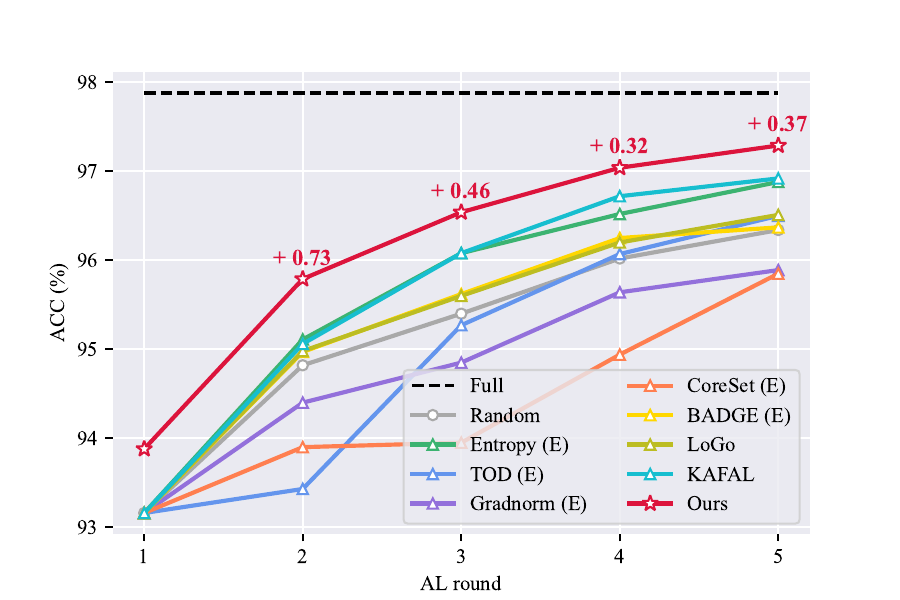}
\caption{Fed-Camelyon ($E$)}
\end{subfigure}
\vspace{-2mm}
\caption{\textbf{Comparison of FAL methods in medical image classification.} (a)-(c) and (d)-(f) depict the results of the Fed-ISIC and Fed-Camelyon datasets, respectively. Performance enhancements over the second-best method in each FAL round are emphasized in \red{red} text.}
\label{fig:main_results}
\vspace{-2.5mm}
\end{figure*}

\subsection{Experimental Settings}
\noindent\textbf{Datasets.}
We evaluated FEAL on five real multi-center medical image datasets, comprising two classification and three segmentation datasets.
The classification datasets included
\begin{itemize}
\item \textbf{Fed-ISIC}: A skin lesion dataset from 4 data sources~\cite{ogier2022flamby} containing \{12413, 3954, 3363, 2259\} images. 
\item \textbf{Fed-Camelyon}: A breast cancer histology dataset from 5 centers~\cite{jiang2022harmofl} comprising \{59436, 34904, 85054, 129838, 146722\} patches. 
\end{itemize}
The segmentation datasets included
\begin{itemize}
\item \textbf{Fed-Polyp}: A endoscopic polyp dataset from 4 centers~\cite{wang2022personalizing} with \{1000, 196, 379, 612\} samples.
\item \textbf{Fed-Prostate}: A prostate MRI dataset from 6 data sources~\cite{liu2021feddg} with \{261, 384, 158, 468, 421, 175\} slices.
\item \textbf{Fed-Fundus}: A retinal fundus dataset from 4 centers~\cite{liu2021feddg} with \{101, 159, 400, 400\} samples.
\end{itemize}
In our study, each dataset was divided using an 8:2 train-to-test split ratio at the patient level. Details of these datasets are provided in Appendix~C.1.

\noindent\textbf{Evaluation metrics.}
For classification, we utilized the Balanced Multi-class Accuracy (BMA) for skin lesion classification \cite{cassidy2022analysis} and measured accuracy (ACC) for breast cancer histology classification. In the context of segmentation, we used the Dice score and the 95\% Hausdorff Distance (HD95) to assess segmentation results.

\noindent\textbf{Implementation details.}
We conducted $R=5$ rounds of FAL involving federated model training and data annotation. The annotation budget $B_k$ is $500$ for Fed-ISIC and Fed-Camelyon, $50$ for Fed-Polyp, and $20$ for Fed-Prostate and Fed-Fundus.
During model training, we followed the previous work~\cite{wicaksana2023fca, jiang2022harmofl, wang2022personalizing} to utilize EfficientNet-B0~\cite{tan2019efficientnet} for Fed-ISIC, DenseNet-121~\cite{huang2017densely} for Fed-Camelyon, and U-Net~\cite{ronneberger2015u,liu2021feddg} for segmentation datasets. Notably, both EfficientNet-B0 and DenseNet-121 were pre{-}trained on ImageNet~\cite{deng2009imagenet}. 
Each experiment was conducted three times using different random seeds, and the average results were reported. More details are in Appendix~C.1.

\noindent\textbf{Comparison methods.}
We compared FEAL with eight FAL methods, including random sampling (Random), entropy-based sampling (Entropy)~\cite{shannon1948mathematical}, TOD~\cite{huang2021semi}, Gradnorm~\cite{wang2022boosting}, CoreSet~\cite{sener2017active}, BADGE~\cite{ash2020deep}, LoGo~\cite{kim2023re}, and KAFAL~\cite{cao2023knowledge}. The first six strategies are primarily developed for standard active learning, whereas LoGo and KAFAL are specifically tailored for decentralized scenarios. To incorporate these standard AL strategies into the FAL framework, we implemented them in three distinct manners: using only the global model (referred to as $G$), depending solely on the local model ($L$), or employing a simple ensemble method with both models ($E$). It guarantees a comprehensive evaluation of these strategies in FAL. Details of comparison methods are summarized in Appendix~C.1.

\subsection{Results}
\begin{table*}[t]
\caption{\textbf{Comparison of FAL methods in medical image segmentation.} Dice scores for three segmentation datasets are reported. For Fed-Fundus, Dice scores for both optic disc and optic cup segmentation and their average are presented. $G$ and $L$ stand for sampling solely with the global or local model, while $E$ represents sampling with both models. \red{Red} and \blue{blue} highlight the Top-1 and Top-2 results. 
}
\vspace{-2.5mm}
\scriptsize
\setlength{\tabcolsep}{2.7pt}
\begin{tabularx}{\linewidth}{c|l|cccc|cccc|cccc}
\bottomrule
\multirow{2}{*}{Model} & \multicolumn{1}{c|}{\multirow{2}{*}{Method}} & \multicolumn{4}{c|}{Fed-Polyp (\%)} & \multicolumn{4}{c|}{Fed-Prostate (\%)} & \multicolumn{4}{c}{Fed-Fundus (\%)} \\
& & R2 & R3 & R4 & R5 & R2 & R3 & R4 & R5 & R2 & R3 & R4 & R5 \\ \hline 
- & Full & \multicolumn{4}{c|}{78.18} & \multicolumn{4}{c|}{88.02} & \multicolumn{4}{c}{94.32 / 85.70 (90.01)} \\ \hline
- & Random & 67.70 & 72.16 & 75.58 & 76.32 & 80.29 & 82.70 & 83.94 & 84.77 & 92.30 / 81.41 (86.85)& 93.33 / 84.45 (88.89)& 94.29 / 84.80 (89.54) & 94.46 / 85.05 (89.76)\\ \hline
\multirow{5}{*}{$G$} & Entropy~\cite{shannon1948mathematical} & \multicolumn{1}{c}{67.45} & 74.65 & 75.30 & 76.69 & \multicolumn{1}{c}{82.17} & 82.53 & 84.05 & 86.10 & 93.19 / 82.61 (87.90) & 93.84 / 84.35 (89.10) & 94.27 / 85.34 (89.80) & 94.47 / 85.29 (89.88)\\
& TOD~\cite{huang2021semi} & 64.99 & 74.61 & 76.24 & 78.26 & 80.75 & 83.48 & 84.31 & 85.82 & 92.70 / 82.49 (87.60) & 93.95 / \blue{85.01} (89.48) & 94.27 / 85.63 (89.95) & \blue{94.71} /  85.58 (90.14) \\
& Gradnorm~\cite{wang2022boosting} & 69.14 & 74.58 & 75.79 & 78.51 & 82.10 & 83.01 & 84.85 & 86.02 & 93.20 / 82.01 (87.60) & 94.12 / 84.71 (89.41) & 94.33 / 85.38 (89.85) & 94.56 / 85.43 (89.99)\\
& CoreSet~\cite{sener2017active} & 69.50 & 73.37 & 76.71 & 78.18 & 82.11 & 83.68 & 84.56 & 85.86 & 93.00 / 83.07 (88.03) & 93.90 / 84.75 (89.32) & 94.16 / 85.35 (89.75)& 94.51 / 85.63 (90.07)\\
& BADGE~\cite{ash2020deep} & 70.09 & 74.11 & 76.38 & 76.55 & \blue{82.78} & 83.91 & 85.39 & 85.97 & 93.17 / 82.54 (87.85)& 94.07 / 84.46 (89.26) & 94.40 / 85.37 (89.89) & 94.58 / 85.19 (89.89)\\ \hline

\multirow{5}{*}{$L$}  & Entropy & 67.48 & 73.41 & 75.07 & 78.63 & 81.08 & 82.22 & 84.36 & 85.19 & 93.19 / \blue{83.22} (88.21) & 93.83 / 84.49 (89.16)& 94.36 / 84.97 (89.66) & 94.63 / 85.68 (90.15)\\
& TOD~\cite{huang2021semi} & 65.95 & 72.92 & 75.19 & 77.97 & 79.59 & 83.74 & 85.50 & 86.03 & 92.82 / 82.34 (87.58)& 93.98 / 85.00 (89.49) & 94.37 / 85.28 (89.83)& 94.65 / 85.56 (90.10)\\
& Gradnorm~\cite{wang2022boosting} & 70.06 & 74.69 & 77.25 & \blue{ 78.84} & 80.52 & 83.43 & 84.94 & 86.04 & 93.29 / 83.04 (88.16) & \blue{94.13} / 84.69 (89.41) & 94.33 / 85.60 \blue{(89.97)} & 94.42 / 85.53 (89.98) \\
& CoreSet~\cite{sener2017active} & 68.92 & 74.06 & 75.59 & 77.75 & 81.49 & 83.49 & 84.65 & 86.19 & 92.80 / 83.20 (88.00) & 93.87 / 84.70 (89.28) & 94.28 / 85.42 (89.85) & 94.47 / 85.54 (90.00)\\
& BADGE~\cite{ash2020deep} & \blue{ 70.28} & 73.96 & 76.21 & 77.63 & 82.07 & 83.54 & 85.30 & 86.06 & 93.06 / 82.65 (87.85)& 93.95 / 84.44 (89.19)& 94.34 / 85.02 (89.68)& 94.54 / 85.52 (90.03)\\ \hline

\multirow{8}{*}{$E$} &Entropy~\cite{shannon1948mathematical} & 67.85 & 75.10 & 76.80 & 77.20 & 80.95 & 83.66 & 84.81 & 85.42 & 93.26 / 82.77 (88.01)& 94.04 / 84.69 (89.36)& 94.33 / 85.31 (89.82)& 94.38 / 85.10 (89.74)\\
& TOD~\cite{huang2021semi} & 67.25&70.43 &74.84 &77.53 & 81.45 & 84.46 & 84.51 & 85.65 & 93.13 / 82.70  (87.92)& 93.63 / 84.64 (89.14)& 94.31 / 85.30 (89.81)& 94.54 / 85.82 (90.18)\\
& Gradnorm~\cite{wang2022boosting} & 68.01 & 75.75 & \blue{77.73} & 75.67& 81.21 & 83.43 & 85.30 & 85.13 & 93.36 / 83.09 \blue{(88.23)}& 93.83 / 84.91 (89.37)& 94.33 / 85.59 (89.96)& 94.65 / 85.52 (90.08)\\
& CoreSet~\cite{sener2017active} & 67.77 & 74.28 & 77.69 & 75.87 & 81.30 & 84.52 & 84.75 & \blue{86.50} & 93.24 / 82.55 (87.89)& 93.63 / 84.86 (89.24) & 94.20 / 85.50 (89.85) & 94.62 / \blue{85.89} \blue{(90.25)}\\
& BADGE~\cite{ash2020deep} & 69.12 & 75.45 & 77.37 & 76.24 & 81.31 & 84.34 & \blue{85.92} & 85.55 & \blue{93.37} / 82.95 (88.16)& 93.99 / 85.00 \blue{(89.50)} & \blue{94.50} / 85.22 (89.86) & 94.62 / 85.44 (90.03)\\
& LoGo~\cite{kim2023re} & 69.07 & \blue{ 75.76} & 74.63 & 77.24 & 82.35 & \blue{ 84.56} & 85.53 & 85.97 & 93.14 / 83.01 (88.08) & 93.93 / 84.55 (89.24)& 94.18 / \blue{85.68} (89.93) & 94.61 / 85.64 (90.12)\\
& KAFAL~\cite{cao2023knowledge} & 69.69 & 73.83 & 75.38 & 77.97 & 82.65 & 83.49 & 85.58 & 85.96 & 93.11 / 82.75 (87.93)& 94.01 / 84.12 (89.06) & 94.37 / 85.16 (89.77) & 94.46 / 85.02 (89.74)\\
& \textbf{FEAL (Ours)} & \red{72.06} & \red{76.39} & \red{78.62} & \red{80.18} & \red{82.94} & \red{85.29} & \red{86.77} & \red{87.42} & \red{93.53 / 83.72 (88.63)} & \red{94.25 / 85.19 (89.72)} & \red{94.60 / 85.96 (90.28)} & \red{94.89 / 86.27 (90.58)}\\ \toprule
\end{tabularx}
\vspace{-2.5mm}
\label{tab:seg_dice}
\end{table*}

\noindent\textbf{Image classification.}
The comparative analysis of image classification results in Fig.~\ref{fig:main_results} indicates that FEAL achieves superior results on both Fed-ISIC and Fed-Camelyon datasets. 
As depicted in Fig.~\ref{fig:main_results}, the performance of all methods exhibits a general trend of improvement with the incremental inclusion of labeled samples. 
However, an exception to this trend is observed in the Fed-ISIC dataset as shown in Fig.~\ref{fig:main_results}(a). 
As observed, the exclusive use of Entropy, Gradnorm, and CoreSet with a single model, whether it is a global (see Fig.~\ref{fig:main_results}(a)) or local model (see Fig.~\ref{fig:main_results}(b)), results in suboptimal performance, leading to a notable decrease in effectiveness beginning from the third round. The global model delivers unreliable uncertainty evaluations, which may result in suboptimal data selection and adversely affect the ability of the model to generalize effectively. Moreover, selecting data based on evaluations from the local model can cause overfitting to its specific client, negatively impacting the performance.
Conversely, methods like Gradnorm ($E$) and TOD ($E$) that combine both global and local models often outperform those relying solely on the global model, benefiting from the additional domain-specific knowledge of the local model. However, it is important to note that without proper calibration of the global model, the combined use of both models does not always guarantee better performance than solely using the local model.

\vspace{-1mm}
Remarkably, FEAL consistently outperforms state{-}of-the{-}art FAL methods on Fed-ISIC, as shown in Fig.~\ref{fig:main_results}(a)-(c). This superiority is especially noticeable in the fifth FAL round, where FEAL achieves a substantial performance gain of $1.62\%$ over the second-best method, CoreSet ($E$), as demonstrated in Fig.~\ref{fig:main_results}(c).
Additionally, it is noteworthy that FEAL achieves a performance comparable to training with the fully annotated dataset in the third round and even exceeds the fully supervised performance by $0.84\%$ in the fifth round. These advancements are primarily attributable to the effective uncertainty calibration and demonstrate the efficacy of FEAL.
It is noteworthy that the baseline methods KAFAL and LoGo, designed for FAL underperform in real-world federated scenarios. Despite showing impressive results in simulated federated datasets, they fail to replicate this success in actual multi-center federated scenarios. This is mainly due to the inherent domain shift characteristics of multi-center medical data.
As depicted in Fig.~\ref{fig:main_results}(d)-(f), FEAL also achieves superior performance on the large{-}scale dataset Fed-Camelyon, where each local client contains tens of thousands of patches. By employing a low-data regime, where merely about $3.43\%$ of the total training samples are annotated in the active learning process, FEAL attains $99.40\%$ of fully supervised performance after five rounds of FAL. This achievement represents a significant improvement compared to the second-best method, KAFAL, which reaches $98.93\%$ of the fully supervised performance, demonstrating the effectiveness of uncertainty calibration in FEAL. Additional results with different annotation budgets/ratios are available in Appendix~C.2.

\noindent\textbf{Image segmentation.}
To further evaluate the effectiveness of FEAL in segmentation tasks, we conducted experiments on three real multi-center datasets: Fed-Polyp, Fed-Prostate, and Fed-Fundus, with the results summarized in Tab.~\ref{tab:seg_dice}. As can be seen, FEAL exhibits superior performance on three multi-center segmentation datasets, as evidenced by its higher Dice scores. Specifically, for Fed-Polyp, FEAL yields a Dice score of $80.18\%$ in the fifth round, outperforming the second-best method Gradnorm ($L$) by $1.34\%$ and surpassing fully-supervised training by $2.00\%$. For Fed-Prostate, FEAL demonstrates improvements of $0.85\%$ and $0.92\%$ over the second-best method in the fourth and fifth FAL rounds, respectively. For Fed-Fundus, FEAL not only surpasses other methods in segmenting both the optic disc and optic cup but also outperforms fully supervised training in the fourth and fifth rounds of FAL. 
Complete results including HD95 and standard deviation are available in Appendix~C.2.

\subsection{Discussion}
\begin{figure}[htbp]
\centering
\includegraphics[width=\linewidth]{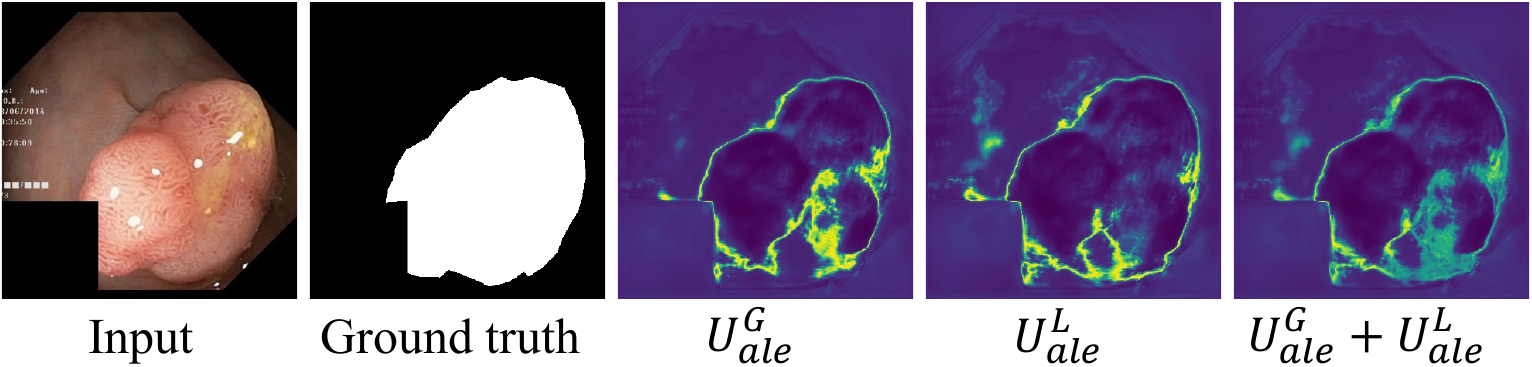}
\vspace{-5mm}
\caption{Visualization of aleatoric uncertainty on Fed-Polyp. $U_{\text{ale}}^G$ and $U_{\text{ale}}^L$ denote the aleatoric uncertainty in the global and local models, respectively.}
\vspace{-1mm}
\label{fig:udata}
\end{figure}

\noindent\textbf{Effect of uncertainty calibration.}
We conducted experiments on Fed-ISIC to evaluate the effects of different uncertainty combinations: $U_{\text{epi}}^G$, $U_{\text{ale}}^G$, and $U_{\text{ale}}^L$. As summarized in Tab.~\ref{tab:sampling_strategy}, combining aleatoric uncertainty from both global and local models proves more effective than relying on just one model. The best results are obtained with $U_{\text{epi}}^G$, $U_{\text{ale}}^G$, and $U_{\text{ale}}^L$, showcasing the effectiveness of the proposed uncertainty calibration. The ablation results on Fed-Polyp are in Appendix~C.3.
Moreover, we visualize the aleatoric uncertainty in both models on Fed-Polyp in Fig.~\ref{fig:udata}. It is noticeable that $U_{\text{ale}}^G$ and $U_{\text{ale}}^L$ highlight different regions, underscoring the importance of combining aleatoric uncertainty in both models for a more comprehensive assessment.
\vspace{-2mm}
\begin{table}[htbp]
\caption{Ablation study of uncertainty calibration on Fed-ISIC. 
}
\vspace{-2.5mm}
\footnotesize
\setlength{\tabcolsep}{3pt}
\begin{tabularx}{\linewidth}{ccc|cccc}
\bottomrule
$U_{\text{epi}}^G$ & $U_{\text{ale}}^G$ & $U_{\text{ale}}^L$ & Round 2 & Round 3 & Round 4 & Round 5 \\  
\hline

- & \checkmark & - & $60.61_{\pm1.57}$ & $66.60_{\pm0.33}$ & $67.09_{\pm1.02}$ & $66.57_{\pm1.21}$\\ 

- & - & \checkmark & $62.20_{\pm3.56}$ & $66.84_{\pm1.99}$ & $66.13_{\pm1.52}$ & $67.45_{\pm0.69}$\\    

- & \checkmark & \checkmark & $63.43_{\pm1.11}$ & $67.18_{\pm0.55}$ & $66.58_{\pm1.02}$ & $66.70_{\pm0.28}$\\

\hline

\checkmark & - & - & $61.97_{\pm1.25}$ & $65.87_{\pm0.59}$ & $67.09_{\pm1.24}$ & $66.41_{\pm1.10}$\\ 

\checkmark & \checkmark & - & $61.95_{\pm2.12}$ & $66.08_{\pm0.40}$ & $67.19_{\pm1.02}$ & $66.85_{\pm0.84}$ \\ 
\checkmark & - & \checkmark & $61.07_{\pm1.24}$ & $65.17_{\pm1.58}$ & $67.16_{\pm0.73}$ & $65.92_{\pm1.95}$\\   
\rowcolor{gray!20}\checkmark & \checkmark & \checkmark & $65.18_{\pm 0.41}$ & $67.77_{\pm1.31}$ & $68.41_{\pm1.01}$ & $68.46_{\pm0.37}$\\ 
\toprule
\end{tabularx}
\label{tab:sampling_strategy}
\vspace{-1mm}
\end{table}

\begin{figure}[htbp]    
\vspace{-1mm}
\centering
\begin{subfigure}{0.45\linewidth}
\centering
\includegraphics[width=\linewidth]{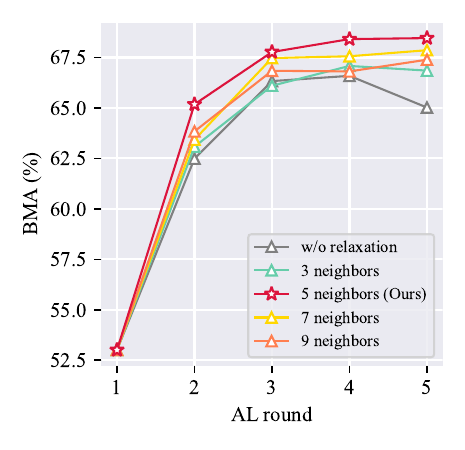}
\caption{Minimum neighbor size $n$}
\end{subfigure} 
\hspace{2mm}
\begin{subfigure}{0.45\linewidth}
\centering
\includegraphics[width=\linewidth]{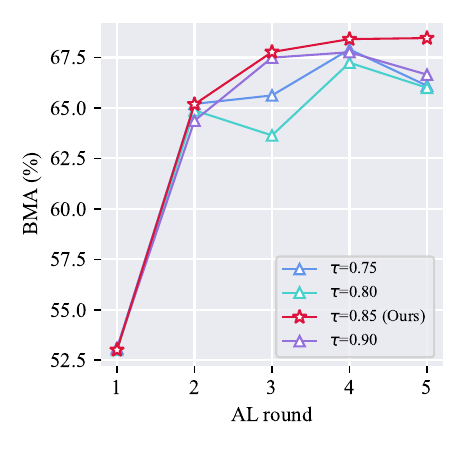}
\caption{Similarity threshold $\tau$}
\end{subfigure} 
\vspace{-2.5mm}
\caption{Ablation study of diversity relaxation on Fed-ISIC.}
\label{fig:relaxation}
\vspace{-1mm}
\end{figure}

\noindent\textbf{Effect of diversity relaxation.}
We analyzed the impact of hyperparameters, \textit{i.e.} minimum neighbor size  $n$ and similarity threshold $\tau$, on Fed-ISIC. As depicted in Fig.~\ref{fig:relaxation}(a), eliminating diversity relaxation (`w/o relaxation') results in a notable reduction in BMA in the fifth round, and the best performance is achieved with $n{=}5$ and $\tau{=}0.85$. The ablation results on Fed-Polyp are reported in Appendix~C.3.

\noindent\textbf{Effect of evidential model training.}
We performed experiments to compare the evidential loss ($\mathcal{L}$ in Eq.~\ref{eq:overall_loss}) against cross-entropy loss (CE) on Fed-ISIC and against dice loss (Dice) on Fed-Polyp. The results are detailed in Tab.~\ref{tab:loss}. As can be seen, training with evidential loss results in an average performance gain of $1.03\%$ on Fed-ISIC and $1.16\%$ on Fed-Polyp, respectively. This improvement can be primarily attributed to evidence regularization, demonstrating the efficacy of evidential model training. The ablation results on the other three datasets are available in Appendix~C.3.
\begin{table}[htbp]
\vspace{-1mm}
\caption{Ablation study of loss function.}
\vspace{-2.5mm}
\label{tab:loss}
\footnotesize
\setlength{\tabcolsep}{3pt}
\begin{tabularx}{\linewidth}{c|c|cccc}
\bottomrule
Dataset & Loss & Round 2 & Round 3 & Round 4 & Round 5 \\
\hline
\multirow{2}{*}{Fed-ISIC} & CE & $64.28_{\pm1.64}$ & $66.69_{\pm0.95}$ & $67.32_{\pm1.16}$ & $67.40_{\pm0.22}$ \\
& \cellcolor{gray!20}$\mathcal{L}$ & \cellcolor{gray!20}$65.18_{\pm 0.41}$ & \cellcolor{gray!20}$67.77_{\pm1.31}$ & \cellcolor{gray!20}$68.41_{\pm1.01}$ & \cellcolor{gray!20}$68.46_{\pm0.37}$\\
\hline
\multirow{2}{*}{Fed-Polyp} & Dice & $70.14_{\pm0.10}$ & $75.77_{\pm0.67}$ & $77.23_{\pm0.21}$ & $79.48_{\pm0.62}$ \\
& \cellcolor{gray!20}$\mathcal{L}$ & \cellcolor{gray!20}$72.06_{\pm0.72}$ & \cellcolor{gray!20}$76.39_{\pm0.66}$ & \cellcolor{gray!20}$78.62_{\pm1.44}$ & \cellcolor{gray!20}$80.18_{\pm0.10}$\\
\toprule
\end{tabularx}
\vspace{-1mm}
\end{table}

\noindent\textbf{Effect of trade{-}off weight $\lambda$.}
We further determined the optimal setting for the hyperparameter $\lambda$ on Fed-ISIC, choosing from the candidate set $\{ 1e{-}3,5e{-}3, 1e{-}2, 5e{-}2, 1e{-}1\}$, the results are detailed in Tab.~\ref{tab:hyper}. As can be seen, the best performance is achieved when $\lambda=1e{-}2$. The ablation results on Fed-Polyp are reported in Appendix~C.3.

\begin{table}[htbp]
\caption{Ablation study of trade{-}off weight $\lambda$ on Fed-ISIC.}
\vspace{-2.5mm}
\centering
\footnotesize

\begin{tabularx}{\linewidth}{c|cccc}
\bottomrule
$\lambda$ & Round 2 & Round 3 & Round 4 & Round 5 \\
 \hline
$1e{-}3$ & $63.49_{\pm3.00}$ & $64.57_{\pm2.70}$ & $66.25_{\pm1.17}$ & $65.45_{\pm1.06}$\\
$5e{-}3$ & $63.10_{\pm2.07}$ & $65.79_{\pm2.57}$ & $66.00_{\pm2.09}$ & $66.48_{\pm0.86}$\\
\rowcolor{gray!20}$1e{-}2$ & $65.18_{\pm 0.41}$ & $67.77_{\pm1.31}$ & $68.41_{\pm1.01}$ & $68.46_{\pm0.37}$\\ 
$5e{-}2$ & $62.12_{\pm0.99}$ & $67.21_{\pm1.42}$ & $66.92_{\pm0.70}$ & $66.90_{\pm0.93}$\\
$1e{-}1$ & $63.53_{\pm2.03}$ &  $66.21_{\pm0.35}$ & $66.03_{\pm2.34}$ & $67.78_{\pm1.17}$\\
\toprule
\end{tabularx}
\label{tab:hyper}
\vspace{-1mm}
\end{table}

\begin{figure}[htbp]
\vspace{-1mm}
\centering
\includegraphics[width=\linewidth]{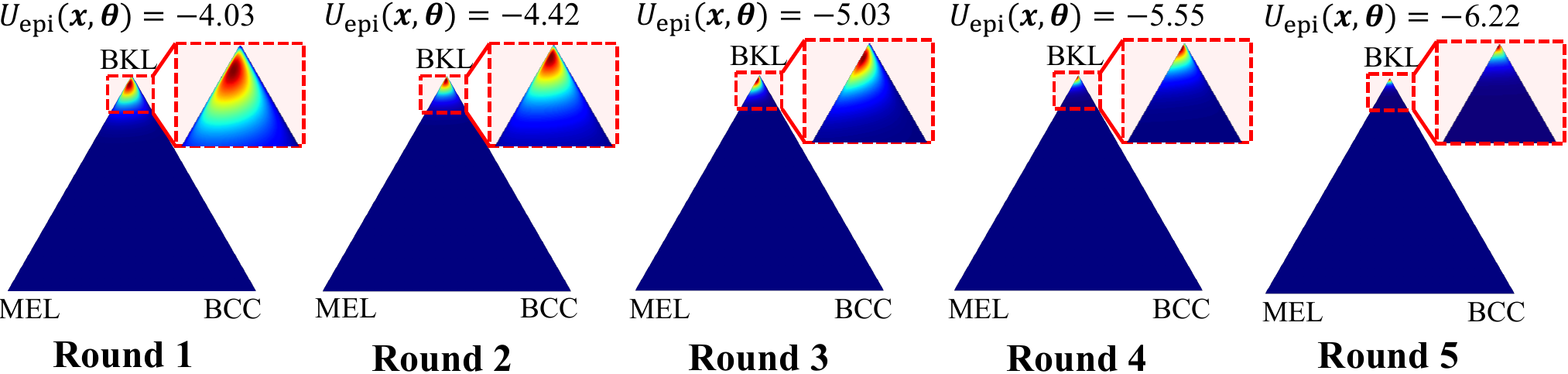}
\vspace{-5mm}
\caption{Visualization of the Dirichlet simplex for unlabeled samples across five FAL rounds using FEAL.}
\label{fig:udis_trend}
\vspace{-1mm}
\end{figure}

\begin{figure}[htbp]
\vspace{-1mm}
\centering
\includegraphics[width=\linewidth]{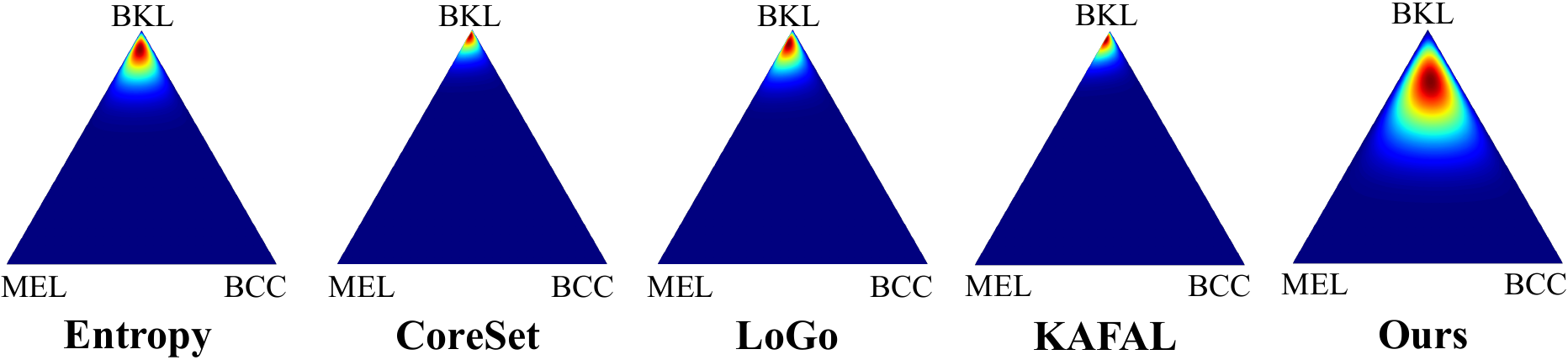}
\vspace{-5mm}
\caption{Visualization of the Dirichlet simplex for samples selected in the second FAL round using various sampling strategies.}
\label{fig:simplex}
\vspace{-1mm}
\end{figure}

\noindent\textbf{Analysis of Dirichlet simplex.}
We analyze the Dirichlet simplex on a subset of Fed-ISIC encompassing three classes. As illustrated in Fig.~\ref{fig:udis_trend}, when selecting samples with FEAL, the Dirichlet distribution becomes more concentrated at the simplex's corner for unlabeled samples, indicating reduced epistemic uncertainty in the global model. This trend verifies the effectiveness of CES in addressing domain shifts. Additionally, starting with an identical set of labeled samples, we tracked the selection of samples in the second FAL round utilizing various FAL methods. The Dirichlet simplexes of different methods are visualized in Fig.~\ref{fig:simplex}. As can be seen, the Dirichlet distribution of samples selected by FEAL showcases a broader spread across the simplex, indicating that FEAL effectively models the global model's knowledge of local data and prioritizes selecting samples characterized by high epistemic uncertainty. More details and results are available in Appendix~C.3.

\section{Conclusion and Social Impact}
\vspace{-2mm}
To address the challenge of unreliable data assessment using the global model under domain shifts, we proposed a method FEAL, which places a Dirichlet prior over categorical probabilities to treat the prediction as a distribution over the probability simplex and leverages both aleatoric uncertainty and epistemic uncertainty to calibrate the uncertainty evaluation, enhancing the reliability of data assessment and incorporating diversity relaxation to maintain sample diversity. Extensive results verify the effectiveness.
This work holds the potential to advance healthcare by preserving data privacy and facilitating collaborative research, ultimately leading to more accessible and effective patient care.

\clearpage
{
    \small
    \bibliographystyle{ieeenat_fullname}
    \bibliography{main}

\begin{thebibliography}{82}
\providecommand{\natexlab}[1]{#1}
\providecommand{\url}[1]{\texttt{#1}}
\expandafter\ifx\csname urlstyle\endcsname\relax
  \providecommand{\doi}[1]{doi: #1}\else
  \providecommand{\doi}{doi: \begingroup \urlstyle{rm}\Url}\fi

\bibitem[Ahn et~al.(2022)Ahn, Kim, Koh, and Li]{ahn2022federated}
Jin-Hyun Ahn, Kyungsang Kim, Jeongwan Koh, and Quanzheng Li.
\newblock Federated active learning (f-al): an efficient annotation strategy for federated learning.
\newblock \emph{arXiv preprint arXiv:2202.00195}, 2022.

\bibitem[Ash et~al.(2020)Ash, Zhang, Krishnamurthy, Langford, and Agarwal]{ash2020deep}
Jordan~T Ash, Chicheng Zhang, Akshay Krishnamurthy, John Langford, and Alekh Agarwal.
\newblock Deep batch active learning by diverse, uncertain gradient lower bounds.
\newblock In \emph{ICLR}, 2020.

\bibitem[Bandi et~al.(2018)Bandi, Geessink, Manson, Van~Dijk, Balkenhol, Hermsen, Bejnordi, Lee, Paeng, Zhong, et~al.]{bandi2018detection}
Peter Bandi, Oscar Geessink, Quirine Manson, Marcory Van~Dijk, Maschenka Balkenhol, Meyke Hermsen, Babak~Ehteshami Bejnordi, Byungjae Lee, Kyunghyun Paeng, Aoxiao Zhong, et~al.
\newblock From detection of individual metastases to classification of lymph node status at the patient level: the camelyon17 challenge.
\newblock \emph{IEEE Trans. Med. Imaging.}, 38\penalty0 (2):\penalty0 550--560, 2018.

\bibitem[Bernal et~al.(2015)Bernal, S{\'a}nchez, Fern{\'a}ndez-Esparrach, Gil, Rodr{\'\i}guez, and Vilari{\~n}o]{bernal2015wm}
Jorge Bernal, F~Javier S{\'a}nchez, Gloria Fern{\'a}ndez-Esparrach, Debora Gil, Cristina Rodr{\'\i}guez, and Fernando Vilari{\~n}o.
\newblock Wm-dova maps for accurate polyp highlighting in colonoscopy: Validation vs. saliency maps from physicians.
\newblock \emph{Comput. Med. Imag. Grap.}, 43:\penalty0 99--111, 2015.

\bibitem[Bloch et~al.(2015)Bloch, Madabhushi, Huisman, Freymann, Kirby, Grauer, Enquobahrie, Jaffe, Clarke, and Farahani]{bloch2015nci}
Nicholas Bloch, Anant Madabhushi, Henkjan Huisman, John Freymann, Justin Kirby, Michael Grauer, Andinet Enquobahrie, Carl Jaffe, Larry Clarke, and Keyvan Farahani.
\newblock Nci-isbi 2013 challenge: automated segmentation of prostate structures.
\newblock \emph{The Cancer Imaging Archive}, 370\penalty0 (6):\penalty0 5, 2015.

\bibitem[Cao et~al.(2023)Cao, Shi, Yu, Wang, and Tao]{cao2023knowledge}
Yu-Tong Cao, Ye Shi, Baosheng Yu, Jingya Wang, and Dacheng Tao.
\newblock Knowledge-aware federated active learning with non-iid data.
\newblock In \emph{ICCV}, pages 22279--22289, 2023.

\bibitem[Caramalau et~al.(2021)Caramalau, Bhattarai, and Kim]{caramalau2021sequential}
Razvan Caramalau, Binod Bhattarai, and Tae-Kyun Kim.
\newblock Sequential graph convolutional network for active learning.
\newblock In \emph{CVPR}, pages 9583--9592, 2021.

\bibitem[Cassidy et~al.(2022)Cassidy, Kendrick, Brodzicki, Jaworek-Korjakowska, and Yap]{cassidy2022analysis}
Bill Cassidy, Connah Kendrick, Andrzej Brodzicki, Joanna Jaworek-Korjakowska, and Moi~Hoon Yap.
\newblock Analysis of the isic image datasets: Usage, benchmarks and recommendations.
\newblock \emph{Med. Image. Anal.}, 75:\penalty0 102305, 2022.

\bibitem[Chen et~al.(2021)Chen, Zhu, Yang, and Yuan]{chen2021personalized}
Zhen Chen, Meilu Zhu, Chen Yang, and Yixuan Yuan.
\newblock Personalized retrogress-resilient framework for real-world medical federated learning.
\newblock In \emph{MICCAI}, pages 347--356, 2021.

\bibitem[Deng et~al.(2009)Deng, Dong, Socher, Li, Li, and Fei-Fei]{deng2009imagenet}
Jia Deng, Wei Dong, Richard Socher, Li-Jia Li, Kai Li, and Li Fei-Fei.
\newblock Imagenet: A large-scale hierarchical image database.
\newblock In \emph{CVPR}, pages 248--255. IEEE, 2009.

\bibitem[Elhamifar et~al.(2013)Elhamifar, Sapiro, Yang, and Sasrty]{elhamifar2013convex}
Ehsan Elhamifar, Guillermo Sapiro, Allen Yang, and S~Shankar Sasrty.
\newblock A convex optimization framework for active learning.
\newblock In \emph{ICCV}, pages 209--216, 2013.

\bibitem[Fumero et~al.(2011)Fumero, Alay{\'o}n, Sanchez, Sigut, and Gonzalez-Hernandez]{fumero2011rim}
Francisco Fumero, Silvia Alay{\'o}n, Jos{\'e}~L Sanchez, Jose Sigut, and M Gonzalez-Hernandez.
\newblock Rim-one: An open retinal image database for optic nerve evaluation.
\newblock In \emph{CBMS}, pages 1--6. IEEE, 2011.

\bibitem[Gal and Ghahramani(2016)]{gal2016dropout}
Yarin Gal and Zoubin Ghahramani.
\newblock Dropout as a bayesian approximation: representing model uncertainty in deep learning.
\newblock In \emph{ICML}, 2016.

\bibitem[Gal et~al.(2017)Gal, Islam, and Ghahramani]{gal2017deep}
Yarin Gal, Riashat Islam, and Zoubin Ghahramani.
\newblock Deep bayesian active learning with image data.
\newblock In \emph{ICML}, 2017.

\bibitem[Gao et~al.(2020)Gao, Zhang, Yu, Ar{\i}k, Davis, and Pfister]{gao2020consistency}
Mingfei Gao, Zizhao Zhang, Guo Yu, Sercan~{\"O} Ar{\i}k, Larry~S Davis, and Tomas Pfister.
\newblock Consistency-based semi-supervised active learning: Towards minimizing labeling cost.
\newblock In \emph{ECCV}, pages 510--526, 2020.

\bibitem[Guo et~al.(2023)Guo, Tang, and Lin]{guo2023fedbr}
Yongxin Guo, Xiaoying Tang, and Tao Lin.
\newblock Fedbr: improving federated learning on heterogeneous data via local learning bias reduction.
\newblock In \emph{ICML}, pages 12034--12054, 2023.

\bibitem[Huang et~al.(2017)Huang, Liu, Van Der~Maaten, and Weinberger]{huang2017densely}
Gao Huang, Zhuang Liu, Laurens Van Der~Maaten, and Kilian~Q Weinberger.
\newblock Densely connected convolutional networks.
\newblock In \emph{CVPR}, pages 4700--4708, 2017.

\bibitem[Huang et~al.(2021)Huang, Wang, Xiong, Huan, and Dou]{huang2021semi}
Siyu Huang, Tianyang Wang, Haoyi Xiong, Jun Huan, and Dejing Dou.
\newblock Semi-supervised active learning with temporal output discrepancy.
\newblock In \emph{ICCV}, pages 3447--3456, 2021.

\bibitem[Huang et~al.(2023)Huang, Ye, Shi, Li, and Du]{huang2023rethinking}
Wenke Huang, Mang Ye, Zekun Shi, He Li, and Bo Du.
\newblock Rethinking federated learning with domain shift: A prototype view.
\newblock In \emph{CVPR}, pages 16312--16322. IEEE, 2023.

\bibitem[Jha et~al.(2020)Jha, Smedsrud, Riegler, Halvorsen, de~Lange, Johansen, and Johansen]{jha2020kvasir}
Debesh Jha, Pia~H Smedsrud, Michael~A Riegler, P{\aa}l Halvorsen, Thomas de Lange, Dag Johansen, and H{\aa}vard~D Johansen.
\newblock Kvasir-seg: A segmented polyp dataset.
\newblock In \emph{MMM}, pages 451--462, 2020.

\bibitem[Jiang et~al.(2022)Jiang, Wang, and Dou]{jiang2022harmofl}
Meirui Jiang, Zirui Wang, and Qi Dou.
\newblock Harmofl: Harmonizing local and global drifts in federated learning on heterogeneous medical images.
\newblock In \emph{AAAI}, pages 1087--1095, 2022.

\bibitem[Jiang et~al.(2023)Jiang, Roth, Li, Yang, Zhao, Nath, Xu, Dou, and Xu]{jiang2023fair}
Meirui Jiang, Holger~R Roth, Wenqi Li, Dong Yang, Can Zhao, Vishwesh Nath, Daguang Xu, Qi Dou, and Ziyue Xu.
\newblock Fair federated medical image segmentation via client contribution estimation.
\newblock In \emph{CVPR}, pages 16302--16311, 2023.

\bibitem[J{\o}sang(2016)]{josang2016subjective}
Audun J{\o}sang.
\newblock \emph{Subjective logic}.
\newblock 2016.

\bibitem[Karimireddy et~al.(2020)Karimireddy, Kale, Mohri, Reddi, Stich, and Suresh]{karimireddy2020scaffold}
Sai~Praneeth Karimireddy, Satyen Kale, Mehryar Mohri, Sashank Reddi, Sebastian Stich, and Ananda~Theertha Suresh.
\newblock Scaffold: Stochastic controlled averaging for federated learning.
\newblock In \emph{ICML}, pages 5132--5143, 2020.

\bibitem[Kim et~al.(2023)Kim, Bae, Song, and Yun]{kim2023re}
SangMook Kim, Sangmin Bae, Hwanjun Song, and Se-Young Yun.
\newblock Re-thinking federated active learning based on inter-class diversity.
\newblock In \emph{CVPR}, pages 3944--3953, 2023.

\bibitem[Kingma and Ba(2014)]{kingma2014adam}
Diederik~P Kingma and Jimmy Ba.
\newblock Adam: A method for stochastic optimization.
\newblock \emph{arXiv preprint arXiv:1412.6980}, 2014.

\bibitem[Kone{\v{c}}n{\`y} et~al.(2016)Kone{\v{c}}n{\`y}, McMahan, Ramage, and Richt{\'a}rik]{konevcny2016federated}
Jakub Kone{\v{c}}n{\`y}, H~Brendan McMahan, Daniel Ramage, and Peter Richt{\'a}rik.
\newblock Federated optimization: Distributed machine learning for on-device intelligence.
\newblock \emph{arXiv preprint arXiv:1610.02527}, 2016.

\bibitem[Kullback and Leibler(1951)]{kullback1951information}
Solomon Kullback and Richard~A Leibler.
\newblock On information and sufficiency.
\newblock \emph{The annals of mathematical statistics}, 22\penalty0 (1):\penalty0 79--86, 1951.

\bibitem[Kutsuna et~al.(2012)Kutsuna, Higaki, Matsunaga, Otsuki, Yamaguchi, Fujii, and Hasezawa]{kutsuna2012active}
Natsumaro Kutsuna, Takumi Higaki, Sachihiro Matsunaga, Tomoshi Otsuki, Masayuki Yamaguchi, Hirofumi Fujii, and Seiichiro Hasezawa.
\newblock Active learning framework with iterative clustering for bioimage classification.
\newblock \emph{Nat. Commun.}, 3\penalty0 (1):\penalty0 1032, 2012.

\bibitem[Lema{\^\i}tre et~al.(2015)Lema{\^\i}tre, Mart{\'\i}, Freixenet, Vilanova, Walker, and Meriaudeau]{lemaitre2015computer}
Guillaume Lema{\^\i}tre, Robert Mart{\'\i}, Jordi Freixenet, Joan~C Vilanova, Paul~M Walker, and Fabrice Meriaudeau.
\newblock Computer-aided detection and diagnosis for prostate cancer based on mono and multi-parametric mri: a review.
\newblock \emph{Comput. Biol. Med.}, 60:\penalty0 8--31, 2015.

\bibitem[Li and Yin(2020)]{li2020attention}
Haohan Li and Zhaozheng Yin.
\newblock Attention, suggestion and annotation: a deep active learning framework for biomedical image segmentation.
\newblock In \emph{MICCAI}, pages 3--13, 2020.

\bibitem[Li et~al.(2022)Li, Nan, Del~Ser, and Yang]{li2022region}
Hao Li, Yang Nan, Javier Del~Ser, and Guang Yang.
\newblock Region-based evidential deep learning to quantify uncertainty and improve robustness of brain tumor segmentation.
\newblock \emph{Neural Comput. Appl.}, pages 1--15, 2022.

\bibitem[Li et~al.(2024)Li, Huang, Xu, Yang, Zhang, Ji, Xie, Yuan, Liu, and Chen]{li2024octa}
Mingchao Li, Kun Huang, Qiuzhuo Xu, Jiadong Yang, Yuhan Zhang, Zexuan Ji, Keren Xie, Songtao Yuan, Qinghuai Liu, and Qiang Chen.
\newblock Octa-500: a retinal dataset for optical coherence tomography angiography study.
\newblock \emph{Med. Image. Anal.}, 93:\penalty0 103092, 2024.

\bibitem[Li et~al.(2021{\natexlab{a}})Li, He, and Song]{li2021model}
Qinbin Li, Bingsheng He, and Dawn Song.
\newblock Model-contrastive federated learning.
\newblock In \emph{CVPR}, pages 10713--10722, 2021{\natexlab{a}}.

\bibitem[Li et~al.(2020)Li, Sahu, Zaheer, Sanjabi, Talwalkar, and Smith]{li2020federated}
Tian Li, Anit~Kumar Sahu, Manzil Zaheer, Maziar Sanjabi, Ameet Talwalkar, and Virginia Smith.
\newblock Federated optimization in heterogeneous networks.
\newblock \emph{MLSys}, 2:\penalty0 429--450, 2020.

\bibitem[Li et~al.(2021{\natexlab{b}})Li, Jiang, Zhang, Kamp, and Dou]{li2021fedbn}
Xiaoxiao Li, Meirui Jiang, Xiaofei Zhang, Michael Kamp, and Qi Dou.
\newblock Fedbn: Federated learning on non-iid features via local batch normalization.
\newblock \emph{arXiv preprint arXiv:2102.07623}, 2021{\natexlab{b}}.

\bibitem[Li et~al.(2023)Li, Lin, Shang, and Wu]{li2023revisiting}
Zexi Li, Tao Lin, Xinyi Shang, and Chao Wu.
\newblock Revisiting weighted aggregation in federated learning with neural networks.
\newblock \emph{arXiv preprint arXiv:2302.10911}, 2023.

\bibitem[Litjens et~al.(2014)Litjens, Toth, Van De~Ven, Hoeks, Kerkstra, Van~Ginneken, Vincent, Guillard, Birbeck, Zhang, et~al.]{litjens2014evaluation}
Geert Litjens, Robert Toth, Wendy Van De~Ven, Caroline Hoeks, Sjoerd Kerkstra, Bram Van~Ginneken, Graham Vincent, Gwenael Guillard, Neil Birbeck, Jindang Zhang, et~al.
\newblock Evaluation of prostate segmentation algorithms for mri: the promise12 challenge.
\newblock \emph{Med. Image. Anal.}, 18\penalty0 (2):\penalty0 359--373, 2014.

\bibitem[Liu et~al.(2021)Liu, Chen, Qin, Dou, and Heng]{liu2021feddg}
Quande Liu, Cheng Chen, Jing Qin, Qi Dou, and Pheng-Ann Heng.
\newblock Feddg: Federated domain generalization on medical image segmentation via episodic learning in continuous frequency space.
\newblock In \emph{CVPR}, pages 1013--1023, 2021.

\bibitem[Liu et~al.(2020)Liu, Wang, Owens, and Li]{liu2020energy}
Weitang Liu, Xiaoyun Wang, John Owens, and Yixuan Li.
\newblock Energy-based out-of-distribution detection.
\newblock \emph{NeurIPS}, 33:\penalty0 21464--21475, 2020.

\bibitem[Ma et~al.(2023)Ma, Feng, Chen, Li, and Xia]{ma2023federated}
Benteng Ma, Yu Feng, Geng Chen, Changyang Li, and Yong Xia.
\newblock Federated adaptive reweighting for medical image classification.
\newblock \emph{Pattern Recogn.}, 144:\penalty0 109880, 2023.

\bibitem[Ma et~al.(2024)Ma, Zhang, Xia, and Tao]{ma2024VNAS}
Benteng Ma, Jing Zhang, Yong Xia, and Dacheng Tao.
\newblock Vnas: Variational neural architecture search.
\newblock \emph{Int. J. Comput. Vis.}, 2024.

\bibitem[MacQueen et~al.(1967)]{macqueen1967some}
James MacQueen et~al.
\newblock Some methods for classification and analysis of multivariate observations.
\newblock In \emph{Proceedings of the fifth Berkeley symposium on mathematical statistics and probability}, pages 281--297, 1967.

\bibitem[Malinin and Gales(2018)]{malinin2018predictive}
Andrey Malinin and Mark Gales.
\newblock Predictive uncertainty estimation via prior networks.
\newblock \emph{NeurIPS}, 31, 2018.

\bibitem[McMahan et~al.(2017)McMahan, Moore, Ramage, Hampson, and y~Arcas]{mcmahan2017communication}
Brendan McMahan, Eider Moore, Daniel Ramage, Seth Hampson, and Blaise~Aguera y Arcas.
\newblock Communication-efficient learning of deep networks from decentralized data.
\newblock In \emph{AISTATS}, pages 1273--1282, 2017.

\bibitem[Monarch(2021)]{monarch2021human}
Robert~Munro Monarch.
\newblock \emph{Human-in-the-Loop Machine Learning: Active learning and annotation for human-centered AI}.
\newblock Simon and Schuster, 2021.

\bibitem[Ng et~al.(2011)Ng, Tian, and Tang]{ng2011dirichlet}
Kai~Wang Ng, Guo-Liang Tian, and Man-Lai Tang.
\newblock Dirichlet and related distributions: Theory, methods and applications.
\newblock 2011.

\bibitem[Nguyen and Smeulders(2004)]{nguyen2004active}
Hieu~T Nguyen and Arnold Smeulders.
\newblock Active learning using pre-clustering.
\newblock In \emph{ICML}, page~79, 2004.

\bibitem[Ogier~du Terrail et~al.(2022)Ogier~du Terrail, Ayed, Cyffers, Grimberg, He, Loeb, Mangold, Marchand, Marfoq, Mushtaq, et~al.]{ogier2022flamby}
Jean Ogier~du Terrail, Samy-Safwan Ayed, Edwige Cyffers, Felix Grimberg, Chaoyang He, Regis Loeb, Paul Mangold, Tanguy Marchand, Othmane Marfoq, Erum Mushtaq, et~al.
\newblock Flamby: Datasets and benchmarks for cross-silo federated learning in realistic healthcare settings.
\newblock \emph{NeurIPS}, 35:\penalty0 5315--5334, 2022.

\bibitem[Orlando et~al.(2020)Orlando, Fu, Breda, Van~Keer, Bathula, Diaz-Pinto, Fang, Heng, Kim, Lee, et~al.]{orlando2020refuge}
Jos{\'e}~Ignacio Orlando, Huazhu Fu, Jo{\~a}o~Barbosa Breda, Karel Van~Keer, Deepti~R Bathula, Andr{\'e}s Diaz-Pinto, Ruogu Fang, Pheng-Ann Heng, Jeyoung Kim, JoonHo Lee, et~al.
\newblock Refuge challenge: A unified framework for evaluating automated methods for glaucoma assessment from fundus photographs.
\newblock \emph{Med. Image. Anal.}, 59:\penalty0 101570, 2020.

\bibitem[Pandey and Yu(2023)]{pandey2023learn}
Deep~Shankar Pandey and Qi Yu.
\newblock Learn to accumulate evidence from all training samples: Theory and practice.
\newblock In \emph{ICML}, pages 26963--26989, 2023.

\bibitem[Parvaneh et~al.(2022)Parvaneh, Abbasnejad, Teney, Haffari, van~den Hengel, and Shi]{parvaneh2022active}
Amin Parvaneh, Ehsan Abbasnejad, Damien Teney, Gholamreza~Reza Haffari, Anton van~den Hengel, and Javen~Qinfeng Shi.
\newblock Active learning by feature mixing.
\newblock In \emph{CVPR}, pages 12237--12246, 2022.

\bibitem[Pearce(2020)]{pearce2020uncertainty}
Tim Pearce.
\newblock \emph{Uncertainty in neural networks; bayesian ensembles, priors \& prediction intervals}.
\newblock PhD thesis, University of Cambridge, 2020.

\bibitem[Pearce et~al.(2021)Pearce, Brintrup, and Zhu]{pearce2021understanding}
Tim Pearce, Alexandra Brintrup, and Jun Zhu.
\newblock Understanding softmax confidence and uncertainty.
\newblock \emph{arXiv preprint arXiv:2106.04972}, 2021.

\bibitem[Rehman et~al.(2023)Rehman, Gao, De~Gusm{\~a}o, Alibeigi, Shen, and Lane]{rehman2023dawa}
Yasar Abbas~Ur Rehman, Yan Gao, Pedro Porto~Buarque De~Gusm{\~a}o, Mina Alibeigi, Jiajun Shen, and Nicholas~D Lane.
\newblock L-dawa: Layer-wise divergence aware weight aggregation in federated self-supervised visual representation learning.
\newblock In \emph{ICCV}, pages 16464--16473, 2023.

\bibitem[Ronneberger et~al.(2015)Ronneberger, Fischer, and Brox]{ronneberger2015u}
Olaf Ronneberger, Philipp Fischer, and Thomas Brox.
\newblock U-net: Convolutional networks for biomedical image segmentation.
\newblock In \emph{MICCAI}, pages 234--241, 2015.

\bibitem[Sener and Savarese(2017)]{sener2017active}
Ozan Sener and Silvio Savarese.
\newblock Active learning for convolutional neural networks: A core-set approach.
\newblock \emph{arXiv preprint arXiv:1708.00489}, 2017.

\bibitem[Sensoy et~al.(2018)Sensoy, Kaplan, and Kandemir]{sensoy2018evidential}
Murat Sensoy, Lance Kaplan, and Melih Kandemir.
\newblock Evidential deep learning to quantify classification uncertainty.
\newblock \emph{NeurIPS}, 31, 2018.

\bibitem[Sentz and Ferson(2002)]{sentz2002combination}
Kari Sentz and Scott Ferson.
\newblock Combination of evidence in dempster-shafer theory.
\newblock 2002.

\bibitem[Settles(2009)]{settles2009active}
Burr Settles.
\newblock Active learning literature survey.
\newblock 2009.

\bibitem[Shannon(1948)]{shannon1948mathematical}
Claude~Elwood Shannon.
\newblock A mathematical theory of communication.
\newblock \emph{Bell Syst. Tech. J.}, 27\penalty0 (3):\penalty0 379--423, 1948.

\bibitem[Shen et~al.(2023)Shen, Bu, Sattigeri, Ghosh, Das, and Wornell]{shen2023post}
Maohao Shen, Yuheng Bu, Prasanna Sattigeri, Soumya Ghosh, Subhro Das, and Gregory Wornell.
\newblock Post-hoc uncertainty learning using a dirichlet meta-model.
\newblock In \emph{AAAI}, pages 9772--9781, 2023.

\bibitem[Shoham et~al.(2019)Shoham, Avidor, Keren, Israel, Benditkis, Mor-Yosef, and Zeitak]{shoham2019overcoming}
Neta Shoham, Tomer Avidor, Aviv Keren, Nadav Israel, Daniel Benditkis, Liron Mor-Yosef, and Itai Zeitak.
\newblock Overcoming forgetting in federated learning on non-iid data.
\newblock \emph{arXiv preprint arXiv:1910.07796}, 2019.

\bibitem[Silva et~al.(2014)Silva, Histace, Romain, Dray, and Granado]{silva2014toward}
Juan Silva, Aymeric Histace, Olivier Romain, Xavier Dray, and Bertrand Granado.
\newblock Toward embedded detection of polyps in wce images for early diagnosis of colorectal cancer.
\newblock \emph{Int. J. Comput. Assist. Radiol. Surg.}, 9:\penalty0 283--293, 2014.

\bibitem[Sivaswamy et~al.(2015)Sivaswamy, Krishnadas, Chakravarty, Joshi, Tabish, et~al.]{sivaswamy2015comprehensive}
Jayanthi Sivaswamy, S Krishnadas, Arunava Chakravarty, G Joshi, A~Syed Tabish, et~al.
\newblock A comprehensive retinal image dataset for the assessment of glaucoma from the optic nerve head analysis.
\newblock \emph{JSM Biomedical Imaging Data Papers}, 2\penalty0 (1):\penalty0 1004, 2015.

\bibitem[Tajbakhsh et~al.(2015)Tajbakhsh, Gurudu, and Liang]{tajbakhsh2015automated}
Nima Tajbakhsh, Suryakanth~R Gurudu, and Jianming Liang.
\newblock Automated polyp detection in colonoscopy videos using shape and context information.
\newblock \emph{IEEE Trans. Med. Imaging.}, 35\penalty0 (2):\penalty0 630--644, 2015.

\bibitem[Tan and Le(2019)]{tan2019efficientnet}
Mingxing Tan and Quoc Le.
\newblock Efficientnet: Rethinking model scaling for convolutional neural networks.
\newblock In \emph{ICML}, pages 6105--6114, 2019.

\bibitem[Ulmer et~al.(2023)Ulmer, Hardmeier, and Frellsen]{ulmer2023prior}
Dennis Ulmer, Christian Hardmeier, and Jes Frellsen.
\newblock Prior and posterior networks: A survey on evidential deep learning methods for uncertainty estimation.
\newblock \emph{TMLR}, 2023.

\bibitem[Urner et~al.(2013)Urner, Wulff, and Ben-David]{urner2013plal}
Ruth Urner, Sharon Wulff, and Shai Ben-David.
\newblock Plal: Cluster-based active learning.
\newblock In \emph{COLT}, pages 376--397, 2013.

\bibitem[Wang et~al.(2023)Wang, Han, Zhang, He, and Yin]{wang2023mhpl}
Fan Wang, Zhongyi Han, Zhiyan Zhang, Rundong He, and Yilong Yin.
\newblock Mhpl: Minimum happy points learning for active source free domain adaptation.
\newblock In \emph{CVPR}, pages 20008--20018, 2023.

\bibitem[Wang et~al.(2022{\natexlab{a}})Wang, Jin, and Wang]{wang2022personalizing}
Jiacheng Wang, Yueming Jin, and Liansheng Wang.
\newblock Personalizing federated medical image segmentation via local calibration.
\newblock In \emph{ECCV}, pages 456--472, 2022{\natexlab{a}}.

\bibitem[Wang et~al.(2022{\natexlab{b}})Wang, Li, Yang, Hu, Zeng, Huang, Xu, and Xu]{wang2022boosting}
Tianyang Wang, Xingjian Li, Pengkun Yang, Guosheng Hu, Xiangrui Zeng, Siyu Huang, Cheng-Zhong Xu, and Min Xu.
\newblock Boosting active learning via improving test performance.
\newblock In \emph{AAAI}, pages 8566--8574, 2022{\natexlab{b}}.

\bibitem[Wicaksana et~al.(2023)Wicaksana, Yan, and Cheng]{wicaksana2023fca}
Jeffry Wicaksana, Zengqiang Yan, and Kwang-Ting Cheng.
\newblock Fca: Taming long-tailed federated medical image classification by classifier anchoring.
\newblock \emph{arXiv preprint arXiv:2305.00738}, 2023.

\bibitem[Wu et~al.(2023)Wu, Yu, Yang, Cheng, and Yan]{wu2023fediic}
Nannan Wu, Li Yu, Xin Yang, Kwang-Ting Cheng, and Zengqiang Yan.
\newblock Fediic: Towards robust federated learning for class-imbalanced medical image classification.
\newblock In \emph{MICCAI}, pages 692--702, 2023.

\bibitem[Wu et~al.(2022)Wu, Pei, Chen, Zhu, Wang, Qian, Zhang, Sun, and Guo]{wu2022federated}
Xing Wu, Jie Pei, Cheng Chen, Yimin Zhu, Jianjia Wang, Quan Qian, Jian Zhang, Qun Sun, and Yike Guo.
\newblock Federated active learning for multicenter collaborative disease diagnosis.
\newblock \emph{IEEE Trans. Med. Imaging.}, 2022.

\bibitem[Xie et~al.(2023)Xie, Li, Zhang, and Liu]{xie2023dirichlet}
Mixue Xie, Shuang Li, Rui Zhang, and Chi~Harold Liu.
\newblock Dirichlet-based uncertainty calibration for active domain adaptation.
\newblock \emph{arXiv preprint arXiv:2302.13824}, 2023.

\bibitem[Yang et~al.(2015)Yang, Ma, Nie, Chang, and Hauptmann]{yang2015multi}
Yi Yang, Zhigang Ma, Feiping Nie, Xiaojun Chang, and Alexander~G Hauptmann.
\newblock Multi-class active learning by uncertainty sampling with diversity maximization.
\newblock \emph{Int. J. Comput. Vis.}, 113:\penalty0 113--127, 2015.

\bibitem[Ye et~al.(2023)Ye, Fang, Du, Yuen, and Tao]{ye2023heterogeneous}
Mang Ye, Xiuwen Fang, Bo Du, Pong~C Yuen, and Dacheng Tao.
\newblock Heterogeneous federated learning: State-of-the-art and research challenges.
\newblock \emph{ACM Comput. Surv.}, 56\penalty0 (3):\penalty0 1--44, 2023.

\bibitem[Yoo and Kweon(2019)]{yoo2019learning}
Donggeun Yoo and In~So Kweon.
\newblock Learning loss for active learning.
\newblock In \emph{CVPR}, 2019.

\bibitem[Yuan et~al.(2023)Yuan, Zhang, Yan, Chen, Shi, Li, and Qiao]{yuan2023bi3d}
Jiakang Yuan, Bo Zhang, Xiangchao Yan, Tao Chen, Botian Shi, Yikang Li, and Yu Qiao.
\newblock Bi3d: Bi-domain active learning for cross-domain 3d object detection.
\newblock In \emph{CVPR}, pages 15599--15608, 2023.

\bibitem[Zhang et~al.(2023)Zhang, Xu, Yao, Zhang, Tian, and Wang]{zhang2023federated}
Ruipeng Zhang, Qinwei Xu, Jiangchao Yao, Ya Zhang, Qi Tian, and Yanfeng Wang.
\newblock Federated domain generalization with generalization adjustment.
\newblock In \emph{CVPR}, pages 3954--3963, 2023.

\bibitem[Zhou and Konukoglu(2023)]{zhou2023fedfa}
Tianfei Zhou and Ender Konukoglu.
\newblock Fedfa: Federated feature augmentation.
\newblock \emph{arXiv preprint arXiv:2301.12995}, 2023.

\end{thebibliography}
}
\clearpage
\setcounter{page}{1}
\setcounter{section}{0}
\setcounter{algorithm}{1}
\setcounter{equation}{9}
\setcounter{table}{4}
\setcounter{figure}{7}

\renewcommand\thesection{\Alph{section}}
\maketitlesupplementary
In the supplementary material, we provide the workflow of FEAL in Sec.~\ref{sec:workflow}, formula derivations in Sec.~\ref{sec:deri}, detailed descriptions of experimental settings, results, and discussions in Sec.~\ref{sec:exp}, additional experiments conducted on OCTA datasets in Sec.~\ref{sec:octa}, and discussions comparing with other methods in Sec.~\ref{sec:discuss}.

\section{Workflow of FEAL}
\label{sec:workflow}
The detailed workflow of FEAL is summarized in Alg.~\ref{alg:feal}.
\begin{algorithm}[H]
\small
\caption{Workflow of FEAL}
\label{alg:feal}
\begin{algorithmic}[1]
\REQUIRE global model $\btheta$, local models $\{\btheta_k\}_{k=1}^K$, unlabeled sets $\{U_k\}_{k=1}^K$, annotation budget $\{B_k\}_{k=1}^K$, active learning rounds $R$, aggregation weights $\{\alpha_k\}_{k=1}^K$, communication rounds $T$
\ENSURE trained global model $\btheta^*$
\,\\
\blue{/* 1st AL round */}
\FOR{$k=1$ \TO $K$} 
\STATE Randomly annotate $B_k$ samples from the unlabeled pool $U_k$ to construct the initial labeled pool $L_k^1=\{(\bx_i,\boldsymbol{y}_i\}_{i=1}^{B_k}$.
\vspace{-3mm}
\STATE Update the unlabeled pool $U_k^1=U_k\setminus L_k^1$.
\STATE $\{\alpha_k\}_{k=1}^K\leftarrow\frac{\{|L_k^1|\}_{k=1}^K}{\sum_{k=1}^K |L_k^1|}$
\ENDFOR
\STATE$\btheta^*\leftarrow\btheta$
\FOR{$t=1$ \TO $T$}
\FOR{$k=1$ \TO $K$}
\STATE $\btheta_k^1\leftarrow\btheta^*$
\STATE $\btheta_k^1\leftarrow\text{LocalTraining}(\btheta_k^1, L_k^1)$
\ENDFOR
\STATE $\btheta^{*}\leftarrow\text{FedAvg}(\{\btheta_k^1\}_{k=1}^K, \{\alpha_k\}_{k=1}^K)$
\ENDFOR

\blue{/* $2\text{nd}\sim R$-th AL round */}
\FOR{$r=2$ \TO $R$}
\FOR{$k=1$ \TO $K$}
\STATE $Q_k^r=\text{FEAL}(\btheta^*, \btheta_k^{r-1}, U_k^{r-1})$\blue{\COMMENT{Local data annotation}}
\STATE $L_k^r=L_k^{r-1}\cup Q_k^r$
\STATE$U_k^r=U_k^{r-1}\setminus Q_k^r$
\STATE $\{\alpha_k\}_{k=1}^K\leftarrow\frac{\{|L_k^r|\}_{k=1}^K}{\sum_{k=1}^K |L_k^r|}$
\ENDFOR
\FOR{$t=1$ \TO $T$}
\FOR{$k=1$ \TO $K$}
\STATE $\btheta_k^r\leftarrow\btheta^*$\blue{\COMMENT{Model distribution}}
\STATE $\btheta_k^r\leftarrow\text{LocalTraining}(\btheta_k^r, L_k^r)$\blue{\COMMENT{Local training}}
\ENDFOR
\STATE $\btheta^*\leftarrow \text{FedAvg}(\{\btheta_k^{r}\}_{k=1}^K, \{\alpha_k\}_{k=1}^K\})\text{\blue{\COMMENT{Model aggregation}}}$
\ENDFOR
\ENDFOR
\RETURN $\btheta^*$
\end{algorithmic}
\end{algorithm}

The sampling strategy employed by FEAL and the aggregation method utilized in FedAvg are detailed in Alg.~\ref{alg:feal_ss} and Alg.~\ref{alg:fedavg}, respectively.
\begin{algorithm}[H]
\small
\caption{Sampling strategy of FEAL}
\label{alg:feal_ss}
\begin{algorithmic}[1]
\REQUIRE global model $\btheta^*$, local model $\btheta_k^{r-1}$, unlabeled set $U_k^{r-1}$, annotation budget $B_k$
\ENSURE query set $Q_k^r$
\,\\
\blue{/* Uncertainty calibration */}
\FORALL{$\bx\in U_k^{r-1}$}
\STATE Compute $U_{\text{ale}}(\bx,\btheta_k^{r-1})$ and $U_{\text{ale}}(\bx,\btheta^*)$ using Eq.~3.
\STATE Compute $U_{\text{epi}}(\bx,\btheta^*)$ using Eq.~4.
\STATE Compute $U(\bx,\btheta^*, \btheta_k^{r-1})$ by Eq.~5.
\ENDFOR
\,\\
\blue{/* Diversity relaxation */}
\STATE Determine the query set $Q_k^{r}$ according to Alg.~1.
\RETURN $Q_k^r$
\end{algorithmic}
\end{algorithm}

\begin{algorithm}[H]
\small
\caption{Aggregation method of FedAvg}
\label{alg:fedavg}
\begin{algorithmic}[1]
\REQUIRE local models $\{\btheta_k^r\}_{k=1}^K$, aggregation weights $\{\alpha_k\}_{k=1}^K$
\ENSURE global model $\btheta^*$
\STATE $\btheta^*\leftarrow\sum_{k=1}^K\alpha_k\cdot\btheta_k^r$
\RETURN $\btheta^*$
\end{algorithmic}
\end{algorithm}

\section{Derivations}
\label{sec:deri}
\subsection{Dirichlet-based Evidential Model in FAL}
\label{subsec:deri_1}
In Dirichlet-based evidential models, given a sample $\bx$ and a model $\btheta$, the categorical prediction $\brho$ follows a Dirichlet distribution, denoted as $p(\brho|\bx,\btheta)\sim Dir(\brho|\boldsymbol{\alpha})$. The probability density function of $\brho$~\cite{sensoy2018evidential,xie2023dirichlet}, conditioned on the sample $\bx$ and the model $\btheta$, is formulated as:
\begin{equation}
\small
\begin{aligned} 
p(\brho|\bx,\btheta)
    = \left\{
    \begin{aligned}
         \frac{\Gamma(\sum_{c=1}^C\alpha_c)}{\prod_{c=1}^C\Gamma(\alpha_c)}\prod_{c=1}^C\brho_c^{\alpha_c-1}&, (\brho\in \Delta^C) \\
         0\qquad\qquad&, (\text{otherwise}) 
    \end{aligned}
    \right.
\end{aligned}
\label{eq:rho}
\end{equation}
where $\boldsymbol{\alpha}=\{\alpha_1,\alpha_2,\cdots,\alpha_C\}$ denotes the parameters of the Dirichlet distribution for sample $\bx$, $\Gamma(\cdot)$ is the Gamma function, and $\Delta^C=\{\brho|\sum_{c=1}^C\rho_c=1\ \text{and}\ 0<\rho_c<1\}$ is the $C$-dimensional unit simplex.

As stated in~\cite{ng2011dirichlet}, the marginal distributions of the Dirichlet distribution follow Beta distributions. Consequently, given $p(\brho|\bx,\btheta)\sim Dir(\brho|\boldsymbol{\alpha})$, we can express $p(\rho_c|\bx,\btheta)\sim Beta(\rho_c|\alpha_c,S-\alpha_c)$, where $S=\sum_{c=1}^C\alpha_c$ represents the Dirichlet strength. The probability density function of $\rho_c$, given the sample $\bx$ and the model $\btheta$, is formulated as:
\begin{equation}
\small
    p(\rho_c|\bx,\btheta)=\frac{1}{\mathcal{B}(\alpha_c,S-\alpha_c)}\rho_c^{\alpha_c-1}(1-\rho_c)^{S-\alpha_c-1},
\label{eq:rho_c}
\end{equation}
where $\mathcal{B(\cdot,\cdot)}$ is the Beta function and $\mathcal{B}(\alpha_c,S-\alpha_c)=\frac{\Gamma(\alpha_c)\cdot\Gamma(S-\alpha_c)}{\Gamma(\alpha_c+S-\alpha_c)}=\frac{\Gamma(\alpha_c)\cdot\Gamma(S-\alpha_c)}{\Gamma(S)}$.

Combining Eq.~\ref{eq:rho} and Eq.~\ref{eq:rho_c}, the posterior probability for class $c$, given the sample $\bx$ and the model $\btheta$, can be obtained as:
\begin{equation}
\small
\begin{aligned}
    P(y=c|\bx,\btheta)&=\int p(y=c|\brho)\cdot p(\brho|\bx,\btheta)\ \text{d}\brho\\
    &=\int \rho_c\cdot p(\brho|\bx,\btheta)\ \text{d}\brho\\
    &=\int\rho_c\cdot p(\rho_c|\bx,\btheta)\ \text{d}\rho_c\\
    &=\frac{\mathcal{B}(\alpha_c+1,S-\alpha_c)}{\mathcal{B}(\alpha_c,S-\alpha_c)}\int\frac{\rho_c^{\alpha_c}(1-\rho_c)^{S-\alpha_c-1}}{\mathcal{B}(\alpha_c+1,S-\alpha_c)}\ \text{d}\rho_c\\
    & = \frac{\mathcal{B}(\alpha_c+1,S-\alpha_c)}{\mathcal{B}(\alpha_c,S-\alpha_c)}\\
    & = \frac{\Gamma(\alpha_c+1)\cdot\Gamma(S)}{\Gamma(S+1)\cdot\Gamma(\alpha_c)}\\
    & = \frac{\alpha_c}{S}.
\end{aligned}
\end{equation}

The Dirichlet distribution parameter $\boldsymbol{\alpha}$ is closely linked to the evidence $\boldsymbol{e}$ which reflects the support for the model prediction on the given sample $\bx$~\cite{sensoy2018evidential}. The parameter $\boldsymbol{\alpha}$ is formulated as:
\begin{equation}
\small
\boldsymbol{\alpha}=\boldsymbol{e}+1=\mathcal{A}(f(\bx,\btheta))+1,
\end{equation}
where $f(\bx,\btheta)$ denotes the output logits of model $\btheta$ for sample $\bx$ and $\mathcal{A}(\cdot)$ is a non-negative activation function to transform the logits $f(\bx,\btheta)$ into evidence $\boldsymbol{e}$. There are several common activation functions $\mathcal{A}(\cdot)$~\cite{pandey2023learn}, including: $ReLU(\cdot)=\max(0,\cdot)$, $SoftPlus(\cdot)=\log(1+\exp(\cdot))$, and $\exp(\cdot)$. In our study, $ReLU(\cdot)$ was employed as the non-negative activation function.

\subsection{Calibrated Evidential Sampling}
\label{subsec:deri_2}
\paragraph{Aleatoric uncertainty.}
Given a sample $\bx$ and the global model $\btheta$, the expected entropy of all possible predictions is utilized to depict the aleatoric uncertainty, quantifying the inherent complexity or ambiguity present in
sample $\bx$. The aleatoric uncertainty of the sample $\bx$ in the global model $\btheta$ is formulated as:
\begin{equation}
\small
\begin{aligned}
    U_{\text{ale}}(\bx,\btheta)
    &= \mathbb{E}_{p(\brho|\bx,\btheta)}[\mathcal{H}[P(y|\brho)]]\\
        &=-\sum_{c=1}^C \mathbb{E}_{p(\brho|\bx,\btheta)}[\rho_c \cdot \log\rho_c]\\
        &=-\sum_{c=1}^C \mathbb{E}_{p(\rho_c|\bx,\btheta)}[\rho_c \cdot \log\rho_c]\\
        &=-\sum_{c=1}^C\int\log(\rho_c)\cdot\frac{\rho_c^{\alpha_c}(1-\rho_c)^{S-\alpha_c-1}}{\mathcal{B}(\alpha_c,S{-}\alpha_c)}\ \text{d}\rho_c\\
        &=-\sum_{c=1}^C\frac{\mathcal{B}(\alpha_c{+}1,S{-}\alpha_c)}{\mathcal{B}(\alpha_c,S{-}\alpha_c)}\int\log(\rho_c)\cdot\frac{\rho_c^{\alpha_c}(1-\rho_c)^{S-\alpha_c-1}}{\mathcal{B}(\alpha_c{+}1,S{-}\alpha_c)}\ \text{d}\rho_c\\
        &=-\sum_{c=1}^C\frac{\Gamma(\alpha_c+1)\cdot\Gamma(S)}{\Gamma(S+1)\cdot\Gamma(\alpha_c)}\mathbb{E}_{\rho_c\sim Beta(\rho_c|\alpha_c+1,S-\alpha_c)}[\log\rho_c]\\
        &=\sum_{c=1}^C\frac{\alpha_c}{S}\cdot[\psi(S+1)-\psi(\alpha_c+1)],
\end{aligned}
\label{eq:ale_suppl}
\end{equation}
where $\mathcal{H}(\cdot)$ denotes the Shannon entropy~\cite{shannon1948mathematical} and $\psi(\cdot)$ represents the digamma function. The aleatoric uncertainty of the sample $\bx$ in the local model $\btheta_k$ can also be calculated as $U_{\text{ale}}(\bx,\btheta_k)$ according to Eq.~\ref{eq:ale_suppl}.

\paragraph{Epistemic uncertainty.}
The differential entropy of a Dirichlet distribution is employed to quantify the inherent randomness in categorical distributions~\cite{malinin2018predictive}. This metric is beneficial in depicting epistemic uncertainty, which arises due to the global model's lack of knowledge, often caused by domain shifts. Given a sample $\bx$ and the global model $\btheta$, the epistemic uncertainty is defined as:
\begin{equation}
\small
\begin{aligned}
    U_{\text{epi}}(\bx,\btheta)&=\mathcal{H}[p(\brho|\bx,\btheta)]\\
    &=-\int p(\brho|\bx,\btheta)\log p(\brho|\bx,\btheta)\ \text{d}\brho\\
    &=\sum_{c=1}^C\log\frac{\Gamma(\alpha_c)}{\Gamma(S)} - (\alpha_c-1)\cdot[\psi(\alpha_c)-\psi(S)].
\end{aligned}
\end{equation}

\subsection{Evidential Model Training}
\label{subsec:deri_3}
\paragraph{Task loss for classification.}
Dirichlet-based evidential models treat the prediction of a sample as a distribution, allowing for multiple potential predictions to occur with specific probabilities. Taking into account all potential predictions for a sample, we employ the Bayes risk of cross-entropy loss~\cite{sensoy2018evidential} as the task loss for classification. Given an input pair $(\bx,\boldsymbol{y})$, the task loss for classification is derived as follows:
\begin{equation}
\small
\begin{aligned}
    \mathcal{L}_{\text{task}}(\bx,\btheta_k,\boldsymbol{y}) &= \mathbb{E}_{p(\brho|\bx,\btheta_k)}[\mathcal{L}_{CE}(\bx,\btheta_k,\boldsymbol{y})]\\
    &=\int [\sum_{c=1}^C-y_c\log(\rho_c)]\cdot p(\brho|\bx,\btheta_k) \ \text{d}\brho\\
    &=-\sum_{c=1}^C y_c\int \log(\rho_c)\cdot p(\rho_c|\bx,\btheta_k)\ \text{d}\rho_c\\
    &=-\sum_{c=1}^C y_c\cdot\mathbb{E}_{\rho_c\sim Beta(\rho_c|\alpha_c,S-\alpha_c)}[\log\rho_c]\\
    &=\sum_{c=1}^Cy_c\cdot[\psi(S)-\psi(\alpha_c)].
\end{aligned}
\end{equation}

\paragraph{Task loss for segmentation.}
We leverage the Bayes risk of Dice loss as the task loss for segmentation following~\cite{li2022region}. Given an input pair $(\bx,\boldsymbol{y})$, the task loss for segmentation is denoted as follows:
\begin{equation}
\small
\begin{aligned}
    \mathcal{L}_{\text{task}}(\bx,\btheta_k,\boldsymbol{y})&=\mathbb{E}_{p(\brho|\bx,\btheta_k)}[\mathcal{L}_{Dice}(\bx,\btheta_k,\boldsymbol{y})]\\
    &=\mathbb{E}_{p(\brho|\bx,\btheta_k)}[1-\frac{2}{C}\sum_{c=1}^{C}\frac{{\Vert\boldsymbol{y}_{c}\circ \brho_{c}\Vert}_1}{{\Vert\boldsymbol{y}_{c}^2\Vert}_1{+}{\Vert\brho_{c}^2\Vert}_1}]\\
    &=\mathbb{E}_{p(\brho|\bx,\btheta_k)}[1-\frac{2}{C}\sum_{c=1}^C\frac{\sum_{m=1}^M y_{mc}\cdot \rho_{mc}}{\sum_{m=1}^M (y_{mc}^2+\rho_{mc}^2)}]\\
    &=1-\frac{2}{C}\sum_{c=1}^C\frac{\sum_{m=1}^My_{mc}\cdot\mathbb{E}[\rho_{mc}]}{\sum_{m=1}^M(y_{mc}^2+\mathbb{E}[\rho_{mc}^2])},
\end{aligned}
\label{eq:dice}
\end{equation}
where $\circ$ is the Hadamard product and the image $\bx$ is comprised of $M$ pixels. $y_{mc}$ and $\rho_{mc}$ represent the label indicator and categorical probability of pixel $x_m$ \textit{w.r.t.} class $c$, respectively. By using the following equation:
\begin{equation}
\small
    \mathbb{E}[\rho_{mc}^2] = \mathbb{E}[\rho_{mc}]^2 + \text{Var}(\rho_{mc}),
\end{equation}
Eq.~\ref{eq:dice} can be updated to:
\begin{equation}
\small
\begin{aligned}
    \mathcal{L}_{\text{task}}(\bx,\btheta_k,\boldsymbol{y})&=1-\frac{2}{C}\sum_{c=1}^C\frac{\sum_{m=1}^M y_{mc}\cdot\overline{\rho}_{mc}}{\sum_{m=1}^M[y_{mc}^2+\overline{\rho}_{mc}^2+\frac{\overline{\rho}_{mc}(1-\overline{\rho}_{mc})}{S_m+1}]}\\
    &= 1-\frac{2}{C}\sum_{c=1}^C\frac{{\Vert\boldsymbol{y}_{c}\circ\overline{\brho}_{c}\Vert}_1}{{\Vert\boldsymbol{y}_{c}^2\Vert}_1+{\Vert\overline{\brho}_{c}^2\Vert}_1+{\Vert\frac{\overline{\brho}_{c}\circ(\boldsymbol{1}-\overline{\brho}_{c})}{\boldsymbol{S}+\boldsymbol{1}}\Vert}_1}.
\end{aligned}
\end{equation}

\section{Experiments}
\label{sec:exp}
\subsection{Experimental Settings}
\label{subsec:exp_setting}
\paragraph{Datasets.}
\begin{table}[htbp]
    \centering
    \scriptsize
    \setlength{\tabcolsep}{2.5pt}
    \begin{tabularx}{\linewidth}{clccc}
    \bottomrule
        Dataset & \multicolumn{1}{c}{Data source} & \# Train & \# Test & Resolution\\
        \hline
        \multirow{4}{*}{Fed-ISIC} & Client 1: BCN~\cite{ogier2022flamby}&9,930 & 2,483 & \multirow{4}{*}{$224{\times} 224$}\\
        ~ & Client 2: HAM\_vidir\_molemax~\cite{ogier2022flamby}& 3,163 & 791 \\
        ~ &Client 3: HAM\_vidir\_modern~\cite{ogier2022flamby}& 2,691 & 672\\
        ~ &Client 4: HAM\_rosendahl~\cite{ogier2022flamby}& 1,807 & 452\\
        \hline
        \multirow{5}{*}{Fed-Camelyon} & Client 1: Camelyon17~\cite{bandi2018detection} & 47,548 & 11,888 & \multirow{5}{*}{$96{\times} 96$}\\
        ~ & Client 2: Camelyon17~\cite{bandi2018detection} & 27,923 & 6,981\\
        ~ & Client 3: Camelyon17~\cite{bandi2018detection} & 68,043 & 17,011\\
        ~ & Client 4: Camelyon17~\cite{bandi2018detection} & 103,870 & 25,968\\
        ~ & Client 5: Camelyon17~\cite{bandi2018detection} & 117,377 & 29,345\\
        \hline
        \multirow{4}{*}{Fed-Polyp} &
        Client 1: Kvasir~\cite{jha2020kvasir} & 800 & 200 &\multirow{4}{*}{$384{\times} 384$} \\
        ~ & Client 2: ETIS~\cite{silva2014toward} & 157 & 39\\
        ~ & Client 3: ColonDB~\cite{tajbakhsh2015automated} & 304 & 75 \\
        ~ & Client 4: ClinicDB~\cite{bernal2015wm} & 490 & 122\\
        \hline
        \multirow{6}{*}{Fed-Prostate} & 
         Client 1: BIDMC~\cite{litjens2014evaluation} & 225 & 36 & \multirow{6}{*}{$384{\times} 384$}\\
        ~ & Client 2: BMC~\cite{bloch2015nci} & 306 & 78\\
        ~ & Client 3: HK~\cite{litjens2014evaluation} & 134 & 24\\
        ~ & Client 4: I2CVB~\cite{lemaitre2015computer} & 387 & 81\\
        ~ & Client 5: RUNMC~\cite{bloch2015nci} &337 & 84 \\
        ~ & Client 6: UCL~\cite{litjens2014evaluation} & 152 & 23\\
        \hline
        \multirow{4}{*}{Fed-Fundus} & Client 1: Drishti-GS~\cite{sivaswamy2015comprehensive} & 81 & 20 & \multirow{4}{*}{$384{\times} 384$}\\
        ~ & Client 2: RIM-ONE-r3~\cite{fumero2011rim} & 128 & 31\\
        ~ & Client 3: REFUGE~\cite{orlando2020refuge} & 320 & 80\\
        ~ & Client 4: REFUGE~\cite{orlando2020refuge} & 320 & 80\\
    \toprule
    \end{tabularx}
    \caption{Details of multi-center datasets utilized in our study.}
    \label{tab:dataset}
\end{table}

\begin{figure}[htbp]
    \centering
    \includegraphics[width=\linewidth]{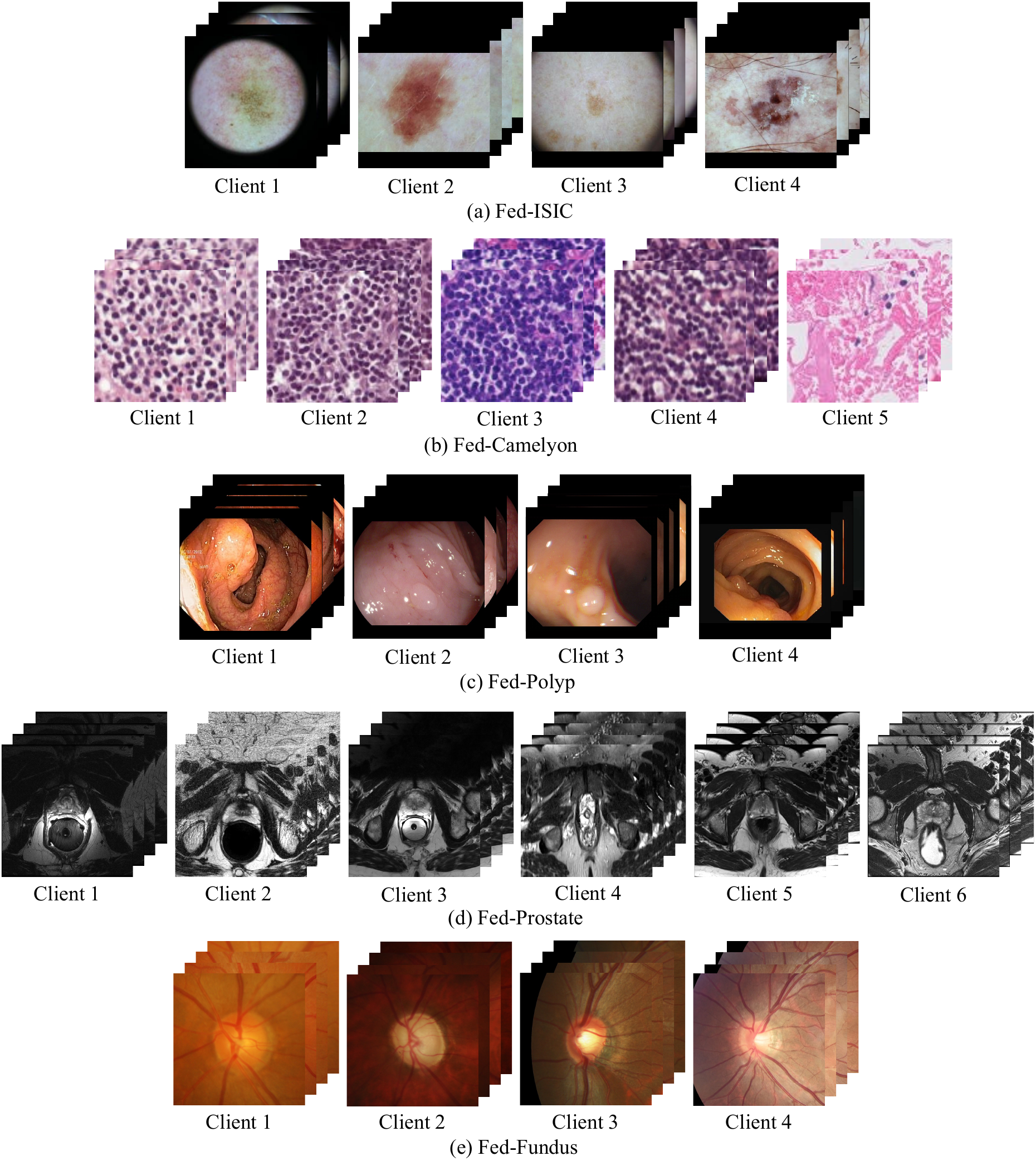}
    \caption{Illustrative samples from each data source within five multi-center medical image datasets utilized in our study.}
    \label{fig:example}
\end{figure}

We verified the effectiveness of FEAL across five real-world medical image datasets, two for classification and three for segmentation. Detailed information, including the data source, the number of samples, and the resolution of each sample, is summarized in Tab.~\ref{tab:dataset}. Illustrative samples from each data source within the five multi-center medical image datasets are showcased in Fig.~\ref{fig:example}. Note that the Camelyon17~\cite{bandi2018detection} dataset comprises five distinct data sources with varying stains, therefore, we partitioned them into five subsets to construct the Fed-Camelyon dataset within the FAL framework. Similarly, the REFUGE~\cite{orlando2020refuge} dataset contains data from two separate sources, each of which was treated as an individual client within the Fed-Fundus dataset. For the three segmentation datasets, we followed~\cite{wang2022personalizing} to resize the images and ground truth to $384{\times}384$ pixels. In the training phase, we implemented data augmentation by randomly cropping patches of $320{\times}320$ pixels. Subsequently, we evaluated the segmentation results on the entire image of $384\times 384$ pixels during the inference phase.

\paragraph{Evaluation metrics.} For classification tasks, we assessed the Balanced Multi-class Accuracy (BMA) for skin lesion classification \cite{cassidy2022analysis} and measured accuracy (ACC) for breast cancer histology classification. Regarding segmentation tasks, we used the Dice score and the 95\% Hausdorff Distance (HD95) to assess segmentation results.

\paragraph{Implemental details.}
We conducted $R=5$ rounds of FAL involving federated model training and data annotation. Federated model training comprises local training and model communication.
During local training, we followed the previous work~\cite{wicaksana2023fca, jiang2022harmofl, wang2022personalizing} to utilize EfficientNet-B0~\cite{tan2019efficientnet} for the Fed-ISIC dataset, DenseNet-121~\cite{huang2017densely} for the Fed-Camelyon dataset, and U-Net~\cite{ronneberger2015u,liu2021feddg} for segmentation datasets. Notably, both EfficientNet-B0 and DenseNet-121 were pre-trained on ImageNet~\cite{deng2009imagenet}. 
In FEAL, we employed the $ReLU(\cdot)$ activation as the non-negative activation function $\mathcal{A}(\cdot)$ for both global and local models.
We trained local models using the Adam optimizer~\cite{kingma2014adam} with a learning rate of $5e{-}4$. The weight decay was set to $5e{-}4$ for the Fed-ISIC dataset and $1e{-}5$ for the other four datasets. In terms of communication, we employed the FedAvg algorithm~\cite{konevcny2016federated}, with all clients participating in each communication round. We conducted $T=100$ rounds of communication to attain a robust global model. Regarding data annotation, we followed the previous work~\cite{kim2023re, cao2023knowledge} to a uniform annotation budget $B_k$ across all local clients $k$. Specifically, annotation budget $B_k$ is allocated based on the size of each dataset. The annotation budget $B_k$ was set to $500$ for both Fed-ISIC and Fed-Camelyon datasets, $50$ for the Fed-Polyp dataset, and $20$ for both Fed-Prostate and Fed-Fundus datasets. 
Furthermore, to account for the different sizes of datasets among local clients, we established a maximum annotation ratio of $85\%$. It means that the clients whose number of labeled samples achieves the threshold ceased further data annotation. All the experiments were conducted three times using different random seeds, and the average results were reported.

\begin{table*}[t]
\caption{Comparison results ($\text{mean}_{\pm \text{std}}$) on medical image datasets for classification. We evaluated the balanced multi-class accuracy (BMA) for Fed-ISIC and the accuracy for Fed-Camelyon and presented the mean result and standard deviation of three random seeds. \red{Red} and \blue{blue} highlights the Top-1 and Top-2 results, respectively.}
\label{tab:cls_suppl}
\resizebox{\textwidth}{!}{
\begin{tabular}{c|l|cccc|cccc}
\bottomrule
\multirow{2}{*}{Model} & \multicolumn{1}{c|}{\multirow{2}{*}{Method}} & \multicolumn{4}{c|}{Fed-ISIC (\%)} & \multicolumn{4}{c}{Fed-Camelyon (\%)} \\
& \multicolumn{1}{c|}{} & R2 & \multicolumn{1}{c}{R3} & \multicolumn{1}{c}{R4} & \multicolumn{1}{c|}{R5} & \multicolumn{1}{c}{R2} & \multicolumn{1}{c}{R3} & \multicolumn{1}{c}{R4} & \multicolumn{1}{c}{R5} \\ \hline
- & Random & $61.59_{\pm1.45}$ & $64.90_{\pm1.53}$ & $65.53_{\pm1.31}$ & $64.99_{\pm1.43}$& $94.82_{\pm0.30}$ & $95.40_{\pm0.24}$ & $96.02_{\pm0.12}$ & $96.34_{\pm0.07}$  \\ \hline
\multirow{5}{*}{$G$} & Entropy~\cite{shannon1948mathematical} & $61.82_{\pm1.38}$ & $65.74_{\pm1.62}$ & $65.99_{\pm0.35}$ & $64.44_{\pm1.21}$&  $94.91_{\pm0.54}$ & $95.98_{\pm0.14}$ & $96.53_{\pm0.16}$ & $96.84_{\pm0.14}$  \\
& TOD~\cite{huang2021semi} & $56.63_{\pm3.05}$ & $64.13_{\pm3.50}$ & $65.32_{\pm1.10}$ & $64.54_{\pm0.76}$ & $93.48_{\pm1.23}$ & $95.47_{\pm0.54}$ & $96.41_{\pm0.31}$ & $96.81_{\pm0.16}$ \\
& Gradnorm~\cite{wang2022boosting} & $63.20_{\pm0.49} $& $65.72_{\pm1.29}$ & $65.31_{\pm2.03}$ & $64.61_{\pm0.76}$ &  $94.10_{\pm0.18}$ & $94.95_{\pm0.23}$ & $95.27_{\pm0.15}$ & $95.73_{\pm0.18}$   \\
& CoreSet~\cite{sener2017active} & $62.19_{\pm1.57}$ & $66.73_{\pm0.49}$ & $66.33_{\pm0.30}$ & $65.62_{\pm1.09}$& $93.89_{\pm0.53}$ & $94.15_{\pm0.26}$ & $95.14_{\pm0.35}$ & $96.04_{\pm0.18}$ \\
& BADGE~\cite{ash2020deep} &$62.26_{\pm1.74}$ &$ 64.98_{\pm1.56}$ & $65.46_{\pm1.94}$ &$ 64.39_{\pm0.63}$& $94.87_{\pm0.38}$ & $95.55_{\pm0.19}$ & $96.07_{\pm0.16}$ & $96.29_{\pm0.25}$ \\ \hline

\multirow{5}{*}{$L$} & Entropy~\cite{shannon1948mathematical} & $62.61_{\pm3.39}$ & $64.95_{\pm1.87}$ & $66.93_{\pm1.44}$ & $65.76_{\pm2.34}$ & $95.07_{\pm0.24}$ & $96.05_{\pm0.03}$ & $96.68_{\pm0.07}$ & $96.73_{\pm0.12}$ \\ 
& TOD~\cite{huang2021semi} & $58.52_{\pm2.89}$ & $65.11_{\pm2.04}$ &$ 64.88_{\pm2.00}$ & $64.81_{\pm3.11}$ &$ 93.79_{\pm1.18}$ & $95.29_{\pm0.47}$ & $96.46_{\pm0.34}$ & $96.83_{\pm0.11}$ \\ 
& Gradnorm~\cite{wang2022boosting} & $62.38_{\pm1.90}$ & $64.96_{\pm3.21}$ & $65.85_{\pm2.79}$ & $64.39_{\pm1.44}$ & $94.22_{\pm0.38} $& $94.98_{\pm0.39}$ & $95.37_{\pm0.54}$ & $95.92_{\pm0.26}$ \\ 
& CoreSet~\cite{sener2017active} & $63.16_{\pm0.70}$ & $66.84_{\pm0.49} $& $66.43_{\pm0.89}$ & $66.40_{\pm0.36}$ & $93.92_{\pm0.40}$ & $94.12_{\pm0.24}$ & $95.24_{\pm0.34}$ & $95.87_{\pm0.17} $\\ 
& BADGE~\cite{ash2020deep} & $63.12_{\pm0.72}$ & $65.54_{\pm1.47}$ & $65.41_{\pm1.77}$ & $64.80_{\pm2.39}$ & \blue{$95.09_{\pm0.23}$} & $95.73_{\pm0.27} $& $96.14_{\pm0.21}$ & $96.50_{\pm0.05} $\\ \hline

\multirow{8}{*}{$E$} & Entropy~\cite{shannon1948mathematical} & $63.21_{\pm0.59}$ & $64.86_{\pm1.09}$ & $66.35_{\pm0.14}$ & $65.57_{\pm1.92}$  & $95.03_{\pm0.01}$ & $96.08_{\pm0.23}$ & $96.52_{\pm0.20}$ & $96.88_{\pm0.18}$  \\ 
 & TOD~\cite{huang2021semi} & $58.10_{\pm1.95}$ & \blue{$66.56_{\pm0.36}$} & $66.26_{\pm1.22}$ & $65.51_{\pm0.75}$  & $93.17_{\pm0.87}$ & $95.27_{\pm0.07}$ & $96.07_{\pm0.09}$ & $96.50_{\pm0.12}$  \\ 
 & Gradnorm~\cite{wang2022boosting} & \blue{$63.23_{\pm1.25}$} & $66.14_{\pm1.51}$ & \blue{$67.02_{\pm1.00}$} & $66.52_{\pm0.75}$  & $94.40_{\pm0.06}$ & $94.85_{\pm0.21}$ & $95.64_{\pm0.04}$ & $95.89_{\pm0.13}$  \\ 
 & CoreSet~\cite{sener2017active} & $62.53_{\pm1.50}$ & $65.91_{\pm0.78}$ & $66.61_{\pm0.20}$ & \blue{$66.84_{\pm0.21}$}  & $93.90_{\pm0.25}$ & $93.95_{\pm0.31}$ & $94.94_{\pm0.09}$ & $95.85_{\pm0.13}$  \\ 
 & BADGE~\cite{ash2020deep} & $59.45_{\pm0.67}$ & $64.27_{\pm0.74}$ & $66.73_{\pm0.46}$ & $64.71_{\pm1.07}$  & $94.97_{\pm0.41}$ & $95.62_{\pm0.11}$ & $96.25_{\pm0.12}$ & $96.37_{\pm0.10}$  \\ 
 & LoGo~\cite{kim2023re} & $62.36_{\pm2.30}$ & $66.43_{\pm0.69}$ & $66.12_{\pm2.64}$ & $66.26_{\pm0.50}$  & $94.98_{\pm0.07}$ & $95.60_{\pm0.15}$ & $96.20_{\pm0.26}$ & $96.51_{\pm0.05}$  \\ 
 & KAFAL~\cite{cao2023knowledge} & $62.34_{\pm0.3}$ & $65.36_{\pm1.15}$ & $66.26_{\pm1.22}$ & $66.24_{\pm1.31}$ & $95.06_{\pm0.17}$ & \blue{$96.08_{\pm0.07}$} & \blue{$96.76_{\pm0.11}$} & \blue{$96.92_{\pm0.04}$} \\ 
& \cellcolor{gray!20}\textbf{FEAL (Ours)} & \cellcolor{gray!20}\red{$65.18_{\pm 0.41}$} & \cellcolor{gray!20}\red{$67.77_{\pm1.31}$} & \cellcolor{gray!20}\red{$68.41_{\pm1.01}$} & \cellcolor{gray!20}\red{$68.46_{\pm0.37}$} & \cellcolor{gray!20}\red{$95.79_{\pm0.68}$} & \cellcolor{gray!20}\red{$96.54_{\pm0.40}$} & \cellcolor{gray!20}\red{$97.04_{\pm0.28}$} & \cellcolor{gray!20}\red{$97.29_{\pm0.35}$}\\ 
\toprule
\end{tabular}
}
\end{table*}

\paragraph{Comparison methods.}
We compared FEAL with eight state-of-the-art FAL approaches, including random sampling (Random), entropy-based sampling (Entropy)~\cite{shannon1948mathematical}, TOD~\cite{huang2021semi}, Gradnorm~\cite{wang2022boosting}, CoreSet~\cite{sener2017active}, BADGE~\cite{ash2020deep}, LoGo~\cite{kim2023re}, and KAFAL~\cite{cao2023knowledge}. The first six methods are designed for standard active learning, whereas LoGo and KAFAL are specifically tailored for federated scenarios. For a comprehensive comparison, we implemented the first six sampling strategies in three distinct manners: (1) utilizing the global model for data evaluation (denoted as $G$), (2) employing the local model for data evaluation ($L$), and (3) integrating an ensemble technique that harnesses both global and local models for evaluation ($E$). The details of comparison methods and ensemble techniques are summarized as follows.
\begin{itemize}
    \item \textbf{Random}: Randomly select $B_k$ unlabeled samples for local client $k$ in each FAL round.
    \item \textbf{Entropy}: Entropy is an uncertainty-based sampling strategy, which prioritizes the top-$B_k$ unlabeled samples with the highest entropy scores in model predictions. Beyond just utilizing the global or local model for data evaluation, we also implemented a simple ensemble strategy that aggregates the entropy scores in both models.
    \item \textbf{TOD}: TOD is an uncertainty-based sampling strategy, which leverages the temporal output discrepancy to quantify the uncertainty of unlabeled samples. It selects the top-$B_k$ unlabeled samples with the highest cyclic output discrepancy (COD) scores. In our implementation, we incorporated an additional ensemble technique~\cite{kim2023re}, fine-tuning, into TOD. Specifically, local client $k$ downloads the global model $\btheta^{r}$ and fine-tunes it with the available labeled samples. Subsequently, local client $k$ employs both the fine-tuned global model and its historical local model $\btheta_k^{r-1}$ to compute the COD score, thereby integrating insights from both global and local models for more effective sampling.
    \item \textbf{Gradnorm}: Gradnorm initially estimates the pseudo loss of unlabeled samples, leveraging either pseudo labels or entropy scores derived from model predictions. This loss estimation is then backpropagated to determine the gradient norm across all model parameters, serving as a measure of data uncertainty to evaluate the potential value of each unlabeled sample. In our study, we aggregated the gradient norm from both models to achieve a fundamental ensemble setting. This approach enabled us to leverage the knowledge of both global and local models, thereby facilitating a more comprehensive evaluation.
    \item \textbf{CoreSet}: CoreSet is a diversity-based sampling strategy that identifies and selects $B_K$ unlabeled samples with feature embeddings exhibiting the greatest dissimilarity to available labeled samples. In a basic ensemble setting, we averaged the feature embeddings from both global and local models for each sample and applied CoreSet on the interpolated feature embedding.
    \item \textbf{BADGE}: BADGE is a hybrid sampling strategy that simultaneously considers uncertainty and diversity metrics. It begins by extracting gradient embeddings of unlabeled samples to ensure uncertainty. Following this, it employs $k$-means clustering on these gradient embeddings to maintain a diverse selection of samples. 
    \item \textbf{LoGo}: LoGo is specifically designed to address class-imbalance issues in federated active learning, harnessing insights from both global and local models. It first performed $k$-means clustering with gradient embeddings extracted from the local model. Then it applied cluster-wise sampling to select the most uncertain sample within each cluster, identified by the highest entropy score in the global model.
    \item \textbf{KAFAL}: KAFAL is another sampling strategy tailored for federated active learning with class imbalance. It employs a knowledge-specialized KL divergence, calculated between the global and local models, to quantify the informativeness of unlabeled samples for both models. For fair comparisons in our study, we implemented KAFAL by exclusively using the supervised loss component.
\end{itemize}

\begin{figure*}[!ht]
    \centering
    \begin{subfigure}{0.3\linewidth}
        \centering
        \includegraphics[width=\linewidth]{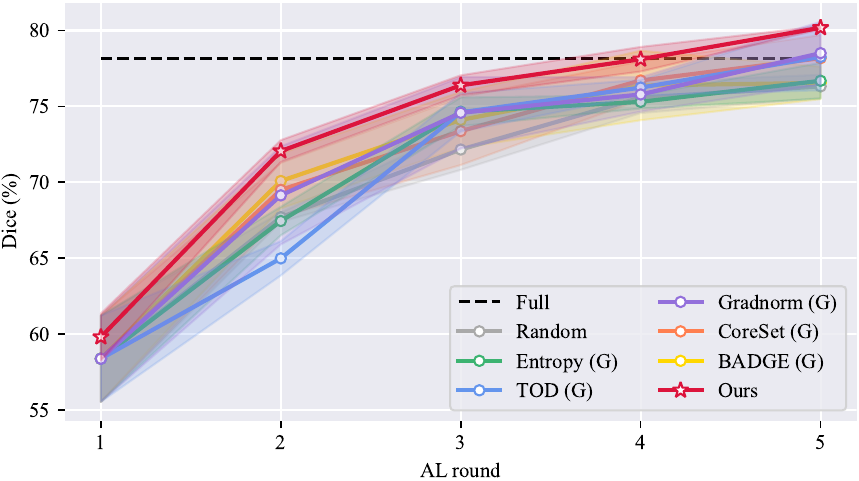}
        \caption{Dice of Fed-Polyp ($G$)}
    \end{subfigure}
    \hspace{2mm}
    \begin{subfigure}{0.3\linewidth}
        \centering
        \includegraphics[width=\linewidth]{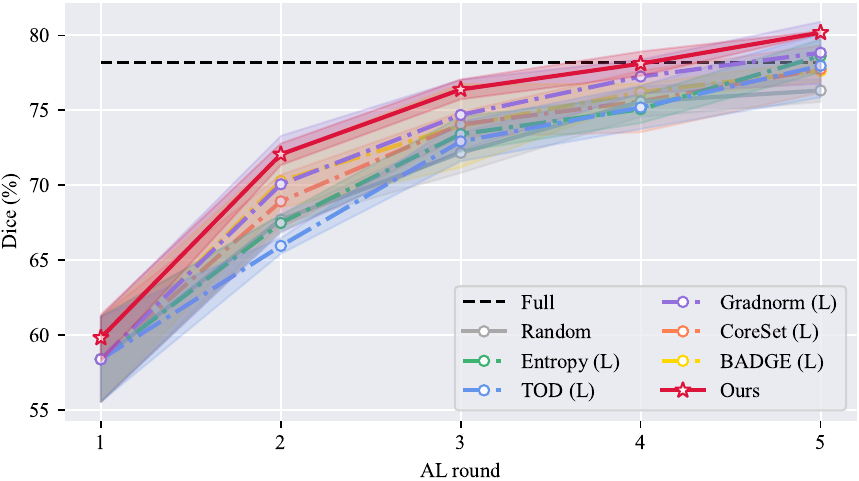}
        \caption{Dice of Fed-Polyp ($L$)}
    \end{subfigure}
    \hspace{2mm}
    \begin{subfigure}{0.3\linewidth}
        \centering
        \includegraphics[width=\linewidth]{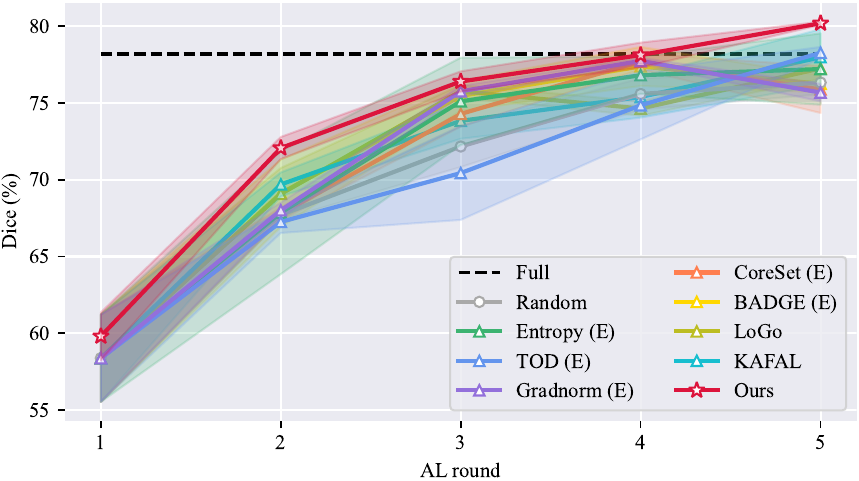}
        \caption{Dice of Fed-Polyp ($E$)}
    \end{subfigure}

    \begin{subfigure}{0.3\linewidth}
        \centering
        \includegraphics[width=\linewidth]{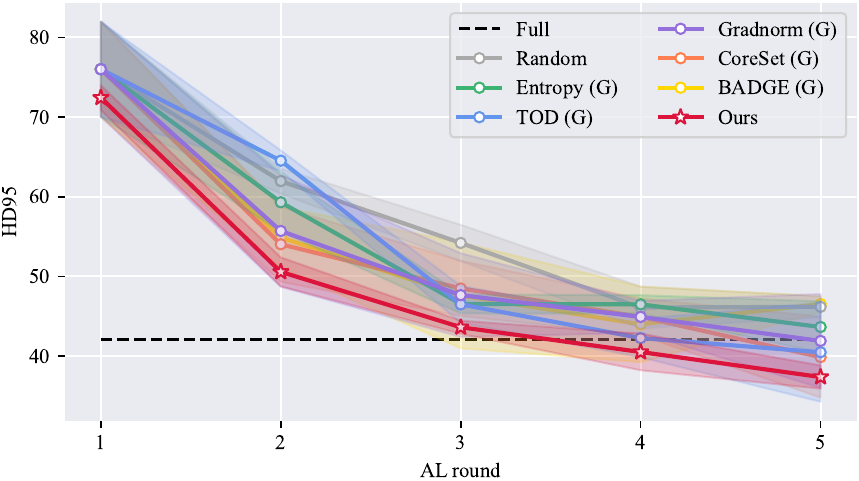}
        \caption{HD95 of Fed-Polyp ($G$)}
    \end{subfigure}
    \hspace{2mm}
    \begin{subfigure}{0.3\linewidth}
        \centering
        \includegraphics[width=\linewidth]{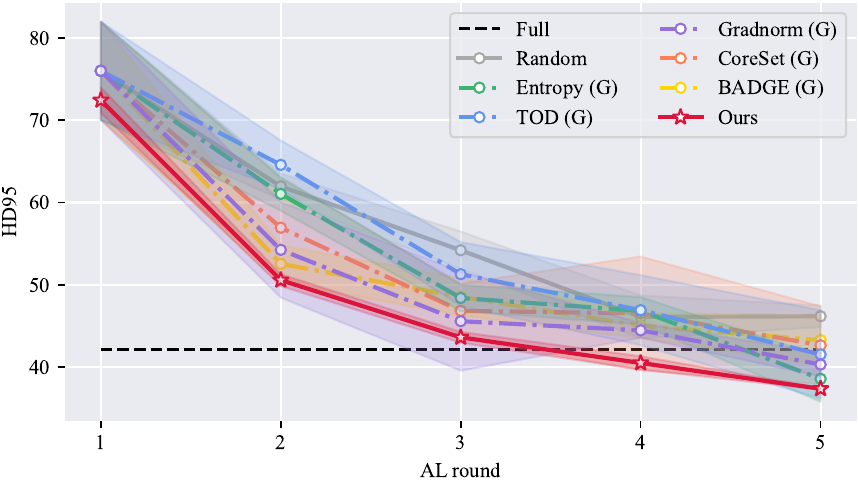}
        \caption{HD95 of Fed-Polyp ($L$)}
    \end{subfigure}
    \hspace{2mm}
    \begin{subfigure}{0.3\linewidth}
        \centering
        \includegraphics[width=\linewidth]{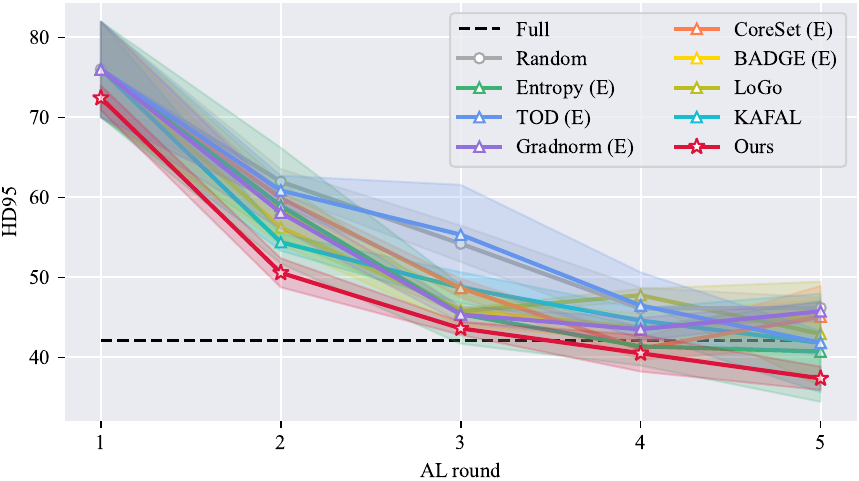}
        \caption{HD95 of Fed-Polyp ($E$)}
    \end{subfigure}
    \vspace{-2mm}
    \caption{Comparison results on the Fed-Polyp dataset.}
    \vspace{-1.5mm}
    \label{fig:polyp_suppl}
\end{figure*}
\begin{figure*}[!ht]
    \centering
    \begin{subfigure}{0.3\linewidth}
        \centering
        \includegraphics[width=\linewidth]{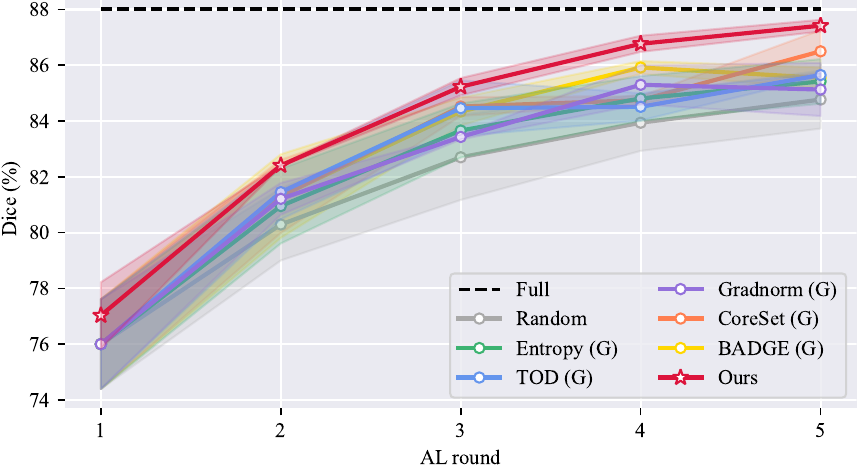}
        \caption{Dice of Fed-Prostate ($G$)}
    \end{subfigure}
    \hspace{2mm}
    \begin{subfigure}{0.3\linewidth}
        \centering
        \includegraphics[width=\linewidth]{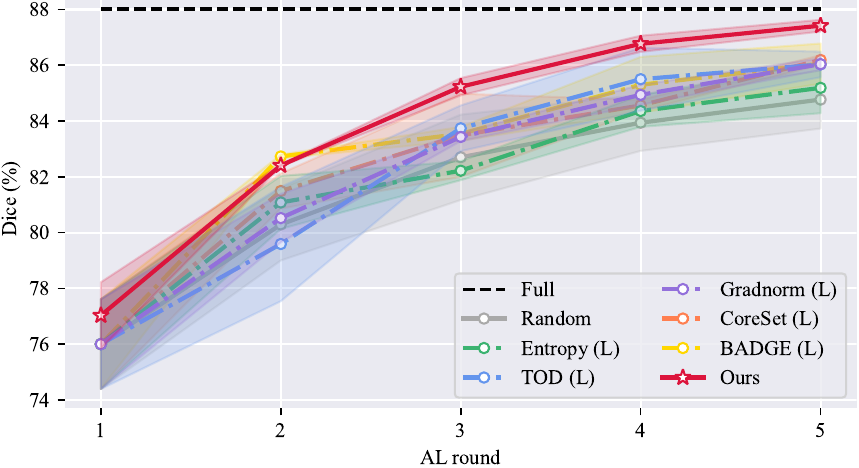}
        \caption{Dice of Fed-Prostate ($L$)}
    \end{subfigure}
    \hspace{2mm}
    \begin{subfigure}{0.3\linewidth}
        \centering
        \includegraphics[width=\linewidth]{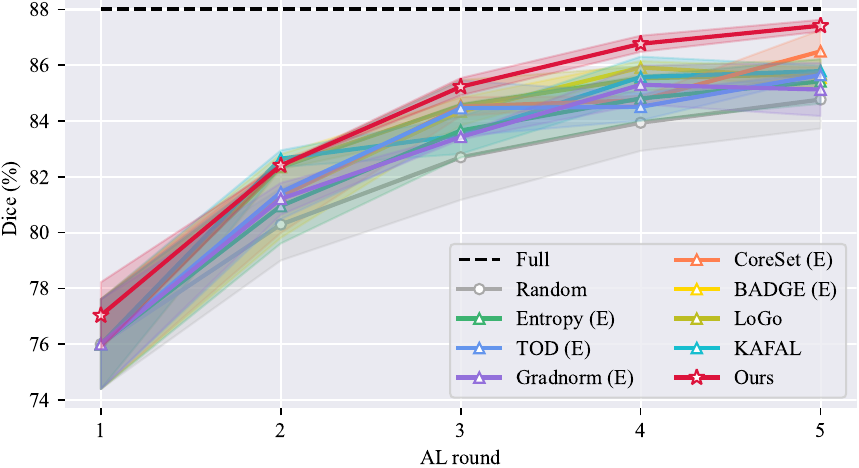}
        \caption{Dice of Fed-Prostate ($E$)}
    \end{subfigure}

    \begin{subfigure}{0.3\linewidth}
        \centering
        \includegraphics[width=\linewidth]{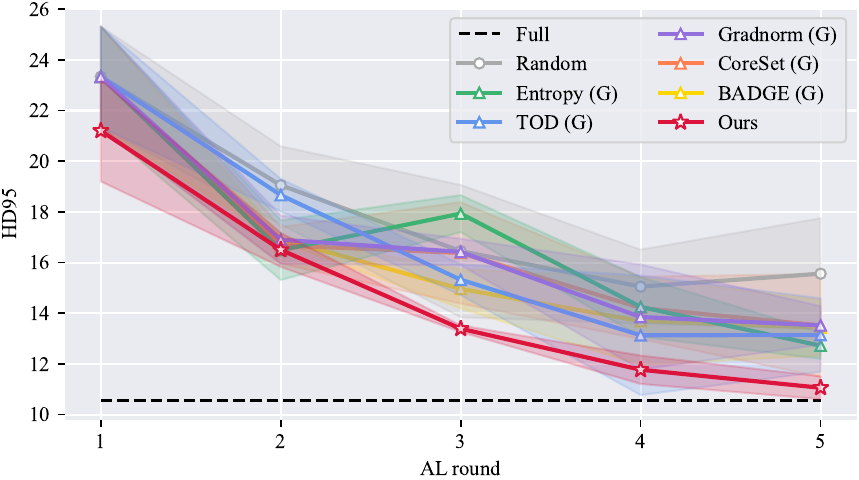}
        \caption{HD95 of Fed-Prostate ($G$)}
    \end{subfigure}
    \hspace{2mm}
    \begin{subfigure}{0.3\linewidth}
        \centering
        \includegraphics[width=\linewidth]{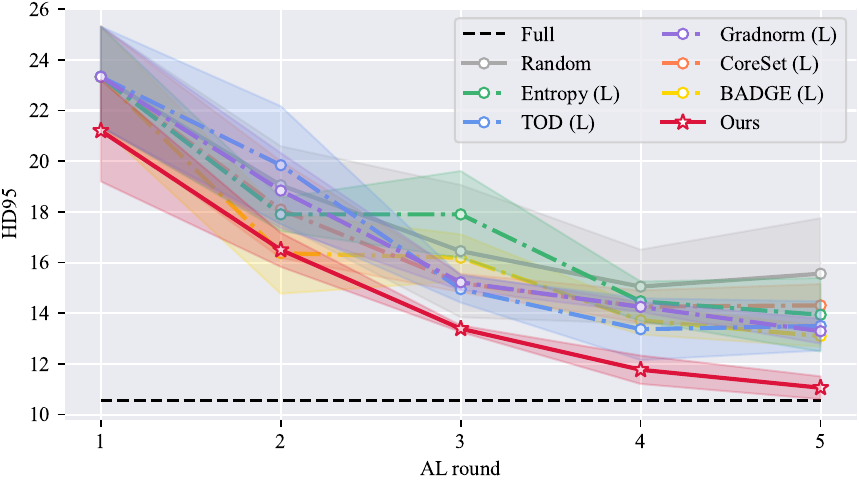}
        \caption{HD95 of Fed-Prostate ($L$)}
    \end{subfigure}
    \hspace{2mm}
    \begin{subfigure}{0.3\linewidth}
        \centering
        \includegraphics[width=\linewidth]{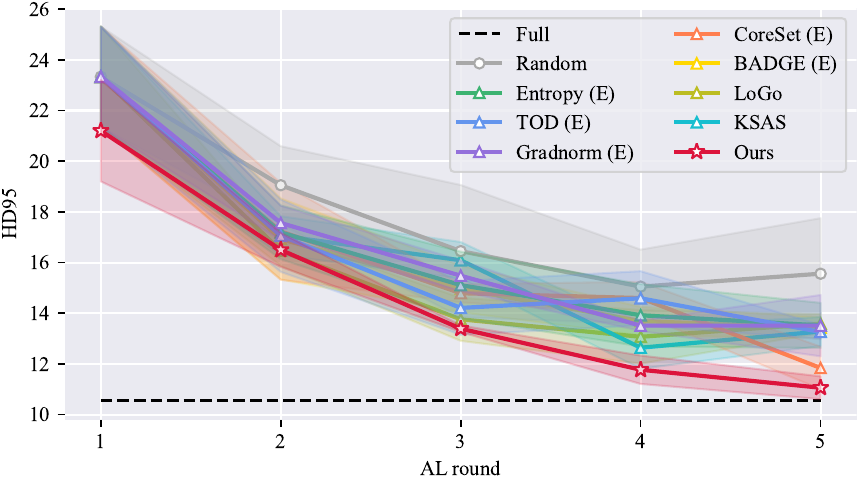}
        \caption{HD95 of Fed-Prostate ($E$)}
    \end{subfigure}
    \vspace{-2mm}
    \caption{Comparison results on the Fed-Prostate dataset.}
    \vspace{-1.5mm}
    \label{fig:prostate_suppl}
\end{figure*}
\begin{figure*}[!ht]
    \centering
    \begin{subfigure}{0.3\linewidth}
        \centering
        \includegraphics[width=\linewidth]{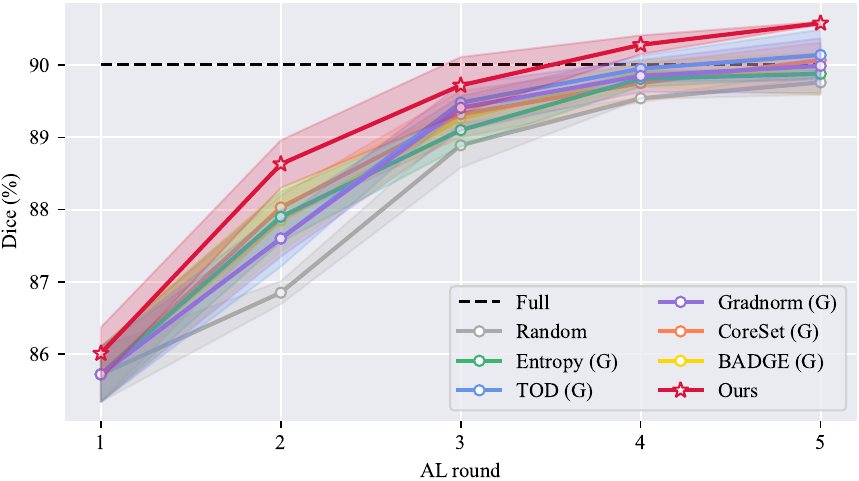}
        \caption{Dice of Fed-Fundus ($G$)}
    \end{subfigure}
    \hspace{2mm}
    \begin{subfigure}{0.3\linewidth}
        \centering
        \includegraphics[width=\linewidth]{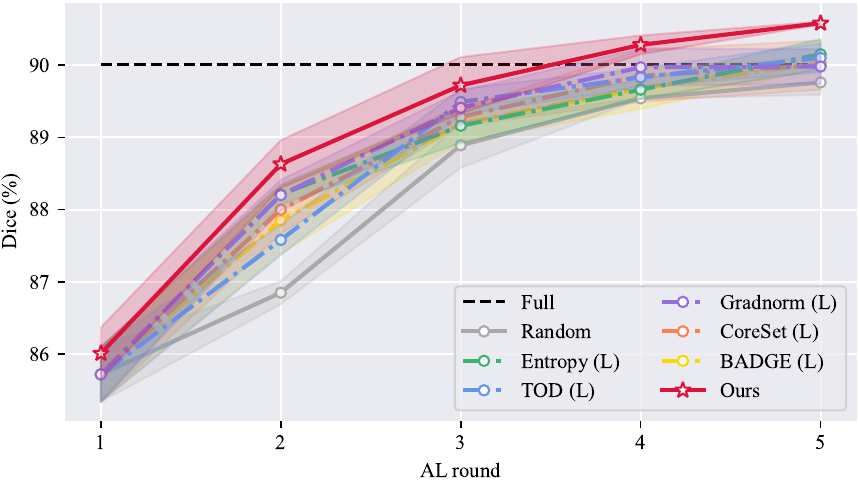}
        \caption{Dice of Fed-Fundus ($L$)}
    \end{subfigure}
    \hspace{2mm}
    \begin{subfigure}{0.3\linewidth}
        \centering
        \includegraphics[width=\linewidth]{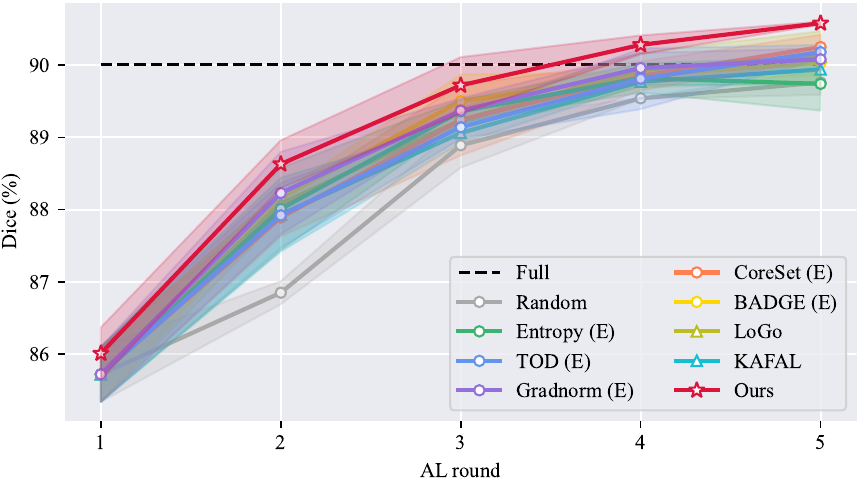}
        \caption{Dice of Fed-Fundus ($E$)}
    \end{subfigure}

    \begin{subfigure}{0.3\linewidth}
        \centering
        \includegraphics[width=\linewidth]{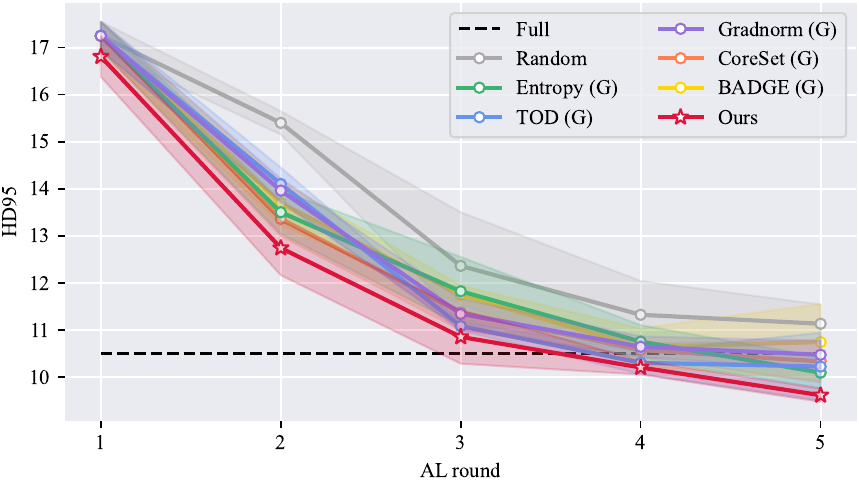}
        \caption{HD95 of Fed-Fundus ($G$)}
    \end{subfigure}
    \hspace{2mm}
    \begin{subfigure}{0.3\linewidth}
        \centering
        \includegraphics[width=\linewidth]{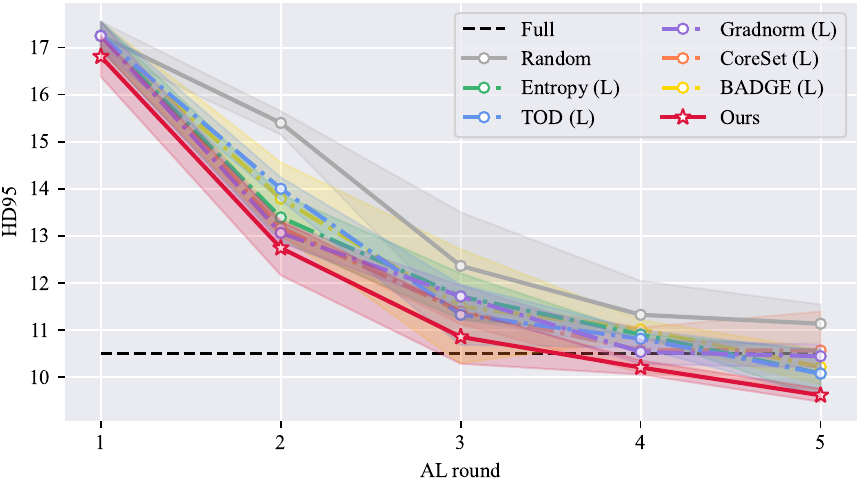}
        \caption{HD95 of Fed-Fundus ($L$)}
    \end{subfigure}
    \hspace{2mm}
    \begin{subfigure}{0.3\linewidth}
        \centering
        \includegraphics[width=\linewidth]{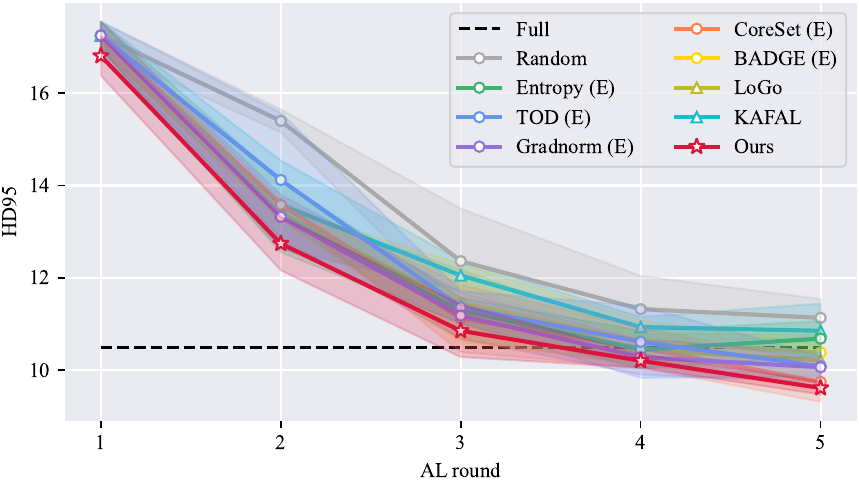}
        \caption{HD95 of Fed-Fundus ($E$)}
    \end{subfigure}
    \vspace{-2mm}
    \caption{Comparison results on the Fed-Fundus dataset.}
    \vspace{-1.5mm}
    \label{fig:fundus_suppl}
\end{figure*}

\subsection{Results}
\label{subsec:result_suppl}
\paragraph{Image classification.}
Tab.~\ref{tab:cls_suppl} summarizes the quantitative results of image classification. From the second to the fifth FAL rounds, FEAL demonstrates superior performance over the second-best method, achieving margins of $1.95\%$, $0.92\%$, $1.39\%$, and $1.62\%$ on the Fed-ISIC dataset, respectively. Furthermore, on the Fed-Camelyon dataset, FEAL maintains a consistent performance advantage, surpassing the second-best method by margins of $0.69\%$, $0.46\%$, $0.3\%$, and $0.37\%$ in these rounds. Additional results with different annotation budgets/ratios on the Fed-ISIC dataset are presented in Sec.~\ref{subsec:discuss_suppl}.

\paragraph{Image segmentaion.}
The mean results and standard deviations of Dice and HD95 metrics for the Fed-Polyp, Fed-Prostate, and Fed-Fundus datasets are illustrated in Fig.~\ref{fig:polyp_suppl}, Fig.~\ref{fig:prostate_suppl}, and Fig.~\ref{fig:fundus_suppl}, respectively. Extensive results in Fig.~\ref{fig:polyp_suppl} - Fig.~\ref{fig:fundus_suppl} indicate that FEAL yields superior performance on three multi-center segmentation datasets, as evidenced by its higher Dice scores and lower HD95 metrics. Notably, FEAL outperforms the second-best method by the margin of $1.78\%$, $0.63\%$, $0.89\%$, and $1.34\%$ from the second to the fifth rounds on the Fed-Polyp dataset.

\subsection{Discussions}
\label{subsec:discuss_suppl}
\begin{figure}[htbp]
    \centering
    \includegraphics[width=\linewidth]{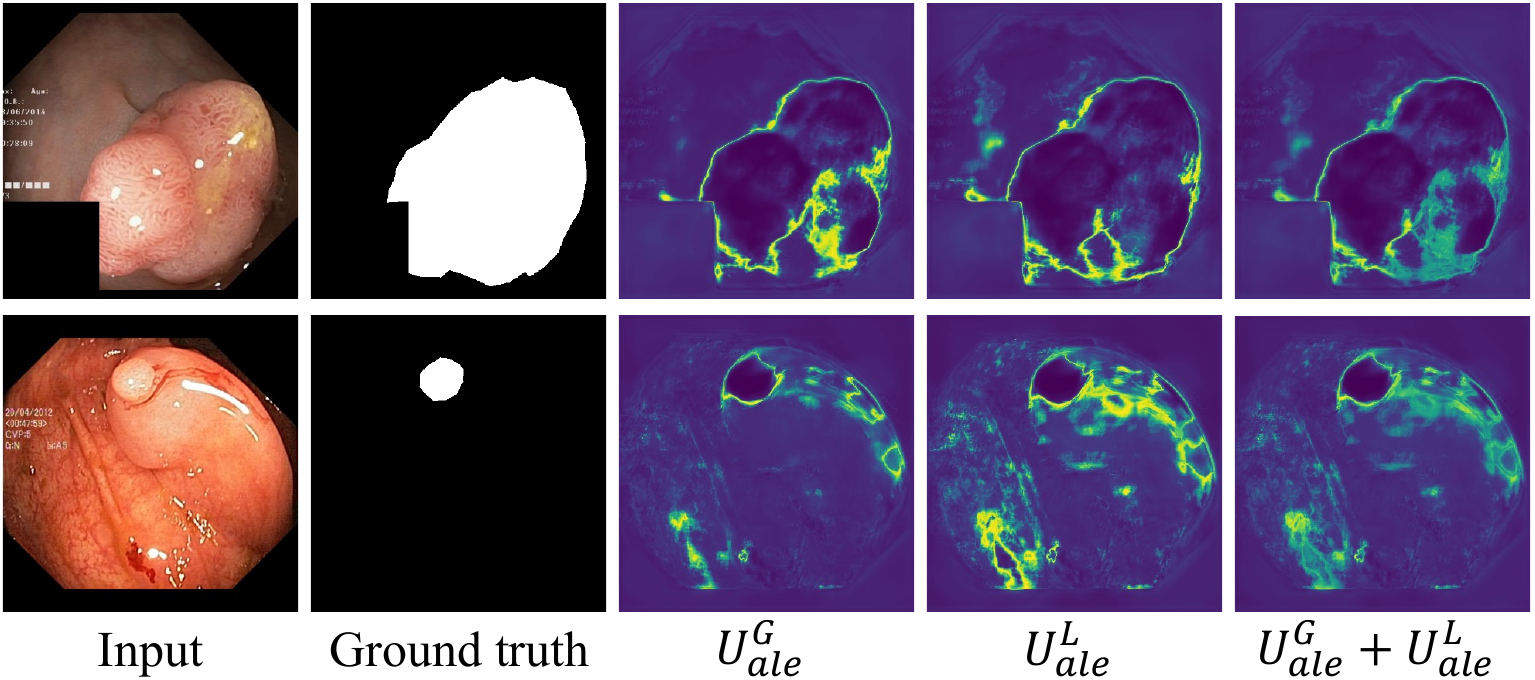}
    \caption{Visualization of aleatoric uncertainty on Fed-Polyp. $U_{\text{ale}}^G$ and $U_{\text{ale}}^L$ denote the aleatoric uncertainty in global and local models, respectively.}
    \label{fig:udata2}
\end{figure}

\paragraph{Effect of uncertainty calibration.}
In addition to the ablation study conducted on the Fed-ISIC dataset for classification, we expanded our analysis to include the Fed-Polyp dataset, specifically examining the impact of uncertainty calibration in segmentation tasks. The results, including the average Dice score and the corresponding standard deviation, are summarized in Tab.~\ref{tab:strategy_suppl}. Here, $U_{\text{epi}}^G$, $U_{\text{ale}}^G$, and $U_{\text{ale}}^L$ represent the epistemic uncertainty in the global model, the aleatoric uncertainty in the global model, and the aleatoric uncertainty in the local model, respectively. As can be seen, the optimal performance is attained when incorporating $U_{\text{epi}}^G$, $U_{\text{ale}}^G$, and $U_{\text{ale}}^L$, demonstrating the effectiveness of the proposed uncertainty calibration method. Furthermore, we visualize the aleatoric uncertainty in both global and local models on the Fed-Polyp dataset in Fig~\ref{fig:udata2}. As depicted in Fig~\ref{fig:udata2}, $U_{\text{ale}}^G$ and $U_{\text{ale}}^L$ highlight different regions of a sample, underscoring the significance of integrating the aleatoric uncertainty in both global and locals for a thorough assessment.
\begin{table}[htbp]
\caption{Ablation study of uncertainty calibration on Fed-Polyp.}
\footnotesize
\setlength{\tabcolsep}{3pt}
\begin{tabularx}{\linewidth}{ccc|cccc}
\bottomrule
 $U_{\text{epi}}^G$ & $U_{\text{ale}}^G$ & $U_{\text{ale}}^L$ & Round 2 & Round 3 & Round 4 & Round 5 \\  
 \hline
 
 - & \checkmark & - & $68.61_{\pm0.48}$ & $73.12_{\pm2.38}$ & $75.19_{\pm1.10}$ & $78.00_{\pm1.14}$\\ 
 
 - & - & \checkmark & $69.13_{\pm1.15}$ & $75.19_{\pm1.29}$ & $77.85_{\pm1.20}$ & $78.12_
{\pm1.29}$\\    
  - & \checkmark & \checkmark & $66.89_{\pm3.41}$ & $74.48_{\pm1.00}$ & $76.56_{\pm1.49}$ & $76.86_{\pm0.43}$\\ 
 
 \hline
 
 \checkmark & - & - & $70.61_{\pm4.22}$ & $74.45_{\pm2.29}$ & $75.07_{\pm2.42}$ & $78.25_{\pm1.29}$\\ 
 
 \checkmark & \checkmark & - & $69.41_{\pm2.17}$ & $75.18_{\pm2.48}$ & $74.66_{\pm0.87}$ & $78.91_{\pm1.04}$\\ 
 \checkmark & - & \checkmark & $69.29_{\pm2.51}$ & $75.93_{\pm1.41}$ & $76.74_{\pm1.08}$ & $78.28_{\pm0.67}$\\   
 \rowcolor{gray!20}\checkmark & \checkmark & \checkmark & $72.06_{\pm0.72}$ & $76.39_{\pm0.66}$ & $78.62_{\pm0.81}$ & $80.18_{\pm0.10}$\\ 
 \toprule
\end{tabularx}
\label{tab:strategy_suppl}
\end{table}

\begin{figure}[htbp]    
    \centering
    \begin{subfigure}{0.45\linewidth}
        \centering
        \includegraphics[width=\linewidth]{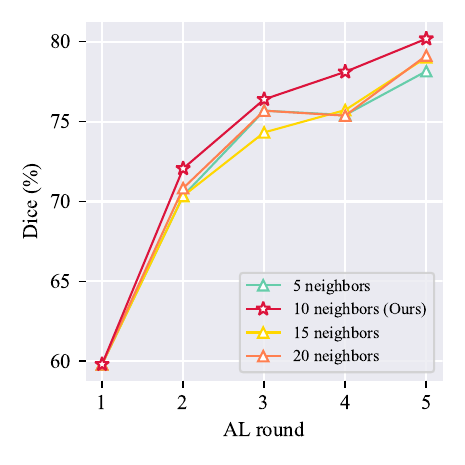}
        \caption{Minimum neighbor size $n$.}
    \end{subfigure} 
    \hspace{4mm}
    \begin{subfigure}{0.45\linewidth}
        \centering
        \includegraphics[width=\linewidth]{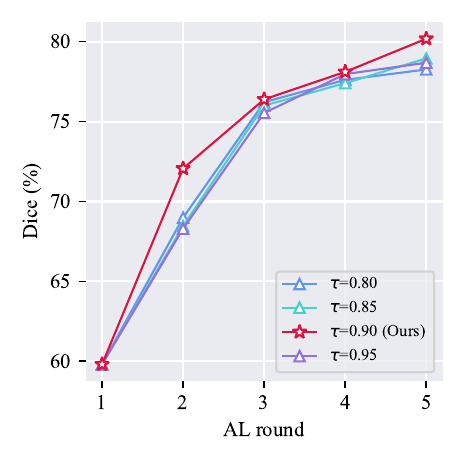}
        \caption{Similarity threshold $\tau$.}
    \end{subfigure} 
    \caption{Ablation study of diversity relaxation on Fed-Polyp.} 
    \label{fig:relaxation_suppl}
\end{figure}

\paragraph{Effect of diversity relaxation.} 
An ablation study was carried out on the Fed-Polyp dataset to investigate the impact of diversity relaxation. As illustrated in Fig.~\ref{fig:relaxation_suppl}, the optimal performance is attained when setting the minimum neighbor size to $n=10$ and the cosine similarity threshold to $\tau=0.90$ on the Fed-Polyp dataset. 

\paragraph{Effect of evidential model training.}
We conducted experiments to compare the evidential loss ($\mathcal{L}$ in Eq.~9) against the cross-entropy loss (CE) on two classification datasets and against dice loss (Dice) on three segmentation datasets.
As summarized in Tab.~\ref{tab:loss_suppl}, the proposed evidential model training yields an average performance gain of $1.03\%$ on the Fed-ISIC dataset, $0.29\%$ on the Fed-Camelyon dataset, $1.16\%$ on the Fed-Polyp dataset, $1.17\%$ on the Fed-Prostate dataset, and $0.36\%$ on the Fed-Fundus dataset.

\begin{table}[hbtp]
\caption{Ablation study of loss function.}
\label{tab:loss_suppl}
\scriptsize
\setlength{\tabcolsep}{3pt}
\begin{tabularx}{\linewidth}{c|c|cccc}
\bottomrule
Dataset & Loss & Round 2 &Round 3 &Round 4 &Round 5 \\
\hline
\multirow{2}{*}{Fed-ISIC} & CE & $64.28_{\pm1.64}$ & $66.69_{\pm0.95}$ & $67.32_{\pm1.16}$ & $67.40_{\pm0.22}$ \\
 & \cellcolor{gray!20}$\mathcal{L}$ & \cellcolor{gray!20}$65.18_{\pm 0.41}$ & \cellcolor{gray!20}$67.77_{\pm1.31}$ & \cellcolor{gray!20}$68.41_{\pm1.01}$ & \cellcolor{gray!20}$68.46_{\pm0.37}$\\
 \hline
 \multirow{2}{*}{Fed-Camelyon} & CE & $95.24_{\pm0.03}$ & $96.21_{\pm0.04}$ & $96.80_{\pm0.07}$ & $97.26_{\pm0.06}$\\
 & \cellcolor{gray!20}$\mathcal{L}$ & \cellcolor{gray!20}$95.79_{\pm0.17}$ & \cellcolor{gray!20}$96.54_{\pm0.08}$ & \cellcolor{gray!20}$97.04_{\pm0.02}$ & \cellcolor{gray!20}$97.29_{\pm0.02}$\\
 \hline
\multirow{2}{*}{Fed-Polyp} & Dice & $70.14_{\pm0.10}$ & $75.77_{\pm0.67}$ & $77.23_{\pm0.21}$ & $79.48_{\pm0.62}$ \\
 & \cellcolor{gray!20}$\mathcal{L}$ & \cellcolor{gray!20}$72.06_{\pm0.72}$ & \cellcolor{gray!20}$76.39_{\pm0.66}$ & \cellcolor{gray!20}$78.62_{\pm1.44}$ & \cellcolor{gray!20}$80.18_{\pm0.10}$\\
 \hline
\multirow{2}{*}{Fed-Prostate} & Dice & $81.43_{\pm0.75}$ & $84.50_{\pm1.02}$ & $85.32_{\pm0.60}$ & $86.50_{\pm0.59}$\\
 & \cellcolor{gray!20}$\mathcal{L}$ & \cellcolor{gray!20}$82.94_{\pm0.04}$ & \cellcolor{gray!20}$85.29_{\pm0.31}$ & \cellcolor{gray!20}$86.77_{\pm0.29}$ & \cellcolor{gray!20}$87.42_{\pm0.21}$\\
 \hline
 \multirow{2}{*}{Fed-Fundus} & Dice & $88.11_{\pm0.3}$ & $89.68_{\pm0.22}$ &$89.84_{\pm0.23}$ & $90.15_{\pm0.19}$\\
 & \cellcolor{gray!20}$\mathcal{L}$ & \cellcolor{gray!20}$88.63_{\pm0.3}$ & \cellcolor{gray!20}$89.72_{\pm0.39}$ & \cellcolor{gray!20}$90.28_{\pm0.13}$ & \cellcolor{gray!20}$90.58_{\pm0.02}$\\
 \toprule
\end{tabularx}
\end{table}

\paragraph{Effect of trade-off weight $\lambda$.}
We additionally conducted experiments to determine the optimal value for the hyperparameter $\lambda$ on the Fed-Polyp dataset, choosing from the candidate set $\{1e{-}5, 5e{-}5, 1e{-}4, 5e{-}4, 1e{-}3\}$. The findings, summarized in Tab.~\ref{tab:hyper_suppl}, reveal the optimal performance is achieved when $\lambda=1e{-}4$ for the Fed-Polyp dataset.
\begin{table}[htbp]
    \caption{Ablation study of trade-off weight $\lambda$ on Fed-Polyp.}
    \centering
        \footnotesize
        \begin{tabularx}{\linewidth}{c|cccc}
        \bottomrule
        $\lambda$ & Round 2 & Round 3 & Round 4 & Round 5\\
         \hline
        $1e{-}5$ & $70.91_{\pm1.89}$ & $76.27_{\pm0.36}$ & $75.98_{\pm1.09}$ & $78.58_{\pm1.85}$\\
        $5e{-}5$ & $70.27_{\pm2.64}$ & $74.68_{\pm1.59}$ & $76.82_{\pm1.00}$ & $78.26_{\pm0.76}$\\
        \rowcolor{gray!20}$1e{-}4$ &$72.06_{\pm0.72}$ & $76.39_{\pm0.66}$ & $78.62_{\pm1.44}$ & $80.18_{\pm0.10}$\\ 
        $5e{-}4$ & $70.90_{\pm1.82}$ & $75.59_{\pm2.02}$ & $77.63_{\pm0.84}$ & $79.76_{\pm1.08}$\\
        $1e{-}3$ & $69.78_{\pm1.64}$ & $74.54_{\pm2.56}$ & $76.92_{\pm1.91}$ & $77.64_{\pm0.88}$\\
        \toprule
        \end{tabularx}
    \label{tab:hyper_suppl}
\end{table}

\begin{figure}[htbp]
    \centering
    \begin{subfigure}{0.49\linewidth}
        \centering
        \includegraphics[width=\linewidth]{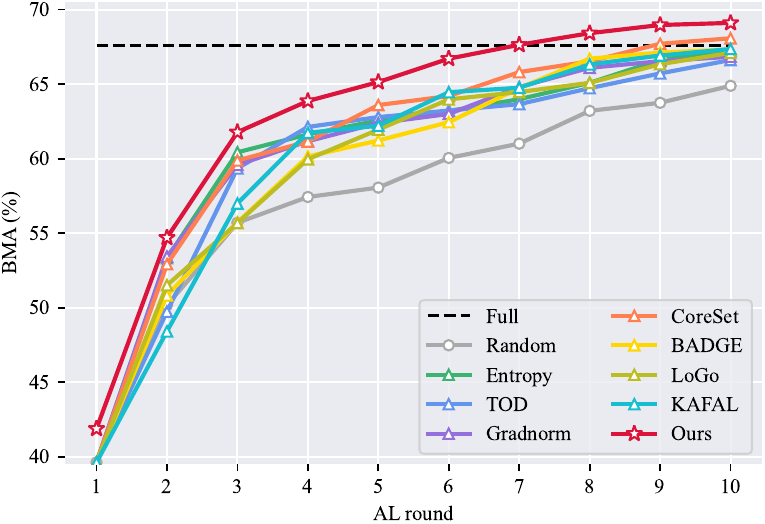}
        \caption{Annotation budget $B_k=250$}
    \end{subfigure}
    \begin{subfigure}{0.49\linewidth}
        \centering
        \includegraphics[width=\linewidth]{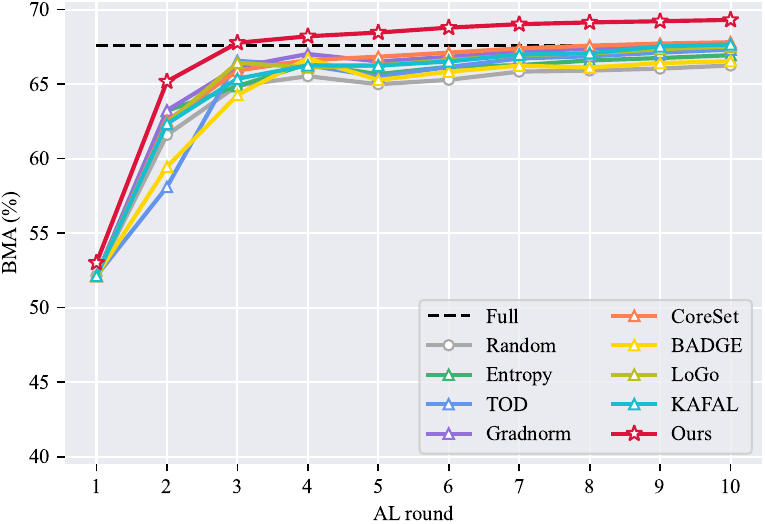}
        \caption{Annotation budget $B_k=500$}
    \end{subfigure}

    \vspace{1mm}
    \begin{subfigure}{0.49\linewidth}
        \centering
        \includegraphics[width=\linewidth]{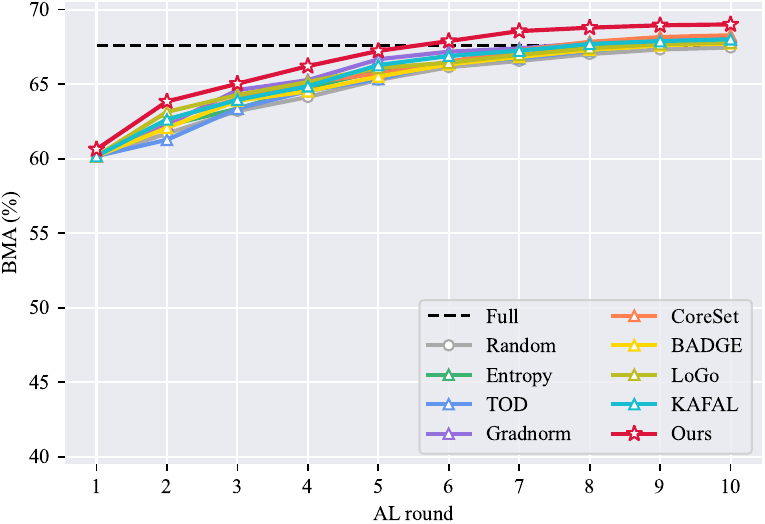}
        \caption{Annotation budget $B_k=750$}
    \end{subfigure}
    \begin{subfigure}{0.49\linewidth}
        \centering
        \includegraphics[width=\linewidth]{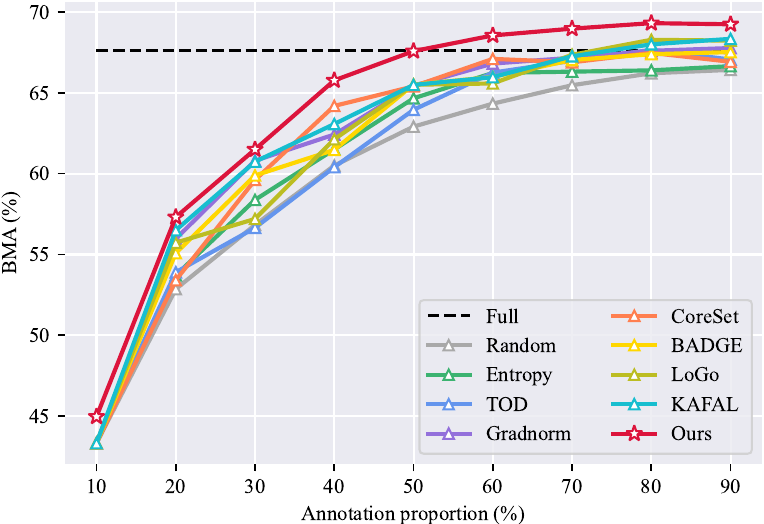}
        \caption{Annotation budget $B_k=\frac{N_k}{10}$}
    \end{subfigure}
    \caption{Ablation study of annotation budget $B_k$ on Fed-ISIC.}
    \label{fig:annotation_budget}
\end{figure}

\paragraph{Effect of annotation budget $B_k$.} \textbf{(1) Fixed-number.} To validate the effectiveness and robustness of FEAL, we further analyzed the impact of annotation budget $B_k$ on the Fed-ISIC dataset. We conducted $R=10$ rounds of FAL with three configurations of $B_k$, \textit{i.e.} $B_k=\{250,500,750\}$, for the Fed-ISIC dataset. As depicted in Fig.~\ref{fig:annotation_budget}(a)-(c), FEAL consistently outperforms other counterparts across all three annotation budget configurations $B_k$, demonstrating its effectiveness and resilience. \textbf{(2) Fixed-ratio.} Furthermore, considering the imbalanced dataset sizes among local clients, we also implemented the fixed-ratio strategy to annotate $10\%$ of the samples for each client in every FAL round. The results depicted in Fig.~\ref{fig:annotation_budget}(d) demonstrate that FEAL surpasses other competing methods even under the fixed-ratio setting. Remarkably, both fixed-number and fixed-ratio strategies exhibit similar performance in later rounds, indicating that FEAL is robust against variations in dataset sizes across local clients.

\begin{figure}[htbp]
    \centering
    \includegraphics[width=\linewidth]{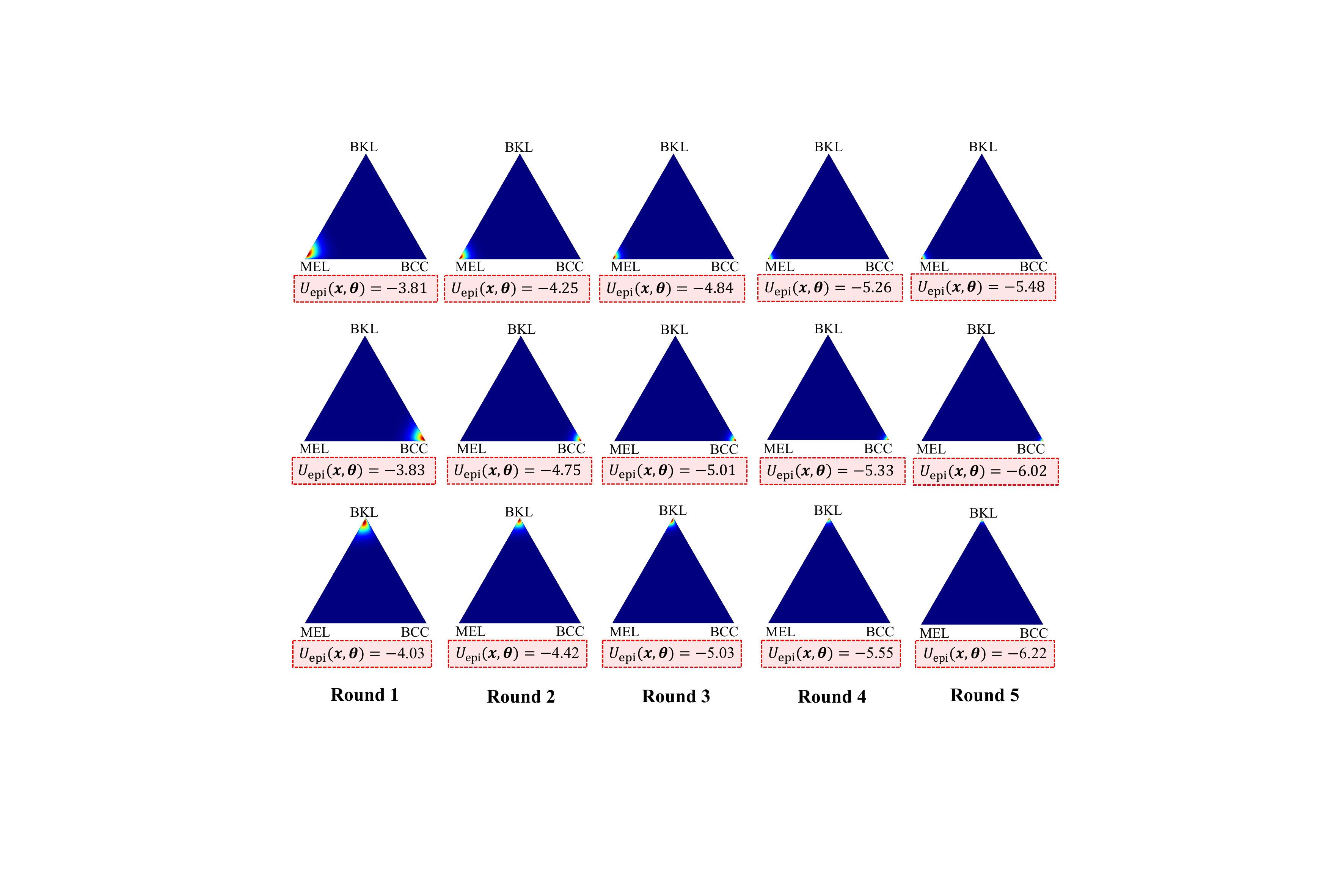}
    \caption{Visualization of the Dirichlet simplex for unlabeled samples across five FAL rounds using FEAL.}
    \label{fig:diri_5rounds}
\end{figure}

\begin{figure}[htbp]
    \centering
    \includegraphics[width=0.95\linewidth]{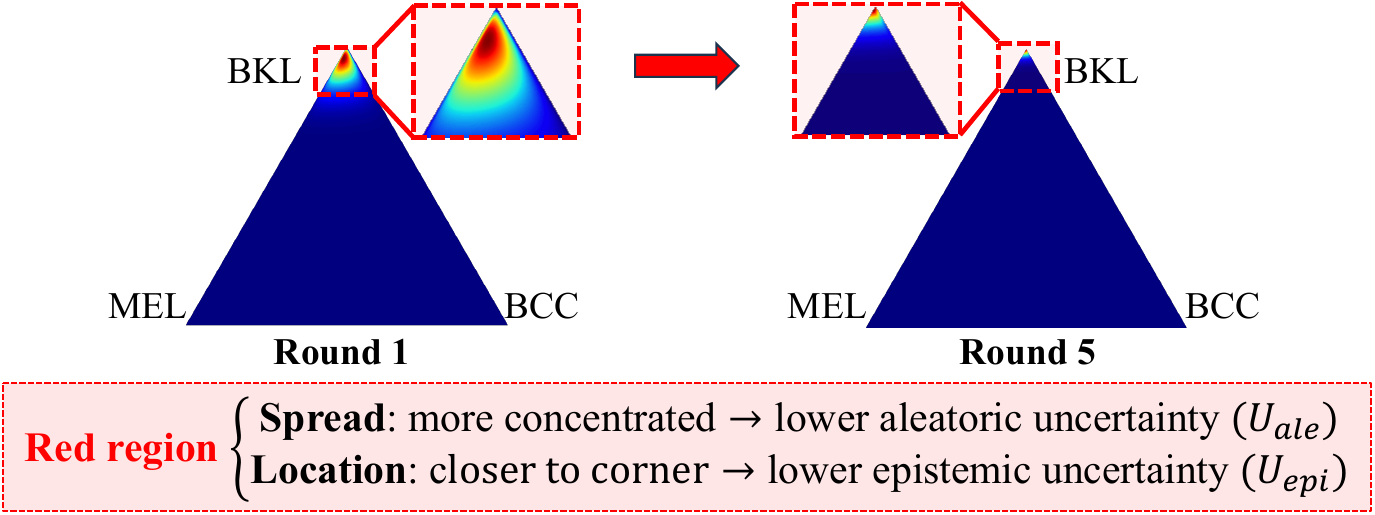}
    \caption{Comparison of the Dirichlet simplex for unlabeled samples in the first and fifth FAL round using FEAL.}
    \label{fig:r1_r5}
\end{figure}

\begin{figure}[htbp]
    \centering
    \includegraphics[width=\linewidth]{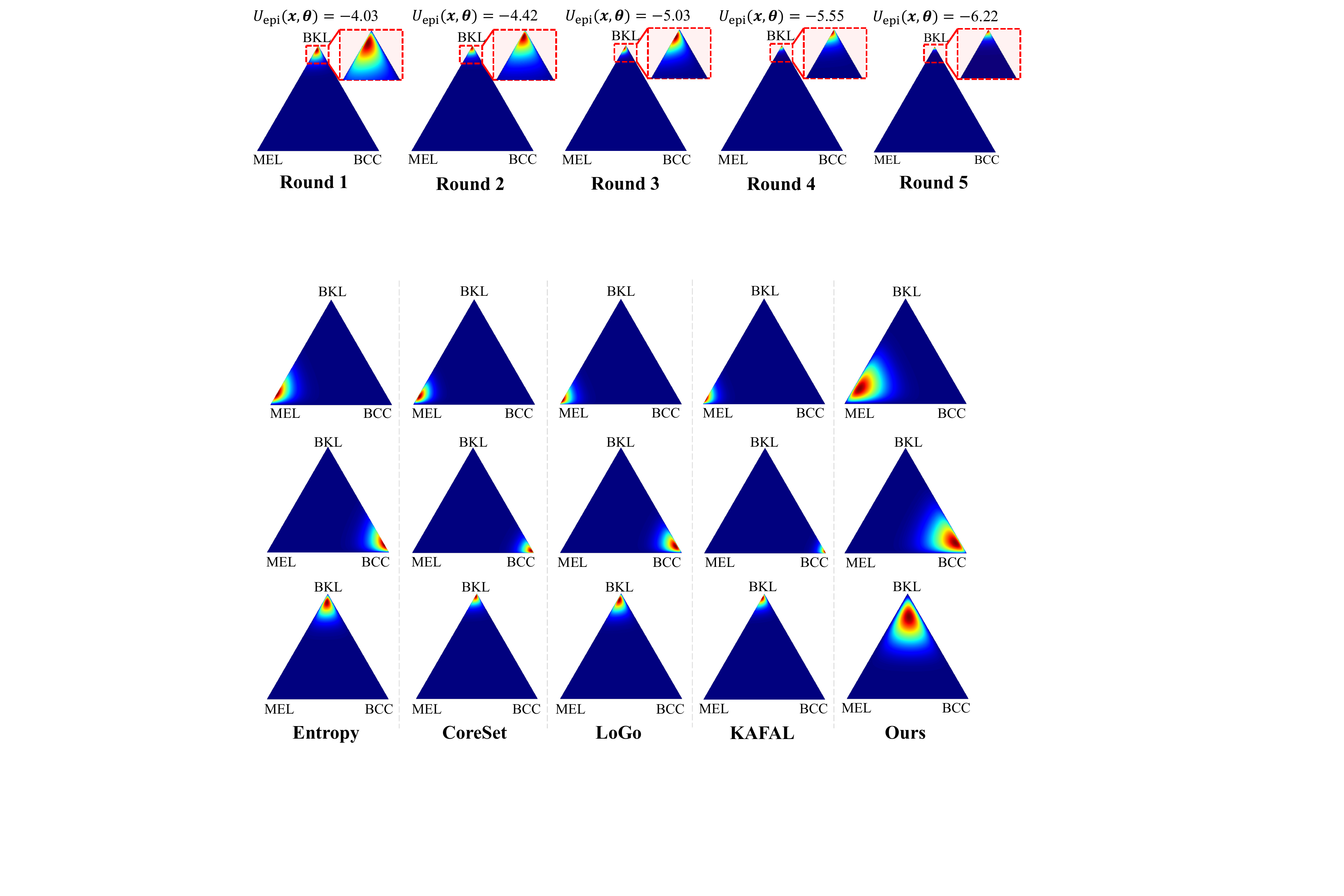}
    \caption{Visualization of the Dirichlet simplex for samples selected in the second FAL round using multiple sampling strategies.}
    \label{fig:diri_suppl}
\end{figure}

\paragraph{Analysis of Dirichlet simplex.}
We analyzed the Dirichlet simplex on a subset of the Fed-ISIC, specifically encompassing three classes: MEL, BCC, and BKL. In Fig.~\ref{fig:diri_5rounds} and Fig.~\ref{fig:diri_suppl}, the unlabeled samples are predicted to belong to MEL, BCC, and BKL from the first to the third rows, respectively. Within the Dirichlet simplex, a concentrated red region indicates low epistemic uncertainty $U_{\text{epi}}$, while a red region near the corner suggests a low aleatoric uncertainty $U_{\text{ale}}$. As depicted in Fig.~\ref{fig:diri_5rounds} and Fig.~\ref{fig:r1_r5}, when selecting samples with FEAL, the Dirichlet distribution becomes narrower and more concentrated for unlabeled local data from the first to the fifth FAL round. This trend suggests a reduction in epistemic uncertainty within the global model, validating the effectiveness of calibrated evidential sampling in mitigating domain shifts.
Moreover, starting with an identical set of labeled samples, we tracked the selection of samples in the second FAL round utilizing multiple FAL methods. The resulting Dirichlet simplexes, corresponding to these different methods, are depicted in Fig.~\ref{fig:diri_suppl}. A critical observation from this analysis is that the Dirichlet distribution of samples selected via FEAL exhibits a notably broader spread across the simplex. This broader spread indicates that FEAL effectively models the global model's understanding of local data and prioritizes the selection of samples characterized by high epistemic uncertainty.


\subsection{Evaluation Time Costs}
All experiments were conducted using an NVIDIA GeForce RTX 2080Ti GPU. The average time cost for one round of data selection across all clients is presented in Tab.~\ref{tab:eval_time}. Note that we reported the time cost for the ensemble settings of Entropy, CoreSet, TOD, Gradnorm, and BADGE.
\begin{table}[H]
\centering
\caption{Time cost (in seconds) for one round of data selection.}
\scriptsize
\setlength{\tabcolsep}{3pt}
\begin{tabularx}{\linewidth}
{l|ccccc}
\bottomrule
\multicolumn{1}{c|}{Method} & Fed-ISIC & Fed-Camelyon & Fed-Polyp & Fed-Prostate & Fed-Fundus\\
\hline
Entropy & 23.49 & 189.97 & 8.11 & 14.20 & 8.39\\
CoreSet & 81.27 & 340.24 & 9.68 & 14.56 & 8.30\\
TOD & 23.26 & 331.01& 14.50 & 55.71 & 28.53 \\
Gradnorm & 1126.95 & 5335.42 & 72.49 & 128.49 & 67.42\\
BADGE & 24.51 & 178.66 & 68.12 & 119.74 & 69.17\\
LoGo & 96.22 & 378.81 & 86.52 & 80.21 &43.20\\
KAFAL & 23.27 & 175.95 & 14.54 & 13.35 & 7.45\\
\rowcolor{gray!20}FEAL (Ours) & 21.76 & 191.71 & 14.88 & 13.59 & 13.98\\
\toprule
\end{tabularx}
\label{tab:eval_time}
\end{table}

\section{Investigations on OCTA Datasets}
\label{sec:octa}
\subsection{Experimental Settings}
\begin{figure}[htbp]
    \centering
    \includegraphics[width=\linewidth]{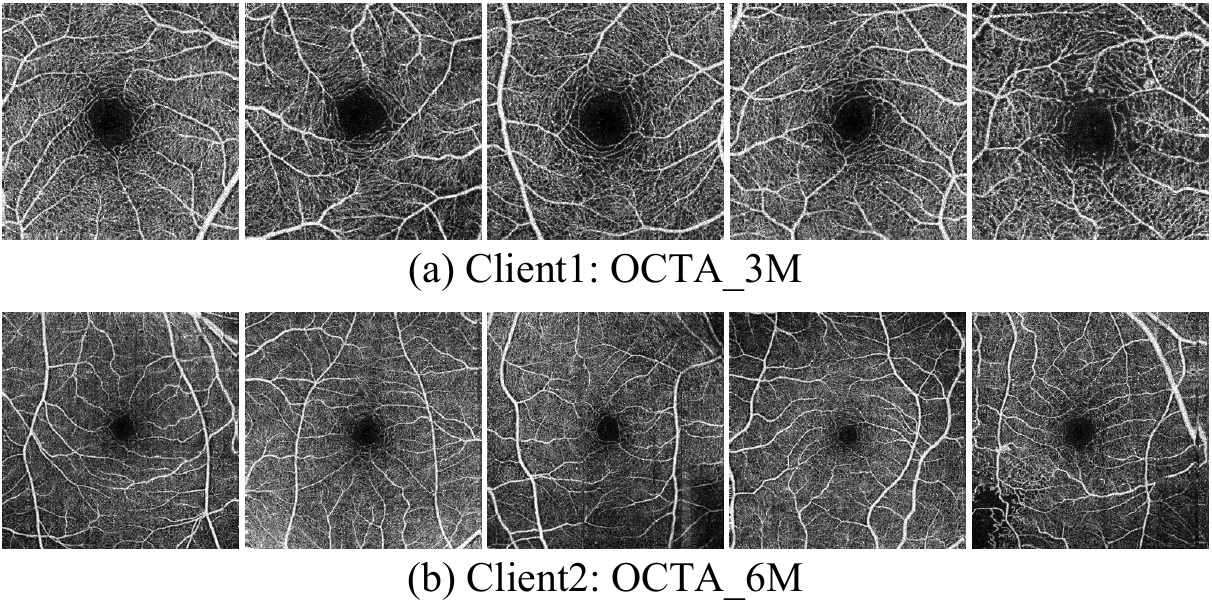}
    \caption{Illustrative samples from each data source within the OCTA-500 dataset.}
    \label{fig:octa_example}
    \vspace{-2mm}
\end{figure}

\paragraph{Dataset.} We further validated the effectiveness of FEAL on another medical image dataset OCTA-500~\cite{li2024octa} for foveal avascular zone (FAZ) segmentation. The OCTA-500 dataset comprises two subsets, namely OCTA\_3M and OCTA\_6M, each providing a distinct field of view (FOV). Specifically, the field of views (FOV) for OCTA\_3M is $3mm\times 3mm\times 2mm$, while that for OCTA\_6M is $6mm\times 6mm\times 2mm$. In our study, we regarded each subset as an individual local dataset within federated scenarios and then divided each local dataset into training and test sets using an 8:2 ratio. Note that we utilized the projection maps between the Internal Limiting Membrane (ILM) layer and the Outer Plexiform Layer (OPL) in experiments. Details of the OCTA-500 dataset are provided in Tab.~\ref{tab:octa500}, while illustrative samples from both data sources are shown in Fig.~\ref{fig:octa_example}.
\begin{table}[htbp]
    \centering
    \caption{Details of multi-center datasets utilized in our study.}
    \setlength{\tabcolsep}{4pt}
    \footnotesize
    \begin{tabularx}{\linewidth}{clccc}
    \bottomrule
        Dataset & \multicolumn{1}{c}{Data source} & \# Train & \# Test & Resolution\\
        \hline
        \multirow{2}{*}{OCTA-500} & Client 1: OCTA\_3M~\cite{li2024octa}&160 & 40 & $304{\times} 304$\\
        ~ & Client 2: OCTA\_6M~\cite{li2024octa}& 240 & 60 & $400{\times}400$\\
    \toprule
    \end{tabularx}
    \label{tab:octa500}
    \vspace{-2mm}
\end{table}

\paragraph{Evaluation metrics.} In line with the aforementioned three image segmentation datasets, we employed the Dice score and the 95\% Hausdorff Distance (HD95) as metrics to quantify segmentation results.

\paragraph{Implemental details.} We conducted $R=5$ rounds of FAL, which comprises federated model training and data annotation. For federated model training, We adopted U-Net~\cite{ronneberger2015u,liu2021feddg} as the backbone and employed the $ReLU(\cdot)$ as the non-negative activation function $\mathcal{A}(\cdot)$ for both global and local models. We trained local models using the Adam optimizer~\cite{kingma2014adam} with a learning rate of $5e{-}4$ and a weight decay of $1e{-}5$. The federated learning process comprises $T=100$ rounds of communication to attain a robust global model, with each local training session lasting for $1$ epochs. Regarding data annotation, the annotation budget $B_k$ was set to $20$ for the OCTA-500 dataset. 
\subsection{Results}
\begin{figure}[htbp]
    \centering
    \begin{subfigure}{0.49\linewidth}
        \centering
        \includegraphics[width=\linewidth]{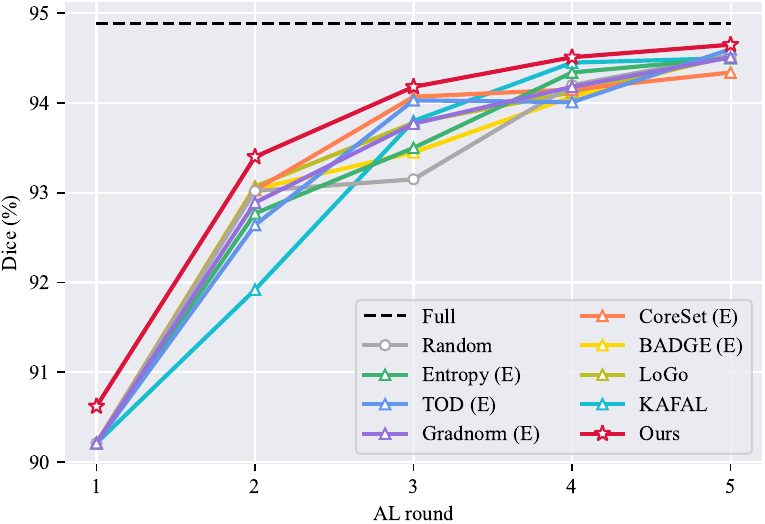}
        \caption{Dice of OCTA-500 ($E$)}
    \end{subfigure}
    \begin{subfigure}{0.49\linewidth}
        \centering
        \includegraphics[width=\linewidth]{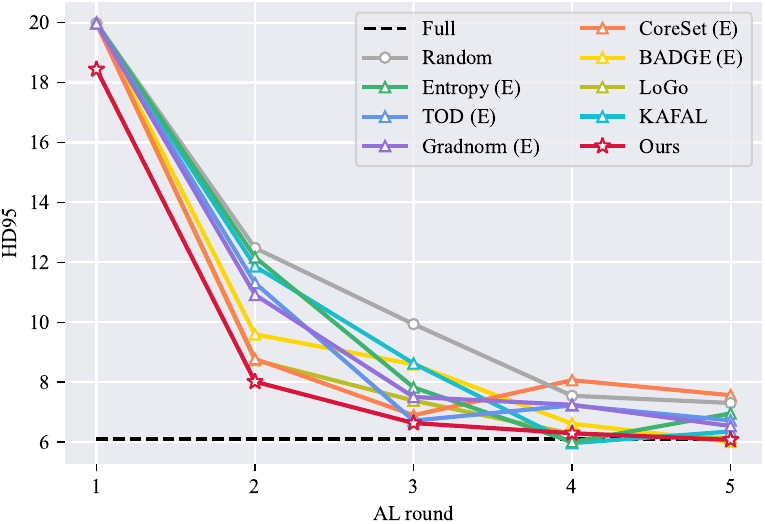}
        \caption{HD95 of OCTA-500 ($E$)}
    \end{subfigure}

    \caption{Comparison results on the OCTA-500 dataset.}
    \label{fig:octa}
\end{figure}

We compared FEAL with eight state-of-the-art FAL approaches and present the results in Fig.~\ref{fig:octa}. The results in Fig.~\ref{fig:octa} verify the effectiveness of FEAL on the OCTA-500 dataset, characterized by superior Dice scores and lower HD95 metrics. It is noteworthy that FEAL achieves a Dice score of $94.18\%$ while utilizing only $24\%$ of annotated samples, which is equivalent to $99.25\%$ of the fully supervised performance.

\section{Discussion with EDL and DUC}
\label{sec:discuss}
\paragraph{Discussion with EDL.} 
While Sensoy \textit{et al.}~\cite{sensoy2018evidential} leveraged evidential deep learning (EDL) to quantify the overall uncertainty of training samples and alleviate model overconfidence, our FEAL differs significantly in both objectives and approaches. Specifically, we employed evidential deep learning to decompose the overall uncertainty into aleatoric and epistemic components, aiming to measure the uncertainty of unlabeled samples and reduce annotation costs in federated learning scenarios with domain shifts.

\paragraph{Discussion with DUC.}
Although DUC~\cite{xie2023dirichlet} employs evidential deep learning to differentiate between aleatoric uncertainty and epistemic uncertainty, there are notable differences in both objectives and methods compared to our FEAL. In terms of objectives, DUC aims to enhance domain adaptation by annotating partial samples from the target domain, whereas FEAL focuses on reducing annotation costs for local clients in realistic medical federated scenarios. Regarding informativeness measurement, DUC evaluates the data informativeness based on a model trained on the source domain, whereas FEAL quantifies data informativeness leveraging both global and local models in federated scenarios. Furthermore, DUC is an uncertainty-based method that overlooks the diversity of selected target samples. By contrast, FEAL identifies and annotates samples considering both uncertainty and diversity measures.
\end{document}